\documentclass[10pt,twocolumn,letterpaper]{article}

\usepackage{cvpr}              

\usepackage{times}
\usepackage{epsfig}
\usepackage{graphicx}
\usepackage{amsmath}
\usepackage{amssymb}
\usepackage{multirow}
\usepackage{multicol}
\usepackage{color}
\usepackage{xcolor}
\usepackage{pifont}
\usepackage{mathtools}
\usepackage{booktabs}
\usepackage{float}
\usepackage[table]{xcolor}

\usepackage[linesnumbered,ruled,vlined]{algorithm2e}
\newtheorem{proposition}{Proposition}
\newtheorem{theorem}{Theorem}

\definecolor{cvprblue}{rgb}{0.21,0.49,0.74}
\usepackage[pagebackref,breaklinks,colorlinks,allcolors=cvprblue]{hyperref}

\def\paperID{****} 
\def\confName{CVPR}
\def\confYear{2026}

\title{Residual Diffusion Bridge Model for Image Restoration}

\author{
	Hebaixu Wang$^{1,3}$, Jing Zhang$^{2,3}$\textsuperscript{\dag}, Haoyang Chen$^{2,3}$, Haonan Guo$^{3,4}$, Di Wang$^{2,3}$, \\ Jiayi Ma$^{1,3,5}$\textsuperscript{\dag}, and Bo Du$^{2,3}$\textsuperscript{\dag}\\
	$^1$School of  Electronic Information, Wuhan University, Wuhan, China\\ 
	$^2$School of Computer Science, Wuhan University, Wuhan, China\\ 
	$^3$Zhongguancun Academy, Beijing, China\\
	$^4$State Key Laboratory of Information Engineering in Surveying, Mapping and Remote Sensing, \\
	Wuhan University, Wuhan, China\\ 
	$^5$School of Robotics, Wuhan University, Wuhan, China\\ 
	{\tt\small $\!$$\!$\{wanghebaixu,$\!$jingzhang.cv,$\!$jyma2010\}@gmail.com;$\!$\{haoyangchen,$\!$haonan.guo,$\!$d\_wang,$\!$dubo\}@whu.edu.cn\!
	}
}
\begin{document}
\maketitle
\begingroup
\renewcommand\thefootnote{\dag}
\footnotetext{Corresponding author.}
\endgroup
\begin{abstract}
	Diffusion bridge models establish probabilistic paths between arbitrary paired distributions and exhibit great potential for universal image restoration. Most existing methods merely treat them as simple variants of stochastic interpolants, lacking a unified analytical perspective. Besides, they indiscriminately reconstruct images through global noise injection and removal, inevitably distorting undegraded regions due to imperfect reconstruction. To address these challenges, we propose the {R}esidual {D}iffusion {B}ridge {M}odel (RDBM). Specifically, we theoretically reformulate the stochastic differential equations of generalized diffusion bridge and derive the analytical formulas of its forward and reverse processes. Crucially, we leverage the residuals from given distributions to modulate the noise injection and removal, enabling adaptive restoration of degraded regions while preserving intact others. Additionally, we unravel the fundamental mathematical essence of existing bridge models, all of which are special cases of RDBM and empirically demonstrate the optimality of our proposed models. Extensive experiments are conducted to demonstrate the state-of-the-art performance of our method both qualitatively and quantitatively across diverse image restoration tasks. Code is publicly available at https://github.com/MiliLab/RDBM.
\end{abstract} 

\section{Introduction}\label{sec:introduction}

\begin{figure}[t]
	\centering
	\includegraphics[width=0.96\linewidth]{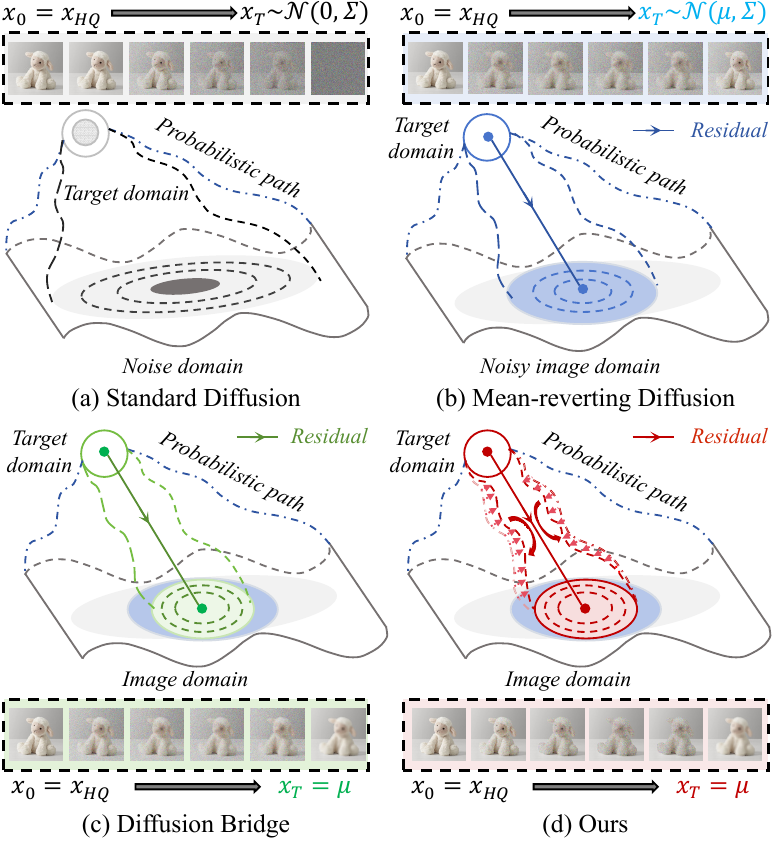}
	\vspace{-0.14in}
	\caption{Typical diffusion processes. (a) Standard diffusion maps high-quality images to Gaussian noise domain. (b) Mean-reverting diffusion drives terminal state toward a low-quality domain with stationary noise. (c) Diffusion bridge establishes direct probabilistic transitions between known distributions. All inject noise globally, disrupting overall structures and constraining transitions. (d) In contrast, our RDBM selectively reconstructs degraded regions (e.g., the doll) while preserving intact areas (e.g., the background).}
	\label{fig:intro}
\end{figure} 

Universal image restoration emerges as a unified paradigm integrating the perception, representation, and elimination of diverse degradations~\cite{jiang2024survey,wang2025deep}, with the aim of restoring high-quality images from the degraded low-quality observation. It typically encompasses a broad range of classical tasks, including denoising, deraining, dehazing, super-resolving, and others~\cite{chen2025towards,tian2024cross,kim2024frequency,wu2024seesr,goyal2024recent}. Owing to the high fidelity of restored details, it has been widely adopted as a precursor across various downstream tasks~\cite{marathe2022restorex,zhang2024ntire,wang2026universal,wang2025deep2}.

Diffusion models have achieved remarkable advances in universal image restoration~\cite{luo2025taming}. Early methods~\cite{songdenoising,xia2023diffir,zhu2023denoising} follow the standard diffusion paradigm~\cite{ho2020denoising,karras2022elucidating} that maps images to a Gaussian distribution, and initializes the reverse inference from pure noise. Some approaches leverage generative priors pretrained from large models~\cite{luo2023diff,ma2025efficient} as conditional guidance for denoising networks. Others treat various restoration tasks as inverse problems by assuming access to degradation kernels~\cite{chungdiffusion,chung2023parallel,wang2025dgsolver,zhang2025improving}. However, the randomness of noise and reliance on specific priors compromise both stability and universality. Subsequent studies~\cite{luo2023image,liu2024residual} incorporate mean-reverting dynamics into diffusion stochastic differential equations (SDEs), clustering forward terminal states around degraded observations to retain task-relevant cues. Additionally, diffusion bridges~\cite{zhoudenoising} directly model point-to-point stochastic transitions between paired distributions, thereby strengthening data associations and improving restoration fidelity. Despite these advances, existing methods still rely on global noise perturbation to construct probabilistic trajectories, requiring rigid reverse denoising processes, as shown in Fig.~\ref{fig:intro}. However, they fail to distinguish regions with varying degradation levels and imperfectly reconstruct intact regions, limiting restoration performance and adaptivity. Besides, a systematic and theoretical framework is absent to elucidate the intricate interconnections among existing diffusion bridge formulations.
 
In this work, we propose a scalable and unified diffusion bridge framework for universal image restoration, termed Residual Diffusion Bridge Model (RDBM), and conduct a comprehensive analysis of optimal distribution transitions. Specifically, we integrate the mean-reverting property of the Ornstein–Uhlenbeck SDEs with Doob’s $h$-transform~\cite{sarkka2019applied} to guide the terminal forward states toward degraded image distribution. Meanwhile, we use residuals from given distributions to dynamically modulate the probabilistic trajectories, thereby allowing the model to learn adaptive restoration of regions with varying degradation levels while mitigating redundant reconstruction in intact areas. Moreover, we theoretically demonstrate that our formulation yields the smooth distributional transition with respect to the residual-to-noise ratio. Building upon this, we uncover the mathematical essence of mainstream diffusion bridge formulations, all of which are special cases within our framework in specific configurations. Extensive experiments are conducted to verify the superiority of our method across diverse tasks including image restoration, translation, and inpainting. Our main contributions are summarized as follows:

\begin{itemize}
\item[1.] We propose a scalable and unified diffusion bridge framework for image restoration. Theoretically, it is characterized as generalized stochastic interpolants that delineate probabilistic transitions between any paired distributions.
\item[2.] Benefiting from the certainty of terminal states, we exploit residuals from paired distributions to modulate noise injection and removal, enabling selective reconstruction of degraded regions while preserving intact areas.
\item[3.] We unify and reinterpret existing bridge models as special instances of our RDBM framework, and substantiate its generality and effectiveness through extensive theoretical analysis and empirical validation.
\end{itemize}

\section{Related Work}\label{sec:relate_work}
Denoising diffusion models~\cite{songscore,songdenoising} were initially developed for image generation. Methods such as DiffIR~\cite{xia2023diffir}, DvSR~\cite{whang2022deblurring}, and SR3~\cite{zamir2020learning} directly repurpose diffusion models conditioned on degraded images for image restoration task, suffering from performance bottlenecks for task incompatibility. I2SB~\cite{liu20232} and ColdDiffusion~\cite{bansal2023cold} bypass explicit noise perturbations and instead learn a degraded diffusion process directly via the network. Besides, RDDM~\cite{liu2024residual}, ResShift~\cite{yue2024efficient}, ResFusion~\cite{shi2024resfusion}, and DiffUIR~\cite{zheng2024selective} incorporate prior distributions into the perturbation kernels to explicitly characterize degradation-aware diffusion processes. Moreover, IRSDE~\cite{luo2023image} employs a mean-reverting process to enforce diffusion trajectories that regress toward noisy degraded images with stationary variance. DDBM~\cite{zhoudenoising}, BBDM~\cite{li2023bbdm} and GOUB~\cite{yue2024image} further apply Doob’s $h$-transform to remove terminal noise, offering a tractable alternative to pave the probability path that connects degraded and clean images, thereby achieving remarkable restoration performance. Flow matching~\cite{lipman2023flow,liuflow,erbach2025solving} discards the stochastic noise and constructs the deterministic distribution transition path, thereby facilitating the optimal transport~\cite{zhu2025diffusion}. In contrast, our RDBM introduces the residual to modulate noise perturbation, enabling spatially adaptive restoration. Besides, RDBM can extend to these diffusion bridge models and flow matching in specific settings.

\begin{figure*}[t]
	\centering
	\includegraphics[width=0.78\linewidth]{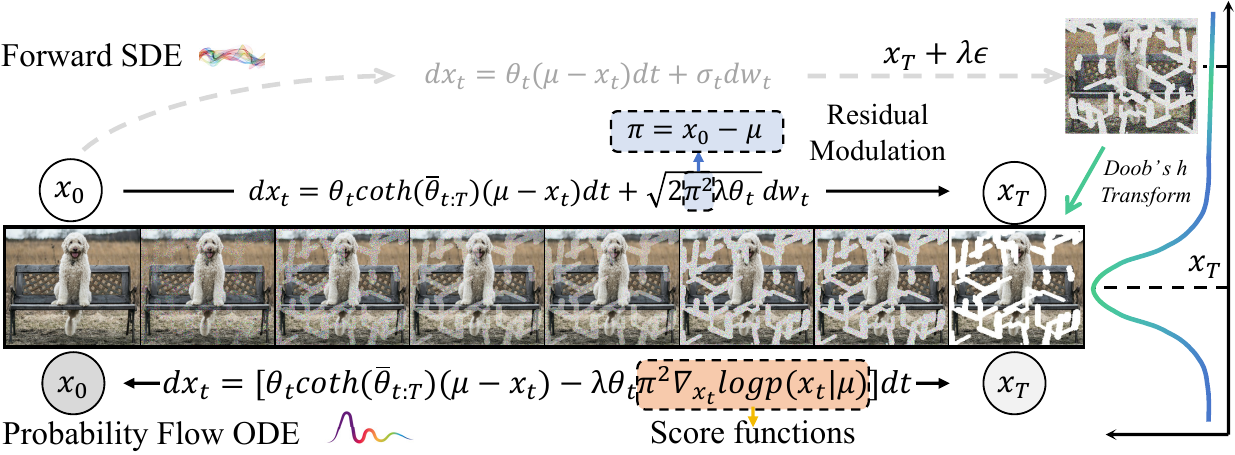}
	\vspace{-0.14in}
	\caption{A schematic of Residual Diffusion Bridge Models. RDBM utilizes a diffusion process guided by Doob’s $h$-transform towards an endpoint $\mathbf{x}_T = \boldsymbol{\mu}$ free from stationary noise $\lambda \epsilon$. Modulated by the residual component $\boldsymbol{\pi} = \mathbf{x}_0 - \mathbf{x}_T$, the noise perturbation is selectively imposed on different regions with diverse degradation levels, thereby constructing probabilistic paths. Besides, it learns to reverse the process by matching the residual bridge score functions, facilitating an adaptive inversion from $\mathbf{x}_T\sim p_{prior}(\boldsymbol{x})$ to $\mathbf{x}_0 \sim p_{data}(\boldsymbol{x})$. }
	\label{fig:method}
\end{figure*} 

\section{Background}\label{sec:background}
\subsection{Diffusion Bridge Models}\label{sec:DBM}
Diffusion SDEs~\cite{sohl2015deep,ho2020denoising} with drift term $\mathbf{f}(\cdot,t)$ and diffusion term $g(t)$ can be generally formulated as~\cite{karras2024analyzing}:
\begin{align}\label{sec:background_eq1}
	d \mathbf{x}_t = \mathbf{f}(\mathbf{x}_t,t)dt + g(t)d\omega_t,
\end{align}
where $\omega_t$ is the standard Wiener process. Eq.~\eqref{sec:background_eq1} describes the stochastic process from initial data $\mathbf{x}_0\sim p_{data}(\mathbf{x})$ to a prior distribution $\mathbf{x}_T\sim p_{prior}(\mathbf{x})$. Its reverse SDEs and probability flow ordinary differential equations (ODEs) that share the same marginal distributions can be derived as:
\begin{align}
	d\mathbf{x}_t &= [\mathbf{f}(\mathbf{x}_t,t) - g^2(t)\nabla_{\mathbf{x}_t} \log p(\mathbf{x}_t)]dt+g(t)d\overline{\omega}_t, \label{sec:background_eq2} \\
	d\mathbf{x}_t &= [\mathbf{f}(\mathbf{x}_t,t) - \frac{1}{2}g^2(t)\nabla_{\mathbf{x}_t} \log p(\mathbf{x}_t)]dt, \label{sec:background_eq3}
\end{align}
where $\nabla_{\mathbf{x}_t} \log p(\mathbf{x}_t)$ is score function. Furthermore, a diffusion process defined in Eq.~\eqref{sec:background_eq1} can be
driven to arrive at a particular point of interest $\boldsymbol{\mu}$ via Doob’s $h$-transform~\cite{rogers2000diffusions}:
\begin{equation}\label{sec:background_eq4}
	d\mathbf{x}_t = [\mathbf{f}(\mathbf{x}_t,t)+ g(t)^2\mathbf{h}(\mathbf{x}_t,t,\mathbf{x}_T,T)]dt + g(t)d\omega_t,
\end{equation}
where $\mathbf{h}(\mathbf{x}_t,t,\mathbf{x}_T,T) = \nabla_{\mathbf{x}_t} \log p(\mathbf{x}_T\vert \mathbf{x}_t)$ is the gradient of the log transition kernel from $t$ to $T$ generated by the original SDE. When both the initial state $\mathbf{x}_0$ and terminal state $\mathbf{x}_T \! =\!\boldsymbol{\mu}$ are fixed, Eq.~\eqref{sec:background_eq4} defines a stochastic process known as a diffusion bridge (see proof in Suppl.~\ref{sec:suppl_A}).

\subsection{Ornstein Uhlenbeck Process}\label{sec:OUP}
Ornstein–Uhlenbeck (OU) process is a stationary Gaussian-Markov process, with its marginal distribution converging toward a stable mean $\boldsymbol{\mu}$ with fixed variance over time. Formally, the OU process is generally defined as follows:
\begin{equation}\label{sec:background_eq5}
	d \mathbf{x}_t = \theta_t (\boldsymbol{\mu} - \mathbf{x}_t) dt + \sigma_t d \omega_t,
\end{equation}
where $\theta_t$ and $\sigma_t$ respectively denote time-dependent drift and diffusion coefficients that characterize the speed of the mean-reversion. The transition probability of Eq.~\eqref{sec:background_eq5} admits a closed-form solution as below:
\begin{align}
	p( \mathbf{x}_t \mid \mathbf{x}_s )	&=\mathcal{N}(\mathbf{\bar m}_{s:t},\bar \sigma_{s:t}^2\boldsymbol{I})=,\label{sec:background_eq6}\\
	\mathcal{N}( \boldsymbol{\mu} +( \mathbf{x}_s - \boldsymbol{\mu} )& e^{-\bar{\theta}_{s:t}},\int_s^t \sigma_z^2 e^{-2\overline{\theta}_{z:t}} dz ) \label{sec:background_eq7}\\
	\bar{\theta}_{s:t} &= \int_s^t{\theta _zdz}\label{sec:background_eq8}.
\end{align}
Driven by the mean-reverting dynamics with Gaussian perturbations, the diffusion trajectory originates from $\mathbf{x}_0 \!\sim\! p_{data}(\mathbf{x})$ at initial time and gradually approaches $\mathbf{x}_T \!=\!\boldsymbol{\mu}\!\sim \! p_{prior}(\mathbf{x})$ at final time $T$. See Suppl.~\ref{sec:suppl_B} for details.

\begin{figure*}[t]
	\centering
	\includegraphics[width=0.96\linewidth]{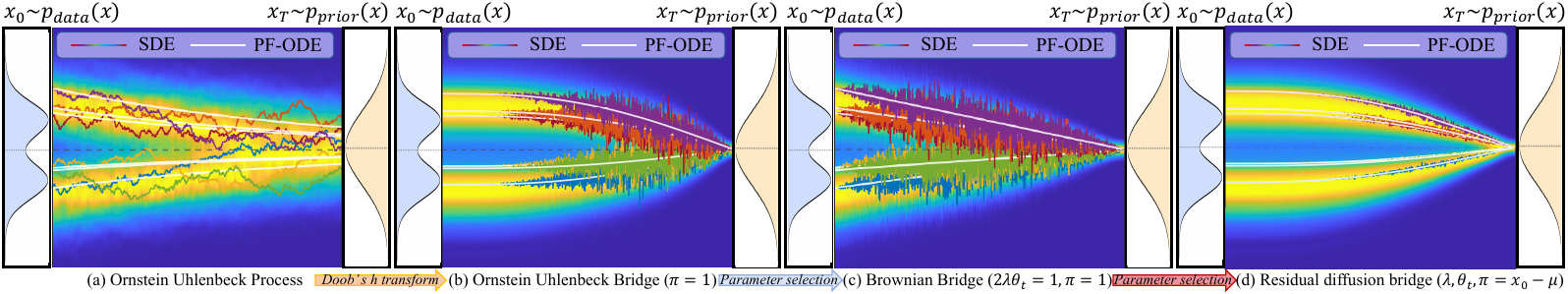}
	\vspace{-0.14in}
	\caption{Overview of mainstream diffusion processes via SDEs, all of which are special cases of our framework. (a) OU process maps the data distribution to the prior distribution with noise. (b) OU bridge constructs probabilistic transition paths between given distributions. (c) Brownian bridge models linear expectations of intermediate states. (d) Our RDBM leverages residuals from paired distributions to adaptively modify the transition trajectories, maintaining a smooth residual-to-noise ratio.}
	\label{fig:sdes}
\end{figure*}

\section{Residual Diffusion Bridge Models}\label{sec:method}
\subsection{Generalized Forward Process}\label{sec:GFP}
We redefine a OU process in Eq.~\eqref{sec:background_eq5} for generality:
\begin{equation}\label{sec:method_eq1}
	d \mathbf{x}_t = \theta_t (\boldsymbol{\mu} - \mathbf{x}_t) dt + \boldsymbol{\pi} \sigma_t d \omega_t,
\end{equation}
where $\boldsymbol{\pi}$ is a predefined value. By applying the Doob’s $h$-transform to Eq.~\eqref{sec:method_eq1}, we can establish the diffusion bridge that connects the high-quality image distribution $\mathbf{x}_0 \sim p_{HQ}(\boldsymbol{x})$ with degraded image distribution $\boldsymbol{\mu} \sim p_{LQ}(\boldsymbol{x})$: 
\begin{proposition}\label{prop_1}
Let $\mathbf{x}_t$ be a finite random variable governed by the generalized diffusion bridge process in Eq.~\eqref{sec:method_eq1}, with terminal condition $\mathbf{x}_T = \boldsymbol{\mu}$. The evolution of its marginal distribution $p(\mathbf{x}_t \mid \mathbf{x}_T)$ satisfies the following SDE under a fixed drift-to-diffusion coefficient ratio $\lambda = \sigma_t^2/(2\theta_t)$:
\begin{equation}\label{sec:method_eq2}
	d\mathbf{x}_t = \theta_t\coth(\overline{\theta}_{t:T})(\boldsymbol{\mu}-\mathbf{x}_t)dt + \sqrt{2\boldsymbol{\pi}^2\lambda \theta_t} d\omega_t,
\end{equation}
where $\overline{\theta}_{s:t} = \int_s^t \theta_z dz$ and $\boldsymbol{\pi}\in \mathbb{R}$. See Suppl.~\ref{sec:suppl_C}.
\end{proposition}
Consequently, Eq.~\eqref{sec:method_eq2} describes the generalized diffusion bridge models governed by $\lambda$, $\theta_t$ and $\boldsymbol{\pi}$. Here, $\lambda$ controls the global noise level, while $\theta_t$ and $\boldsymbol{\pi}$ jointly determine bridge category and dynamical evolution. Furthermore, we can derive its closed-form solution as follows:
\begin{proposition}\label{prop_2}
Given an initial state $\mathbf{x}_0$, the analytical solution of $\mathbf{x}_t$ at time $0<t<T$ of that SDE in Eq.~\eqref{sec:method_eq2} can be formulated as:
\begin{equation}\label{sec:method_eq3}
	\mathbf{x}_t \!=\! \boldsymbol{\mu} \!+\! (\mathbf{x}_0 \!-\! \boldsymbol{\mu})\frac{\sinh(\overline{\theta}_{t:T})}{\sinh(\overline{\theta}_{0:T})}\!+\!\int_0^t\!\! \sqrt{2\boldsymbol{\pi}^2\lambda\theta_s} \frac{\sinh(\overline{\theta}_{t:T})}{\sinh(\overline{\theta}_{s:T})} d\omega_s,
\end{equation}
which satisfies a Gaussian distribution with expectation $E[\mathbf{x}_t]$ and variance $Var[\mathbf{x}_t]$ (proof is provided in Suppl.~\ref{sec:suppl_C}):
\begin{align}
	&\!E[x_t] = \boldsymbol{\mu} \!+\! (\mathbf{x}_0 - \boldsymbol{\mu})\frac{\sinh(\overline{\theta}_{t:T})}{\sinh(\overline{\theta}_{0:T})} \!\coloneqq\! \boldsymbol{\mu} \!+\! (\mathbf{x}_0\!-\!\boldsymbol{\mu})\Theta_t, \label{sec:method_eq4}\\
	&\!Var[x_t]  = 2\boldsymbol{\pi}^2\lambda\frac{\sinh(\overline{\theta}_{0:t})\sinh(\overline{\theta}_{t:T})}{\sinh(\overline{\theta}_{0:T})} \coloneqq \boldsymbol{\pi}^2\Sigma_t^2. \label{sec:method_eq5}
\end{align}
\end{proposition}
Eq.~(\ref{sec:method_eq3}) unveils that the trajectory of probability is dictated by a weighted amalgamation of the residual and Gaussian noise. In order to delineate its temporal dynamic evolution, we define the residual-to-noise ratio (RNR) $R(t,i,j)$ for each pixel $i,j$ at time $t$ as follows (details are in Suppl.~\ref{sec:suppl_D}):
\begin{equation}\label{sec:method_eq6} 
	\!\!\!R(i,j,t) \!=\! \frac{[x_0(i,j)\!-\!\boldsymbol{\mu}(i,j)]^2}{2[\boldsymbol{\pi}(i,j)]^2 \lambda } \!\frac{\sinh(\overline{\theta}_{t:T})}{ \sinh(\overline{\theta}_{0:t})\sinh(\overline{\theta}_{0:T})},
\end{equation} 
which is governed by two terms. The first term depends on the residual component and fixed ratio $\lambda$. The second term is entirely determined by $\theta_t$ series and exhibits a monotonic decline, which diverges to infinity as time $t\!\rightarrow \!0$ and converges to an infinitesimal value as time $t\!\rightarrow \!T$. Previous works~\cite{yue2024image,zhoudenoising,luo2023image} typically set $\boldsymbol{\pi} = 1$, thereby performing the global noise perturbation to uniformly disrupt the overall structure of images. This induces two ill-posed issues: (i) degraded regions with varying levels are treated equally, and intact regions suffer redundant and imperfect reconstruction due to inevitable cumulative error in reverse process. (ii) pixel-wise numerator $[x_0(i,j)\!-\!\boldsymbol{\mu}(i,j)]^2$ may exhibit discontinuous jumps, potentially distorting the smooth monotonic decay of the residual-to-noise ratio. Therefore, to maintain the dynamic equilibrium in transmission trajectories, we fix $\boldsymbol{\pi} = \mathbf{x}_0 - \boldsymbol{\mu}$, thereby deriving our specific formulation within this framework with adaptive noise perturbation and pixel-independent residual-to-noise ratio:
\begin{equation}\label{sec:method_eq7} 
	R(t,i,j) = R(t) \propto \frac{\sinh(\overline{\theta}_{t:T})}{ \sinh(\overline{\theta}_{0:t})\sinh(\overline{\theta}_{0:T})}.
\end{equation} 
 
\subsection{Reverse Process and Training Objective}\label{sec:RPTO}
From Eq.~\eqref{sec:method_eq4}-\eqref{sec:method_eq5}, the transition probability distributions from initial state $\mathbf{x}_0$ to intermediate states  $\mathbf{x}_t$ and $\mathbf{x}_{t-1}$ are:
\begin{align}
	q(\mathbf{x}_t\vert \mathbf{x}_0,\boldsymbol{\mu}) &\!=\! \mathcal{N}(\boldsymbol{\mu} + (\mathbf{x}_0 - \boldsymbol{\mu})\Theta_{t},\boldsymbol{\pi}^2\Sigma^2_{t}\boldsymbol{I}),\label{sec:method_eq9} \\
	q(\mathbf{x}_{t-1}\vert \mathbf{x}_0,\boldsymbol{\mu}) &\!=\! \mathcal{N}(\boldsymbol{\mu} + (\mathbf{x}_0 - \boldsymbol{\mu})\Theta_{t-1},\boldsymbol{\pi}^2\Sigma^2_{t-1}\boldsymbol{I}), \label{sec:method_eq10} 
\end{align}
Supposing that sampling from $\mathbf{x}_t$ to $\mathbf{x}_{t-1}$ follows the Gaussian distribution, we leverage \textit{Bayes' theorem} to derive the deterministic sampling of reverse process (see Suppl.~\ref{sec:suppl_E}):
\begin{align}
	\!\! \mathbf{x}_{t-1} &\!=\! \boldsymbol{\mu} \!+\! {\frac{\Sigma_{t-1}}{\Sigma_{t}}}(\mathbf{x}_t \!-\!\boldsymbol{\mu}) \!+\! (\Theta_{t-1} \!-\! \Theta_{t} {\frac{\Sigma_{t-1}}{\Sigma_{t}}})\boldsymbol{\pi} ,\label{sec:method_eq11} \\
	& \!=\! \boldsymbol{\mu} \!+\! \frac{\Theta_{t-1}}{\Theta_t}(\mathbf{x}_t \!-\! \boldsymbol{\mu}) \!-\! ({\frac{\Theta_{t-1}}{\Theta_{t}}\Sigma_{t} \!-\! \Sigma_{t-1}})\boldsymbol{\pi} \epsilon_t,\label{sec:method_eq12}
\end{align}  
Apparently, Eq.~\eqref{sec:method_eq12} involves two unknowns, the residual $\boldsymbol{\pi}$ and the noise $\epsilon_t$. In theory, the distributions at all time steps should be aligned; thus, the overall training objective is:
\begin{equation}\label{sec:method_eq13}
	\mathcal{L}(\dot{\theta})=D_{KL}(q(\mathbf{x}_{t-1}\vert \mathbf{x}_t,\mathbf{x}_0,\boldsymbol{\mu}) || p_{\dot{\theta}}(\mathbf{x}_{t-1}\vert \mathbf{x}_t,\boldsymbol{\mu})).
\end{equation}
Assuming $p_{{\theta}}(\mathbf{x}_{t-1}\vert \mathbf{x}_t,\boldsymbol{\mu})$ follows a Gaussian distribution centered at $m_{\dot{\theta}}(\mathbf{x}_t,\mathbf{x}_0)$ with a constant variance, minimizing the Kullback-Leibler divergence~\cite{kingmaauto} $D_{KL}$ is equivalent to reducing the distance between the means (see Suppl.~\ref{sec:suppl_G}):
\begin{align}
	\mathcal{L}(\dot{\theta}) &\coloneqq  \mathbb{E}_{q(\mathbf{x}_t\vert \mathbf{x}_0)}[\eta_m\|m(\mathbf{x}_t,\mathbf{x}_0) - m_{\dot{\theta}}(\mathbf{x}_t,\mathbf{x}_0) \|] \label{sec:method_eq14}\\
	& \coloneqq \mathbb{E}_{\mathbf{x}_0,\boldsymbol{\mu},t}[\eta_\epsilon\|\boldsymbol{\pi}_\epsilon^{\dot{\theta}}(\mathbf{x}_t,t,\boldsymbol{\mu}) - (\mathbf{x}_0 - \boldsymbol{\mu})\epsilon_t\|], \label{sec:method_eq15}
\end{align}
where $\eta_m$ and $\eta_\epsilon$ are different weights for different training objectives. Accordingly, we can employ a neural network $\boldsymbol{\pi}_\epsilon^{\dot{\theta}}(\mathbf{x}_t,t,\boldsymbol{\mu})$ to predict the multiplication of residual and noise at once. The detailed algorithms for training and sampling are presented in Alg.~\ref{Alg:train} and Alg.~\ref{Alg:sample}, respectively.

\begin{table}[t]
	\caption{Connections to other mainstream bridge models.}\label{tab:analysis}
	\centering
	\vspace{-0.14in}
	\resizebox{0.98\columnwidth}{!}{
		\begin{tabular}{cccc}
			\toprule
			\multicolumn{3}{c}{Diffusion Bridge Configurations}  & \multirow{1}{*}{{Method}}\\
			\midrule 
			$\theta_t\rightarrow 0$ & $\lambda$ & $\boldsymbol{\pi} = 0$  &  Flow Matching~\cite{lipman2023flow,liuflow}\\
			$\theta_t\rightarrow 0$ & $\lambda\rightarrow \infty$ & $\boldsymbol{\pi} = 1$  &  VE Bridge~\cite{zhoudenoising} \\ 
			$\theta_t\rightarrow 0$ & $\lambda \rightarrow \frac{1}{2}$ & $\boldsymbol{\pi} = 1$  &  VP Bridge~\cite{zhoudenoising} \\
			$\theta_t\rightarrow 0$ & $2\lambda\theta_t \rightarrow 1$ & $\boldsymbol{\pi} = 1$  &  Brownian Bridge~\cite{li2023bbdm,liu20232}\\
			$\theta_t$ & $\lambda$ & $\boldsymbol{\pi} = 1$ & OU Bridge~\cite{yue2024image} \\
			$\theta_t$ & $\lambda$ & $\boldsymbol{\pi} = \mathbf{x}_0 - \boldsymbol{\mu} $ & Ours \\
			\bottomrule
		\end{tabular}
	}
\end{table}

\subsection{Analysis}\label{sec:analysis}
We redefine a general mean-reverting process in Eq.~\eqref{sec:method_eq1} and employ Doob's $h$ transform to derive the generalized diffusion bridge in Eq.~\eqref{sec:method_eq2} that exhibits the property of mean-arrival. Probability paths of several diffusion processes are shown in Fig.~\ref{fig:sdes}. We configure  $\boldsymbol{\pi}$  to serve as residual component for adaptive noise perturbation, yielding a smoothly decaying RNR. Besides, other mainstream bridge models can be concluded in our framework, such as standard diffusion bridge~\cite{zhoudenoising}, Brownian Bridge~\cite{li2023bbdm,liu20232}, OU Bridge~\cite{yue2024image}, Flow Matching~\cite{lipman2023flow,liuflow} and others, as summarized in Tab.~\ref{tab:analysis} (See Suppl.~\ref{sec:suppl_F}).

\begin{algorithm}[t]
	\caption{Training.}
	\label{Alg:train}
	\KwIn{Clean image $\mathbf{x}_0$; Degraded image: $\boldsymbol{\mu}$; Residual map: $\boldsymbol{\pi} = \mathbf{x}_0 - \boldsymbol{\mu}$.}
	
	\Repeat{converged}{
		$\mathbf{x}_0 \sim q(\mathbf{x}_0)$;
		
		$t\sim Uniform(1,\cdots,T)$;
		
		$\epsilon\sim\mathcal{N}(0,\boldsymbol{I})$;
		
		$\mathbf{x}_t = \boldsymbol{\mu} + (\mathbf{x}_0-\boldsymbol{\mu})\Theta_t + \boldsymbol{\pi} \Sigma_t \epsilon$;
		
		Take the gradient descent step on
		
		$\phantom{******}\nabla_\theta \| \boldsymbol{\pi}\epsilon - \boldsymbol{\pi}_\epsilon^{\dot{\theta}}(\mathbf{x}_t,t,\boldsymbol{\mu})\|_1$
		
	}  
\end{algorithm}

\section{Experiments}
\subsection{Datasets and Evaluation Metrics}\label{sec_4_1}
Extensive experiments are conducted to assess the performance of our method on five image restoration tasks, including deraining, low-light enhancement, desnowing, dehazing, and deblurring. For fairness, we collect and mix the most widely used datasets for each task as follows. Besides, dataset details are as summarized in Suppl.~\ref{sec:suppl_H}.\\
\textbf{Image deraining.} We train our model on the merged datasets from Rain13K~\cite{Kui_2020_CVPR} and DeRaindrop~\cite{qian2018attentive}, which cover diverse rain streaks and densities. Evaluation is conducted on both rain- and raindrop-removal tasks using the mixed datasets~\cite{yang2017deep,zhang2018density,qian2018attentive, li2022toward}. In addition, we assess zero-shot generalization on real-world datasets, including GT-Rain~\cite{ba2022gt-rain} without ground-truth for reference.\\
\textbf{Low-light enhancement.} We combine the LOL~\cite{wei2018deep} and VE-LOL-L~\cite{ll_benchmark} datasets, which furnish real and synthetic paired samples across diverse scenes with varying illumination and noise levels. Additionally, we employ the  NPE~\cite{wang2013naturalness}, MEF~\cite{7120119} and DICM~\cite{lee2013contrast} datasets to conduct zero-shot generalization on real-world scenarios.\\
\textbf{Image desnowing.} We adopt the CSD~\cite{chen2021all} dataset as the primary benchmark and evaluate real-world performance on Snow100K-Real~\cite{liu2018desnownet}, which has no ground-truth.\\
\textbf{Image dehazing.} We adopt ITS\_v2~\cite{li2018benchmarking} and D-HAZY~\cite{Ancuti_D-Hazy_ICIP2016} as training benchmarks, encompassing diverse scenes under varying haze densities. The outdoor subset SOTS~\cite{li2018benchmarking} is used for evaluation, while real-world generalization is assessed on Dense-Haze~\cite{ancuti2019dense}, NHRW~\cite{zhang2017fast}, and NH-HAZE~\cite{NH-Haze_2020}.\\
\textbf{Image deblurring.} We use the GoPro~\cite{nah2017deep} dataset to perform deblurring tasks, which contains various levels of blur obtained by averaging the clear images captured in very short intervals. To further validate the generalizability, we perform zero-shot restoration on the RealBlur~\cite{rim_2020_ECCV} dataset.

Benchmarks are evaluated using peak signal-to-noise
ratio (PSNR)~\cite{huynh2008scope}, structural similarity (SSIM)~\cite{wang2004image},  natural image quality evaluator (NIQE)~\cite{lee2013contrast} in RGB space, and the learned perceptual image patch similarity (LPIPS)~\cite{snell2017learning} in feature space.  For fairness, we compare our method with several universal restoration methods~\cite{zamir2022restormer,li2022all,potlapalli2023promptir,zhang2023ingredient,luo2023image,jiang2024autodir,luo2024controlling,yue2024image,cui2024revitalizing,deng2025deepsn,rajagopalan2025awracle,li2025mair,ma2023prores}, which are all re-implemented on the mixed datasets for comparisons.

\subsection{Implementation Details}

\begin{figure*}[t]
	\centering
	\includegraphics[width=0.99\linewidth]{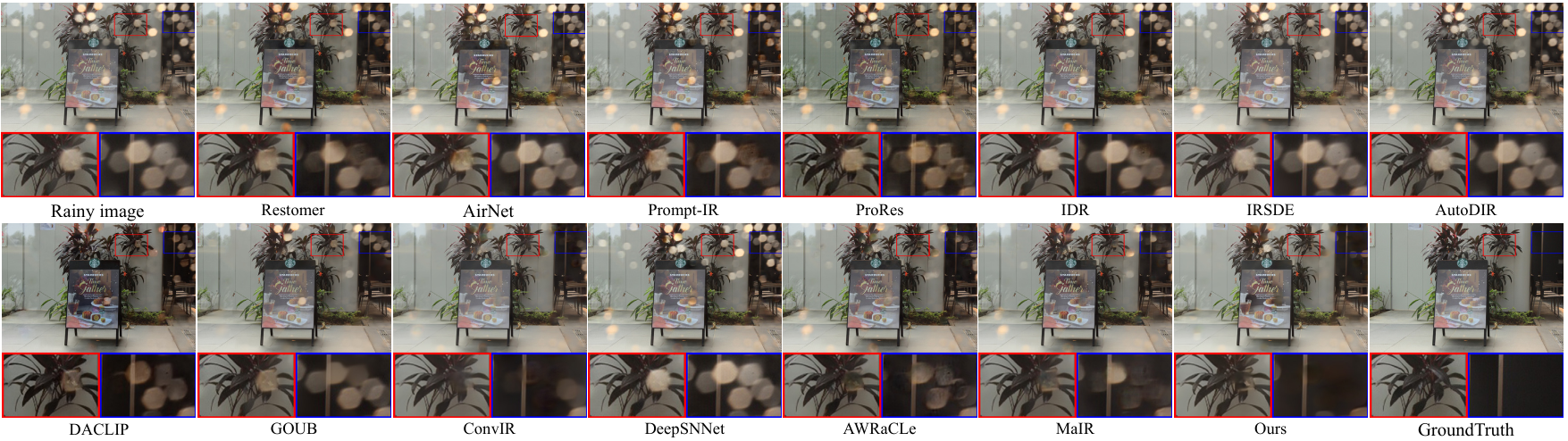}
	\vspace{-0.14in}
	\caption{Visualization comparison with state-of-the-art methods on deraining. Zoom in for best view.}
	\label{fig:rainy}
\end{figure*}

Our method is trained using 8 Nvidia A800 GPUs with PyTorch~\cite{paszke2019pytorch} framework for 128h. Adam optimizer and L1 loss are employed for 500k iterations with a learning rate of $1\text{e}\!-\!4$. We set the batch size as $20$ and distribute it evenly to each task. We randomly crop patches of size $256\times256$ from the original image as network input for training and use 10 timesteps for full-resolution testing. We utilize U-Net ~\cite{ronneberger2015u} architecture as network backbone. We change the channel number of the hidden layers $C$ to obtain different versions with varied parameter quantities:
\begin{itemize}
	\item[$\bullet$] RDBM-T: $C$=32, channel multiplier = \{1,1,1,1\}
	\item[$\bullet$] RDBM-S: $C$=32, channel multiplier = \{1,2,2,4\}
	\item[$\bullet$] RDBM-B: $C$=64, channel multiplier = \{1,2,2,4\}	
	\item[$\bullet$] RDBM-L: $C$=64, channel multiplier = \{1,2,4,8\}
\end{itemize}

\begin{algorithm}[t]
	\caption{Sampling.}
	\label{Alg:sample}
	\KwIn{Degraded image: $\boldsymbol{\mu}$; Neural network $\boldsymbol{\pi}_\epsilon^\theta(\cdot)$.}
		
	\For{\textnormal{$t=T$ to $1$}}{
		$\boldsymbol{\pi}\epsilon = \boldsymbol{\pi}_\epsilon^\theta(x_t,t,\boldsymbol{\mu})$
		
		\If{$t=T$}{
			$\!\mathbf{x}_{T-1} = \boldsymbol{\mu}$
		}
		\Else{
			$\!\mathbf{x}_{t-1} \!=\! \boldsymbol{\mu} \! +\! \frac{\Theta_{t-1}}{\Theta_t}(\mathbf{x}_t \!-\! \boldsymbol{\mu}) \!-\! ({\frac{\Theta_{t-1}}{\Theta_{t}}\Sigma_{t} \!-\! \Sigma_{t-1}})\boldsymbol{\pi} \epsilon$
		}
	}
	\textbf{end}
	
	\KwOut{$\mathbf{x}_{0}$.}  
\end{algorithm}

\begin{table*}[t]
	\caption{Quantitative comparisons of five image restoration tasks. The FLOPS is calculated in the inference stage with 256$\times$256 resolution. The best and second best results of universal models are shown in \textbf{\textcolor{red}{red}} and \textbf{\textcolor{blue}{blue}}, respectively.}\label{Tab_comparative}
	\vspace{-0.14in}
	\centering
	\renewcommand\arraystretch{1.0}
	\resizebox{2.1\columnwidth}{!}{
		\begin{tabular}{c|c|cc|cc|cc|cc|cc|cc|cc}
			\toprule
			\multirow{2}{*}{\textbf{Method}}  & \multirow{2}{*}{\textbf{Year}} & \multicolumn{2}{c|}{Deraining} & \multicolumn{2}{c|}{Enhancement} & \multicolumn{2}{c|}{Desnowing} &  \multicolumn{2}{c|}{Dehazing} &  \multicolumn{2}{c|}{Deblurring} &   \multicolumn{2}{c|}{Average} &  \multicolumn{2}{c}{Complexity} \\ 
			&	& PSNR$\uparrow$ & SSIM$\uparrow$ & PSNR$\uparrow$ & SSIM$\uparrow$ & PSNR$\uparrow$ & SSIM$\uparrow$ & PSNR$\uparrow$ & SSIM$\uparrow$ & PSNR$\uparrow$ & SSIM$\uparrow$ & PSNR$\uparrow$ & SSIM$\uparrow$  & Params(M) & FLOPs(G)\\
			\midrule
			Restomer~\cite{zamir2022restormer}& 2022  & 28.54 & 0.847 & 21.75 & 0.742 & 28.53 & 0.919 & 26.54 & 0.924 & 26.44 & 0.799 & 27.61  & 0.869 & 26.09 & 140.99 \\
			AirNet~\cite{li2022all} & 2022 & 24.78 & 0.774 & 13.05 &  0.485 & 25.80 & 0.885 & 18.53 & 0.827 & 25.76 & 0.782 & 24.01  &  0.809  & 5.76 & 301.27 \\
			Prompt-IR~\cite{potlapalli2023promptir}& 2023 & 28.97  & 0.856 & 20.97  & 0.733 & 29.52  & 0.938 & 25.80  & 0.929 & 26.25  & 0.797 & 27.89  & 0.878 & 32.96 & 158.14 \\
			ProRes~\cite{ma2023prores} & 2023 & 22.42  & 0.752 & 20.31  & 0.741 & 24.53  & 0.859 & 24.81  & 0.888 & 26.08  & 0.792  & 24.08  & 0.814  & 370.26 &  97.17 \\
			IDR~\cite{zhang2023ingredient} & 2023 & 28.40  & 0.844 & 20.95  & 0.706 & 27.77  & 0.911 & 24.48  & 0.914 & 26.33  & 0.799 & 26.96  & 0.863  & 6.19  & 32.16 \\
			IRSDE~\cite{luo2023image} & 2023 & 24.05  & 0.822  & 11.29  & 0.450  & 15.91  & 0.806  & 11.52  & 0.697  & 26.68  & 0.811  & 19.55  & 0.783 & 137.13 &   379.33\\
			AutoDIR~\cite{jiang2024autodir} & 2024  & 29.32  & 0.863  & 15.65  & 0.707  & 15.31  & 0.706 & 19.01 & 0.829 & \textcolor{blue}{\textbf{28.47}}  & \textcolor{blue}{\textbf{0.864}}  & 22.43  & 0.799 & 115.86 & 63.38 \\
			DA-CLIP~\cite{luo2024controlling}& 2024  & 28.63  & 0.854 & 19.50  & 0.730  & 28.23  & 0.934 & 27.26  & 0.941 & 26.47  & 0.818 & 27.54  & 0.881 & 32.96 & 158.14\\
			GOUB~\cite{yue2024image} & 2024 & 28.65  & \textcolor{blue}{\textbf{0.870}}  & 17.80  & 0.723  & 30.39  & \textcolor{blue}{\textbf{0.960}}  & 20.85  & 0.902  & 27.85  & 0.838  & 27.60  & 0.895 & 137.13 & 379.34 \\
			ConvIR~\cite{cui2024revitalizing} & 2024 & 29.18  & 0.867  & 21.36  & \textcolor{blue}{\textbf{0.771}}  & \textcolor{blue}{\textbf{31.43}}  & 0.950  & 29.13  & 0.960  & 28.41  & 0.862  & \textcolor{blue}{\textbf{29.49}}  & 0.903 & 14.82 & 128.93\\
			DeepSNNet~\cite{deng2025deepsn} & 2025 & 28.62  & 0.845  & 17.90  & 0.661  & 30.02  & 0.927  & 28.72  & 0.937  & 25.81  & 0.773  & 28.15  & 0.865  & 17.32 &71.79\\
			AWRaCLe~\cite{rajagopalan2025awracle} & 2025 & 29.15  & 0.860  & 20.41  & 0.756  & 27.70  & 0.927  & 18.38  & 0.789  & 26.37  & 0.818  & 26.31  & 0.861 & 94.18 & 165.42 \\
			MaIR~\cite{li2025mair} & 2025 & \textcolor{blue}{\textbf{29.45}} & 0.864  & \textcolor{blue}{\textbf{21.76}}  & 0.750  & 30.80  & 0.955  & \textcolor{blue}{\textbf{30.39}}  & \textcolor{blue}{\textbf{0.960}}  & 28.28  & 0.859  & 29.51  & \textcolor{blue}{\textbf{0.904}} & 20.71 & 110.44 \\
			\hline
			RDBM-T & - & 27.98  & 0.844  & 21.04  & 0.745  & 28.47  & 0.918  & 26.88  & 0.928  & 25.82  & 0.784  & 27.31  & 0.865  & 0.45 &  5.74\\
			RDBM-S & - & 29.23  & 0.864  & 21.98  & 0.765  & 30.93  & 0.941  & 28.92  & 0.942  & 26.67  & 0.808  & 28.99  & 0.886 & 1.07 &8.01  \\
			RDBM-B & - & 29.70  & 0.875  & 22.00  & 0.761  & 32.48  & 0.956  & 31.56  & 0.966  & 27.81  & 0.842  & 30.24  & 0.904 & 3.65  & 23.97\\  
			RDBM-L & - & \textcolor{red}{\textbf{30.31}}  & \textcolor{red}{\textbf{0.884}}  & \textcolor{red}{\textbf{24.53}}  & \textcolor{red}{\textbf{0.812}}  & \textcolor{red}{\textbf{32.59}}  & \textcolor{red}{\textbf{0.961}}  & \textcolor{red}{\textbf{33.45}}  & \textcolor{red}{\textbf{0.965}}  & \textcolor{red}{\textbf{29.04}}  & \textcolor{red}{\textbf{0.877}}  & \textcolor{red}{\textbf{31.04}}  & \textcolor{red}{\textbf{0.917}} & 7.73  & 32.93 \\
			\bottomrule
	\end{tabular}}
\end{table*}

\subsection{Comparative Experiments}
We compare our RDBM with several representative universal methods across five challenge image restoration tasks. \\
\textbf{Visual comparison.} The qualitative results are illustrated in Fig.~\ref{fig:rainy}. For more results, please refer to Suppl.~\ref{sec:suppl_H}. Obviously, our method generates high-quality results that are the most similar to ground-truth compared with other methods. \\
\textbf{Quantitative evaluation.} We present quantitative results in Tab.~\ref{Tab_comparative}. Clearly, RDBM-L attains great performance improvement across all tasks by a large margin, culminating in average gains of 1.55 dB in PSNR and 0.013 in SSIM. For fair comparisons, we also evaluate several lightweight RDBM variants. Notably, RDBM-B also attains the best average metrics with fewer parameters, highlighting the effectiveness of our design. Moreover, our models exhibit high scalability across different parameter levels. In conclusion, our method outperforms others and is the most competitive. 

\begin{figure*}[t]
	\centering
	\includegraphics[width=0.99\linewidth]{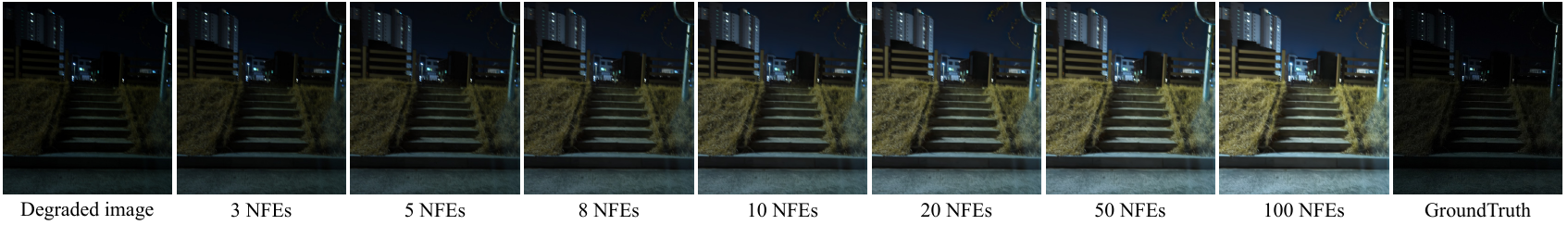}
	\vspace{-0.14in}
	\caption{Visualization results of different NFEs in a blurry night-time scene. Zoom in for best view.}
	\label{fig:compound}
\end{figure*}

\begin{figure*}[t]
	\centering
	\includegraphics[width=0.99\linewidth]{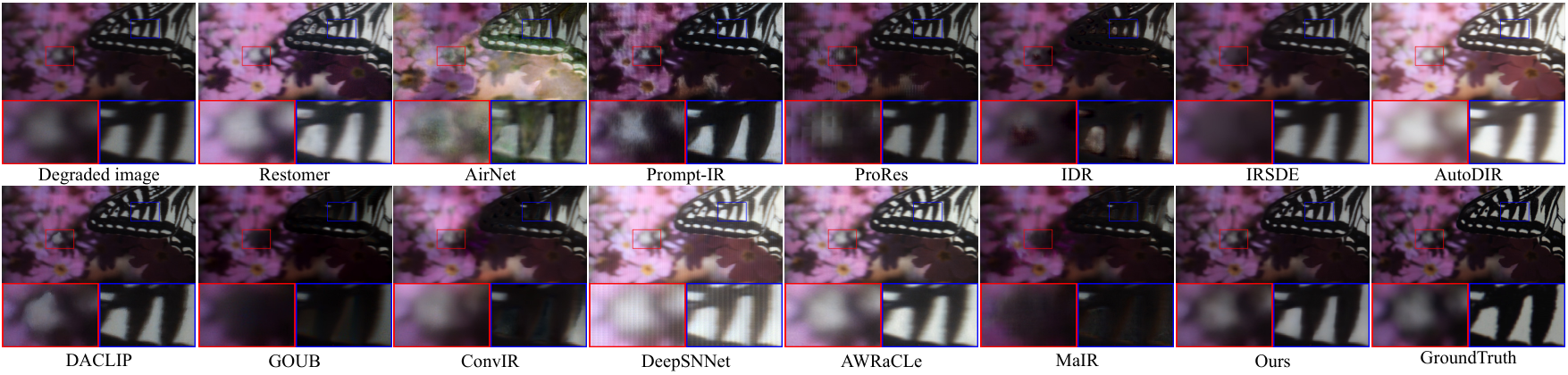}
	\vspace{-0.14in}
	\caption{Visualization results of zero-shot generalization in real-world TOLED dataset. Zoom in for best view.}
	\label{fig:unknown}
\end{figure*}

\begin{table}[t]
	\caption{Performance of different noise schedule ($\lambda = 10/255$).}\label{tab:noise_schedule}
	\vspace{-0.14in}
	\centering
	\renewcommand\arraystretch{1.0}
	\resizebox{1.0\columnwidth}{!}{
		\begin{tabular}{c|cc|cc|cc|cc|cc|cc}
			\toprule
			\multirow{2}{*}{\textbf{Schedule}}  & \multicolumn{2}{c|}{Deraining} & \multicolumn{2}{c|}{Enlighening} & \multicolumn{2}{c|}{Desnowing} &  \multicolumn{2}{c|}{Dehazing} &   \multicolumn{2}{c|}{Deblur} & \multicolumn{2}{c}{Average} \\ 
			& PSNR$\uparrow$ & SSIM$\uparrow$ & PSNR$\uparrow$ & SSIM$\uparrow$ & PSNR$\uparrow$ & SSIM$\uparrow$ & PSNR$\uparrow$ & SSIM$\uparrow$  & PSNR$\uparrow$ & SSIM$\uparrow$ & PSNR$\uparrow$ & SSIM$\uparrow$ \\
			\midrule 
			Linear & 29.63  & 0.878  & 22.39  & 0.774  & 34.15  & 0.965  & 32.00  & 0.958  & 28.48  & 0.864  & 30.99  & 0.912 \\
			\rowcolor{blue!10}
			Cosine & 30.31  & 0.884  & 24.53  & 0.812  & 32.59  & 0.961  & 33.45  & 0.965  & 29.04  & 0.877  & 31.04  & 0.917 \\
			Sigmoid & 29.31  & 0.868  & 22.91  & 0.782  & 33.80  & 0.961  & 32.06  & 0.970  & 28.63  & 0.869  & 30.84  & 0.911   \\
			\bottomrule
	\end{tabular}}
\end{table}

\subsection{Ablation Study}
To thoroughly explore the efficacy of our method, we carry out ablation studies encompassing three distinct categories: (i) the influence of various implementation configurations, (ii) the performance across different sampling steps, and (iii) the impact of diverse diffusion bridge settings.\\
\textbf{Influence of various implementation configurations.} Our RDBM formulations are governed by the schedule $\{\theta_t\}$ and stationary variance $\lambda$. Initially, we adopt the empirical choice $\lambda \!=\! \frac{10}{255}$~\cite{yue2024image} and compare performance across different noise schedules, as reported in Tab.~\ref{tab:noise_schedule}. It is evident that the optimal noise schedules differ for distinct restoration tasks, with the cosine schedule generally yielding the best results. Building on this finding, we further conduct a quantitative comparison among diverse stationary variance $\lambda$, as presented in Tab.~\ref{tab:lambda}. The results indicate that $\lambda \!=\! \frac{10}{255}$ with cosine noise schedule is the optimal configuration.

\begin{table}[t]
	\caption{Performance of varied stationary variance $\lambda$.}\label{tab:lambda}
	\vspace{-0.14in}
	\centering
	\renewcommand\arraystretch{1.0}
	\resizebox{1.0\columnwidth}{!}{
		\begin{tabular}{c|cc|cc|cc|cc|cc|cc}
			\toprule
			\multirow{2}{*}{\textbf{ $\lambda$}}  & \multicolumn{2}{c|}{Deraining} & \multicolumn{2}{c|}{Enlighening} & \multicolumn{2}{c|}{Desnowing} &  \multicolumn{2}{c|}{Dehazing} &   \multicolumn{2}{c|}{Deblur} & \multicolumn{2}{c}{Average} \\ 
			& PSNR$\uparrow$ & SSIM$\uparrow$ & PSNR$\uparrow$ & SSIM$\uparrow$ & PSNR$\uparrow$ & SSIM$\uparrow$ & PSNR$\uparrow$ & SSIM$\uparrow$  & PSNR$\uparrow$ & SSIM$\uparrow$ & PSNR$\uparrow$ & SSIM$\uparrow$ \\
			\midrule 
			1/255  & 30.03  & 0.884  & 23.87  & 0.823  & 31.86  & 0.957  & 31.27  & 0.959  & 28.89  & 0.874  & 30.36  & 0.915  \\
			\rowcolor{blue!10}
			10/255  & 30.31  & 0.884  & 24.53  & 0.812  & 32.59  & 0.961  & 33.45  & 0.965  & 29.04  & 0.877  & 31.04  & 0.917 \\
			20/255 & 29.86  & 0.879  & 22.20  & 0.756  & 33.40  & 0.962  & 31.52  & 0.966  & 28.08  & 0.850  & 30.66  & 0.909 \\
			50/255 & 29.94  & 0.884  & 22.61  & 0.782  & 32.66  & 0.964  & 29.51  & 0.950  & 28.55  & 0.867  & 30.26  & 0.913   \\
			100/255 & 29.98  & 0.880  & 23.49  & 0.794  & 30.09  & 0.950  & 27.03  & 0.940  & 28.53  & 0.865  & 29.08  & 0.906   \\
			\bottomrule
	\end{tabular}}
\end{table}

\noindent{\textbf{Performance across different sampling steps.}} Model efficiency and restoration quality hinge on the sampling steps, quantified by neural function evaluations (NFEs). We provide the restoration performance of different sampling steps in Tab.~\ref{tab:sampling_x0}. Clearly, our model exhibits varying performance across different NFEs. Initially, the restoration performance increases with more steps and peaks at 10 NFEs, reflecting accuracy gains from additional iterations. Beyond this threshold, performance gradually declines as NFEs rise. The underlying reasons are that our model is designed to handle diverse degradation types within a unified framework. In scenarios where samples exhibit multiple degradations, the model tends to prioritize the removal of the primary degradation before addressing secondary ones. Consequently, the restored output may deviate from the available reference, as shown in Fig~\ref{fig:compound}. In conclusion, we adopt 10 sampling steps to ensure performance and efficiency.
 
\begin{table}[t]
	\caption{Restoration performance of different sampling steps.}\label{tab:sampling_x0}
	\vspace{-0.14in}
	\centering
	\renewcommand\arraystretch{1.0}
	\resizebox{1.0\columnwidth}{!}{
		\begin{tabular}{c|cc|cc|cc|cc|cc|cc}
			\toprule
			\multirow{2}{*}{\textbf{NFE}}  & \multicolumn{2}{c|}{Deraining} & \multicolumn{2}{c|}{Enlighening} & \multicolumn{2}{c|}{Desnowing} &  \multicolumn{2}{c|}{Dehazing} &   \multicolumn{2}{c|}{Deblur} & \multicolumn{2}{c}{Average} \\ 
			& PSNR$\uparrow$ & SSIM$\uparrow$ & PSNR$\uparrow$ & SSIM$\uparrow$ & PSNR$\uparrow$ & SSIM$\uparrow$ & PSNR$\uparrow$ & SSIM$\uparrow$  & PSNR$\uparrow$ & SSIM$\uparrow$ & PSNR$\uparrow$ & SSIM$\uparrow$ \\
			\midrule 
			2 & 26.06  & 0.790  & 14.82  & 0.648  & 20.28  & 0.835  & 17.20  & 0.809  & 28.06  & 0.855  & 22.81  & 0.815\\
			5 & 30.05  & 0.876  & 23.35  & 0.809  & 30.47  & 0.947  & 29.01  & 0.929  & 29.13  & 0.878  & 29.61  & 0.905 \\
			\rowcolor{blue!10}
			10 & {30.31}  & {0.884}  & {24.53}  & {0.812}  & {32.60}  & {0.961}  & {33.45}  & {0.965}  & {29.04}  & {0.877}  & {31.04}  & {0.917}    \\
			20 & 30.10  & 0.882  & 24.35  & 0.811  & 31.96  & 0.959  & 32.25  & 0.961  & 28.94  & 0.875  & 30.58  & 0.915  \\
			50 & 29.92  & 0.880  & 24.21  & 0.809  & 31.59  & 0.958  & 31.56  & 0.958  & 28.85  & 0.873  & 30.28  & 0.913   \\
			100 & 29.84  & 0.879  & 24.13  & 0.808  & 31.49  & 0.957  & 31.39  & 0.957  & 28.80  & 0.873  & 30.19  & 0.912  \\
			\bottomrule
	\end{tabular}}
\end{table}

\begin{table}[t]
	\caption{Restoration performance of different $\pi$.}\label{tab:pi}
	\vspace{-0.14in}
	\centering
	\renewcommand\arraystretch{1.0}
	\resizebox{1.0\columnwidth}{!}{
		\begin{tabular}{c|cc|cc|cc|cc|cc|cc}
			\toprule
			\multirow{2}{*}{\textbf{$\pi$}}  & \multicolumn{2}{c|}{Deraining} & \multicolumn{2}{c|}{Enlighening} & \multicolumn{2}{c|}{Desnowing} &  \multicolumn{2}{c|}{Dehazing} &   \multicolumn{2}{c|}{Deblur} & \multicolumn{2}{c}{Average} \\ 
			& PSNR$\uparrow$ & SSIM$\uparrow$ & PSNR$\uparrow$ & SSIM$\uparrow$ & PSNR$\uparrow$ & SSIM$\uparrow$ & PSNR$\uparrow$ & SSIM$\uparrow$  & PSNR$\uparrow$ & SSIM$\uparrow$ & PSNR$\uparrow$ & SSIM$\uparrow$ \\
			\midrule 
			$0$  & 28.10  & 0.841  & 19.68  & 0.722  & 30.24  & 0.927  & 28.13  & 0.936  & 26.58  & 0.806  & 28.21  & 0.872\\
			$1$  & 29.56  & 0.876  & 21.71  & 0.749  & 32.79  & 0.957  & 30.74  & 0.961  & 27.66  & 0.838  & 30.15  & 0.903 \\
			\rowcolor{blue!10}
			$x_0 \!-\! x_T$ &  30.31  & 0.884  & 24.53  & 0.812  & 32.59  & 0.961  & 33.45  & 0.965  & 29.04  & 0.877  & 31.04  & 0.917  \\ 
			$|x_0 \!-\! x_T|$ & 30.36  & 0.883  & 24.37  & 0.812  & 32.40  & 0.957  & 33.19  & 0.965  & 28.99  & 0.876  & 30.94  & 0.915   \\ 
			\bottomrule
	\end{tabular}}
\end{table} 
 
\noindent{\textbf{Impact of diverse diffusion bridge settings.}} By appropriately selecting $\boldsymbol{\pi}$, our method can establish equivalence with other diffusion bridges. Hence, we perform the restoration performance comparisons with different $\boldsymbol{\pi}$ selections, as presented in Tab.~\ref{tab:pi}. The model is akin to flow matching as $\boldsymbol{\pi}=0$, yielding moderate results. It resembles stochastic interpolants and performs better as $\boldsymbol{\pi}=1$. Configuring $\boldsymbol{\pi}$ as the distributional residual or its absolute value is our formulation. These two variants produce similar results and achieve the best overall performance, thus verifying that residual bridge score matching offers a robust and effective paradigm for universal image restoration.

\begin{table}[t]
	\caption{Comparison under unknown tasks setting (under-display camera image restoration) on POLED and TOLED datasets.}\label{tab:real_unknown}
	\vspace{-0.14in}
	\centering
	\renewcommand\arraystretch{1.05}
	\resizebox{1.0\columnwidth}{!}{
		\begin{tabular}{c|cccc|cccc}
			\toprule
			\multirow{2}{*}{\textbf{Method}}  & \multicolumn{4}{c|}{POLED} & \multicolumn{4}{c}{TOLED} \\ 
			&PSNR$\uparrow$ & SSIM$\uparrow$ & MSE$\downarrow$ & LPIPS$\downarrow$ & PSNR$\uparrow$ & SSIM$\uparrow$ & MSE$\downarrow$ & LPIPS$\downarrow$ \\
			\midrule
			Restomer~\cite{zamir2022restormer} & 11.500  & 0.445 & 0.077  & 0.494 & 11.094  & 0.495 & 0.106  & 0.330  \\
			AirNet~\cite{li2022all} & 5.705  & 0.103 & 0.324  & 1.072 & 9.706  & 0.430 & 0.117  & 0.403   \\
			Prompt-IR~\cite{potlapalli2023promptir} & 11.589  & 0.429 & 0.075  & 0.541 & 13.088  & 0.504 & 0.105  & 0.336 \\
			ProRes~\cite{ma2023prores} & 10.284  & 0.433 & 0.103  & 0.473 & \textcolor{blue}{\textbf{28.452}} & \textcolor{blue}{\textbf{0.834}} & \textcolor{blue}{\textbf{0.002}}  & 0.212\\
			IDR~\cite{zhang2023ingredient} & 13.583  & 0.466 & 0.057  & 0.551 & 24.259  & 0.759 & 0.008  & 0.253  \\
			IRSDE~\cite{luo2023image} & \textcolor{blue}{\textbf{16.983}}  & \textcolor{blue}{\textbf{0.615}} & 0.029  & 0.475 & 27.163  & 0.811 & 0.002  & 0.243 \\
			AutoDIR~\cite{jiang2024autodir} & 8.627  & 0.404 & 0.151  & \textcolor{blue}{\textbf{0.406}} & 9.354  & 0.443 & 0.130  & 0.338   \\
			DA-CLIP~\cite{luo2024controlling} & 16.788  & 0.559 & \textcolor{blue}{\textbf{0.025}}  & 0.469 & 27.256  & 0.789 & 0.003  & \textcolor{red}{\textbf{0.201}}\\
			GOUB~\cite{yue2024image}& 12.922  & 0.525 & 0.053  & 0.446 & 23.177  & 0.761 & 0.007  & 0.269   \\
			ConvIR~\cite{cui2024revitalizing} & 9.370  & 0.429 & 0.130  & 0.477 & 13.659  & 0.558 & 0.091  & 0.316  \\
			DeepSNNet~\cite{deng2025deepsn}& 10.195  & 0.411 & 0.113  & 0.534 & 17.394  & 0.576 & 0.075  & 0.303\\
			AWRaCLe~\cite{rajagopalan2025awracle} & 11.208  & 0.431 & 0.091  & 0.513 & 10.540  & 0.495 & 0.121  & 0.331 \\
			MaIR~\cite{li2025mair} & 11.072  & 0.423 & 0.086  & 0.529 & 23.637  & 0.770 & 0.005  & 0.271\\
			RDBM & \textcolor{red}{\textbf{19.834}}  & \textcolor{red}{\textbf{0.715}} & \textcolor{red}{\textbf{0.012}}  & \textcolor{red}{\textbf{0.351}} & \textcolor{red}{\textbf{30.809}}  & \textcolor{red}{\textbf{0.870}} & \textcolor{red}{\textbf{0.001}}  & \textcolor{blue}{\textbf{0.202}}  \\
			\bottomrule
	\end{tabular}}
\end{table}  

\begin{figure*}[t]
	\centering
	\includegraphics[width=0.99\linewidth]{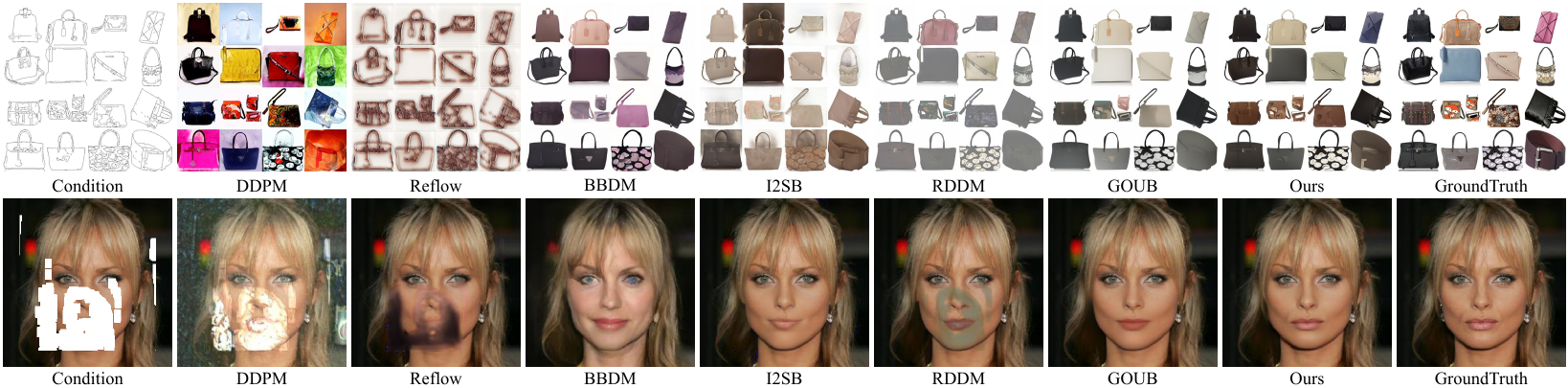}
	\vspace{-0.14in}
	\caption{Visualization results of image translation (top row) and image inpainting (bottom row). Zoom in for best view.}
	\label{fig:vision}
\end{figure*}

\subsection{Zero-Shot Real-world Generation}
To evaluate the generalization ability of our method, we do zero-shot generalization for unknown and known restoration tasks in real-world scenes. ``Unknown'' denotes cases where the degradation type is unspecified and may be compound, whereas ``known'' matches our task specification. As all methods are re-implemented on mixed datasets, they inherently handle diverse degradation types. In comparison, our method achieves strong performance in both settings.\\
\textbf{Unknown task generalization.} POLED and TOLED~\cite{zhou2021image} are captured by under-display cameras in high-resolution with different degradation types, which fully meet the real-world scene. The quantitative results are reported in Tab.~\ref{tab:real_unknown} while the visual comparisons are illustrated in Fig.~\ref{fig:unknown}. Evidently, our method achieves the best metric evaluation and our restored image is the most similar to ground-truth.\\
\textbf{Known task generalization.} As real-world datasets mainly have no ground truth, we use the non-reference metric, i.e., MetaIQA~\cite{zhu2020metaiqa} and NIQE~\cite{mittal2012making}, to assess the perceptual quality, as provided in Tab.~\ref{tab:known}. Results show that our method outperforms other universal models in various benchmarks and remains the most competitive. 

\begin{table}[t]
	\caption{Comparison under known task generalization setting.}\label{tab:known}
	\vspace{-0.14in}
	\centering
	\renewcommand\arraystretch{1.0}
	\resizebox{1.0\columnwidth}{!}{
		\begin{tabular}{c|cc|cc|cc|cc|cc}
			\toprule
			\multirow{2}{*}{\textbf{Method}}  & \multicolumn{2}{c|}{Deraining} & \multicolumn{2}{c|}{Enhancement} & \multicolumn{2}{c|}{Desnowing} & \multicolumn{2}{c|}{Dehazing} & \multicolumn{2}{c}{Deblurring} \\ 
			& MetaIQA$\uparrow$ &NIQE$\downarrow$  & MetaIQA$\uparrow$ &NIQE$\downarrow$ & MetaIQA$\uparrow$&NIQE$\downarrow$ & MetaIQA$\uparrow$&NIQE$\downarrow$ & MetaIQA$\uparrow$ &NIQE$\downarrow$ \\
			\midrule
			Restomer~\cite{zamir2022restormer}
			& 0.231 & 13.115 & 0.328 & 3.828 & 0.357 & 5.845 & 0.437 & 4.400 & 0.303 & 6.734\\
			AirNet~\cite{li2022all}
			& 0.232 & 11.668 & 0.280 & 3.674 & 0.347 & 6.091 & 0.440 & 4.623 & 0.286 & 6.393\\
			Prompt-IR~\cite{potlapalli2023promptir} 
			& 0.232 & 11.439 & 0.308 & 3.797 & 0.361 & 5.840 & 0.437 & 4.962 & 0.286 & 6.670\\
			ProRes~\cite{ma2023prores} 
			& 0.226 & 13.110 & 0.348 & 3.933 & 0.355 & 5.976 & 0.434 & 5.444 & 0.297 & 6.574\\
			IDR~\cite{zhang2023ingredient}
			& 0.231 & 11.100 & 0.324 & 3.866 & 0.363 & 5.850 & 0.453 & 4.634 & 0.300 & 6.683\\
			IRSDE~\cite{luo2023image} 
			&0.230  & 11.391 & 0.351 & 3.809 & 0.357 & 5.874 & 0.427 & 4.134 & 0.285 & 6.289\\
			AutoDIR~\cite{jiang2024autodir} 
			&0.231  & 10.800 & 0.366 & 3.910 & \textcolor{blue}{\textbf{0.373}} & 5.831 & \textcolor{blue}{\textbf{0.470}} & 9.881 & 0.308 & 6.493\\
			DA-CLIP~\cite{luo2024controlling} 
			&0.232  & 10.604 & 0.334 & 3.720 & 0.361 & 5.864 & 0.460 & 6.531 & 0.310 & \textcolor{blue}{\textbf{6.058}}\\
			GOUB~\cite{yue2024image}
			& 0.231 & 11.566 & \textcolor{blue}{\textbf{0.373}} & 3.928 & 0.360 & 5.853 & 0.458 & \textcolor{blue}{\textbf{4.104}} & 0.278 & 6.303\\
			ConvIR~\cite{cui2024revitalizing}
			& \textcolor{blue}{\textbf{0.236}} & \textcolor{blue}{\textbf{10.280}} & 0.370 & 3.723 & 0.364 & \textcolor{blue}{\textbf{5.813}} & 0.446 & 4.645 & \textcolor{blue}{\textbf{0.313}} & 6.465\\
			DeepSNNet~\cite{deng2025deepsn}
			& 0.231 & 11.446 & 0.348 & 3.896 & 0.367 & 5.882 & 0.436 & 4.662 & 0.301 & 6.525\\
			AWRaCLe~\cite{rajagopalan2025awracle}
			& 0.232 & 12.016 & 0.366 & 3.796 & 0.363 & 5.898 & 0.426 & 4.649 & 0.306 & 6.516\\
			MaIR~\cite{li2025mair}
			& 0.234 & 10.804 & 0.350 & \textcolor{blue}{\textbf{3.666}} & 0.363 & 5.890 & 0.245 & 22.446 & 0.284 & 6.590\\
			RDBM
			& \textcolor{red}{\textbf{0.238}} & \textcolor{red}{\textbf{9.559}}  & \textcolor{red}{\textbf{0.397}} & \textcolor{red}{\textbf{3.663}} & \textcolor{red}{\textbf{0.396}} & \textcolor{red}{\textbf{5.482}} & \textcolor{red}{\textbf{0.483}} & \textcolor{red}{\textbf{3.973}}  & \textcolor{red}{\textbf{0.343}} & \textcolor{red}{\textbf{5.671}}\\
			\bottomrule
	\end{tabular}}
\end{table}  

\subsection{Noise Maps Visualization }
\begin{figure}[t]
	\centering
	\includegraphics[width=0.99\linewidth]{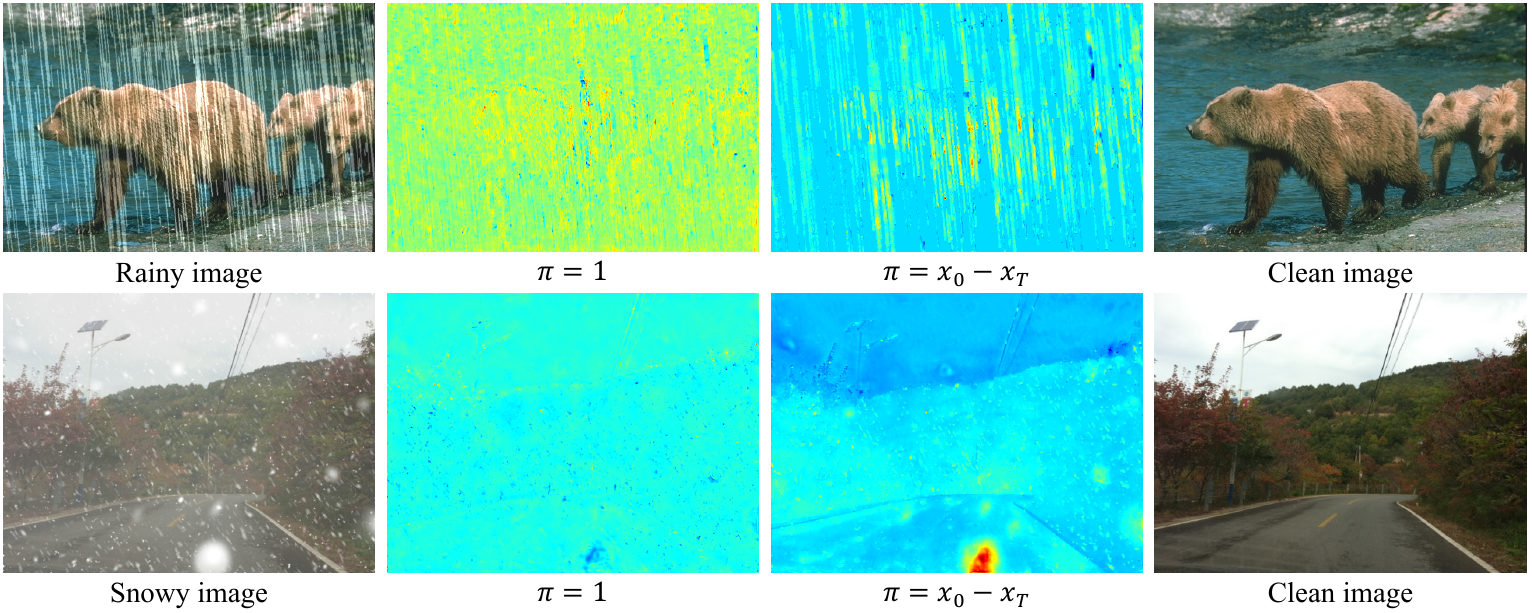}
	\vspace{-0.14in}
	\caption{Visualization of noise maps on different $\boldsymbol{\pi}$.}
	\label{fig:noise}
\end{figure}

To further elucidate the superiority of our method, we visualize the predicted noise maps generated at a random time point in the reverse process of bridge models under different settings of $\boldsymbol{\pi}$, as depicted in Fig.~\ref{fig:noise}. Obviously, naive diffusion bridge ($\boldsymbol{\pi}=1$) blindly conducts global noise removal for the reconstruction of missing details. In contrast, our method ($\boldsymbol{\pi}=\mathbf{x}_0 \! - \!\mathbf{x}_T $) performs adaptive restoration, as the noise maps are concentrated in degraded regions while remaining relatively smooth in non-degraded areas. In summary, our method can adaptively restore degradation in different regions, showcasing its high flexibility.

\subsection{Image Translation and Inpainting}
RDBM owns distinct advantages in mapping the data distribution to the prior distribution, thereby enabling extensive validation on similar computer vision tasks. To this end, we expand our experimental settings on image-to-image translation and image inpainting to fully demonstrate the potential of our method. The former aims to transform an input image from one domain to another while preserving certain essential semantic or structural features. The latter focuses on filling in missing or corrupted regions within an image. Specifically, we adopt the widely used edge2handlebags~\cite{isola2017image} dataset for image-to-image translation and apply the Celebrate-HQ dataset~\cite{karras2018progressive} with masks provided in \cite{liu2018image} for image inpainting. All these datasets are scaled to 256 × 256. We additionally employ Fréchet Inception Distance (FID)~\cite{heusel2017gans} for evaluation. Qualitative comparisons and quantitative results are presented in Fig.~\ref{fig:vision} and Tab.~\ref{tab:vision}, respectively. Clearly, our method achieves the best visual effects and the best metrics assessments.
 
\begin{table}[t]
	\caption{Quantitative results of image translation and inpainting.}\label{tab:vision}
	\vspace{-0.14in}
	\centering
	\renewcommand\arraystretch{1.0}
	\resizebox{1.0\columnwidth}{!}{
		\begin{tabular}{c|cccccccc}
			\toprule
			\multirow{3}{*}{\textbf{Method}}  &  \multicolumn{4}{c}{Image Translation~\cite{isola2017image}}  & \multicolumn{4}{c}{Image Inpainting~\cite{karras2018progressive}}  \\
			\cline{2-9}
			& \multicolumn{4}{c}{Edges→Handbags-256×256} & \multicolumn{4}{c}{Celebrate-HQ-256×256} \\ 
			&PSNR$\uparrow$ & SSIM$\uparrow$ & FID$\downarrow$ & LPIPS$\downarrow$ &PSNR$\uparrow$ & SSIM$\uparrow$ & FID$\downarrow$ & LPIPS$\downarrow$ \\
			\midrule
			DDPM~\cite{songscore} & 8.39 & 0.447 & 8.39 & 0.412  & 19.22 & 0.746 & 0.526 & 0.126 \\
			ReFlow~\cite{liuflow} & 10.46 & 0.442 & 5.76 & 0.374 & 23.53 &  0.822 & 0.307 & 0.149\\
			BBDM~\cite{li2023bbdm} & 14.75 & 0.635 & 7.46  & 0.248 & 20.36 & 0.653 & 0.386 & 0.169 \\
			I$^2$SB~\cite{liu20232} & 12.57 & 0.615 & 6.64 & 0.357 & 27.34  & 0.890  & 0.379  & \textcolor{blue}{\textbf{0.054}}\\
			RDDM~\cite{liu2024residual}  & 14.66 &  0.645  & \textcolor{blue}{\textbf{5.72}} & \textcolor{blue}{\textbf{0.256}}  & 23.94 & 0.852 & \textcolor{blue}{\textbf{0.167}} & 0.119 \\
			GOUB~\cite{yue2024image} & \textcolor{blue}{\textbf{16.58}} &  \textcolor{blue}{\textbf{0.700}}  & 8.76 & 0.288 & \textcolor{blue}{\textbf{31.56}} & \textcolor{blue}{\textbf{0.920}} & 0.321& 0.065 \\
			RDBM & \textcolor{red}{\textbf{19.26}} & \textcolor{red}{\textbf{0.738}} & \textcolor{red}{\textbf{5.38}} & \textcolor{red}{\textbf{0.224}} & \textcolor{red}{\textbf{37.88}}  & \textcolor{red}{\textbf{0.965}} & \textcolor{red}{\textbf{0.147}}  & \textcolor{red}{\textbf{0.031}}\\
			\bottomrule
	\end{tabular}}
\end{table} 

\section{Conclusion}
In this paper, we propose Residual Diffusion Bridge Model, termed as RDBM. Specifically, we theoretically reformulate the formulations of generalized diffusion bridge and derive the analytical formulas of its forward and reverse processes. Crucially, we leverage the residual from given distributions to dynamically modulate the probabilistic trajectories, thereby allowing the model to learn adaptive restoration of different regions with varying degradation levels. Furthermore, we unravel the fundamental mathematical essence of existing bridge models, and empirically verify the superiority of our models. Extensive experiments are conducted to demonstrate the state-of-the-art performance of our method across diverse tasks including image restoration, translation, and inpainting both qualitatively and quantitatively.

\section*{Acknowledgement}
This work was supported in part by the Fundamental and Interdisciplinary Disciplines Breakthrough Plan of the Ministry of Education of China (JYB2025XDXM101), the National Natural Science Foundation of China (62276192, 62225113, 624B2109), the Zhongguancun Academy Project (20240308), the New Generation Artificial Intelligence-National Science and Technology Major Project (2025ZD0123602), the National Key Laboratory of Multispectral Information Intelligent Processing Technology (61421132302), and the Key Technology Research Project of China National Petroleum Corporation (2025ZG82).

{
    \small
    \bibliographystyle{ieeenat_fullname}
    \bibliography{main}
}

\clearpage
\appendix 
\onecolumn
\begin{center}
	\Large
	\textbf{\thetitle}\\
	\vspace{0.5em}Supplementary Material \\
	\vspace{1.0em}
\end{center}


\setcounter{equation}{0}
\renewcommand{\theequation}{\thesection.\arabic{equation}} 
\renewcommand{\thesection}{A}
\section{Doob's $h$ transform}\label{sec:suppl_A}
\begin{theorem}\label{them:A}
	For a given SDE:
	\begin{equation} 
		\mathrm{d}\mathbf{x}_t=\mathbf{f}\left( \mathbf{x}_t,t \right) \mathrm{d}t+g_t \mathrm{d}\mathbf{w}_t,\qquad \mathbf{x}_0\sim p\left( \mathbf{x}_0 \right), 
	\end{equation}
	For a fixed $\mathbf{x}_T$, the evolution of conditional probability $p(\mathbf{x}_t\mid \mathbf{x}_T)$ follows:
	\begin{equation} 
		\mathrm{d}\mathbf{x}_t=\left[\mathbf{f}( \mathbf{x}_t,t) + g^2_t\mathbf{h}(\mathbf{x}_t,t,\mathbf{x}_T,T) \right]\mathrm{d}t + g_t \mathrm{d}\mathbf{w}_t,\qquad \mathbf{x}_0\sim p\left( \mathbf{x}_0 \mid\mathbf{x}_T \right),
	\end{equation}
	where $\mathbf{h}(\mathbf{x}_t,t,\mathbf{x}_T,T)=\nabla_{\mathbf{x}_t}\log p(\mathbf{x}_T\mid \mathbf{x}_t)$.
\end{theorem}
\textit{Proof}: In theory, $p(\mathbf x_t \mid \mathbf x_0)$ and $p(\mathbf x_T \mid \mathbf x_t)$ satisfy the Kolmogorov Forward Equation (KFE) and Kolmogorov Backward Equation (KBE), respectively~\cite{risken1989fokker}, as formulated below:
\begin{align}
	\frac{\partial}{\partial t} p(\mathbf x_t \mid \mathbf x_0) = -\nabla_{\mathbf x_t} \cdot \left[ \mathbf f(\mathbf x_t, t)p(\mathbf x_t \mid \mathbf x_0) \right] + \frac{1}{2}g^2_t \nabla_{\mathbf x_t} \cdot \nabla_{\mathbf x_t} p(\mathbf x_t \mid \mathbf x_0),\\
	-\frac{\partial}{\partial t} p(\mathbf x_T \mid \mathbf x_t) = \mathbf f(\mathbf x_t, t) \cdot \nabla_{\mathbf x_t} p(\mathbf x_T \mid \mathbf x_t) + \frac{1}{2}g^2_t \nabla_{\mathbf x_t} \cdot \nabla_{\mathbf x_t} p(\mathbf x_T\mid \mathbf x_t).
\end{align}
Using Bayes' rule, we have:
\begin{equation}
	\begin{aligned}
		p(\mathbf x_t \mid \mathbf x_0, \mathbf x_T) 
		&=\frac{p(\mathbf x_T \mid \mathbf x_t, \mathbf x_0)p(\mathbf x_t \mid \mathbf x_0)}{p(\mathbf x_T \mid \mathbf x_0)} \\
		&= \frac{p(\mathbf x_T \mid \mathbf x_t)p(\mathbf x_t \mid \mathbf x_0)}{p(\mathbf x_T \mid \mathbf x_0)}\\
	\end{aligned}
\end{equation}
Therefore, the derivative of conditional transition probability $p(\mathbf x_t \mid \mathbf x_0, \mathbf x_T)$ with time follows:
\begin{equation}\label{51}
	\begin{aligned}
		\frac{\partial}{\partial t} p(\mathbf x_t \mid \mathbf x_0, \mathbf x_T) 
		&= \frac{p(\mathbf x_t \mid \mathbf x_0)}{p(\mathbf x_T \mid \mathbf x_0)}\frac{\partial}{\partial t} p(\mathbf x_T \mid \mathbf x_t) + \frac{p(\mathbf x_T \mid \mathbf x_t)}{p(\mathbf x_T \mid \mathbf x_0)} \frac{\partial}{\partial t} p(\mathbf x_t \mid \mathbf x_0)\\
		&= \frac{p(\mathbf x_t \mid \mathbf x_0)}{p(\mathbf x_T \mid \mathbf x_0)}\left[-\mathbf f(\mathbf x_t, t) \cdot \nabla_{\mathbf x_t} p(\mathbf x_T \mid \mathbf x_t) - \frac{1}{2}g^2_t \nabla_{\mathbf x_t} \cdot \nabla_{\mathbf x_t} p(\mathbf x_T\mid \mathbf x_t) \right] 
		\\&\quad + \frac{p(\mathbf x_T \mid \mathbf x_t)}{p(\mathbf x_T \mid \mathbf x_0)} \left\{ -\nabla_{\mathbf x_t} \cdot \left[ \mathbf f(\mathbf x_t, t)p(\mathbf x_t \mid \mathbf x_0) \right] + \frac{1}{2}g^2_t \nabla_{\mathbf x_t} \cdot \nabla_{\mathbf x_t} p(\mathbf x_t \mid \mathbf x_0) \right\}\\
		&= - \left[\frac{p(\mathbf x_t \mid \mathbf x_0)}{p(\mathbf x_T \mid \mathbf x_0)}\mathbf f(\mathbf x_t, t) \cdot \nabla_{\mathbf x_t} p(\mathbf x_T \mid \mathbf x_t) + \frac{p(\mathbf x_T \mid \mathbf x_t)}{p(\mathbf x_T \mid \mathbf x_0)} \mathbf f(\mathbf x_t, t)\nabla_{\mathbf x_t} p(\mathbf x_t\mid \mathbf x_0) \right.\\
		&\quad \left. + \frac{p(\mathbf x_T \mid \mathbf x_t)}{p(\mathbf x_T \mid \mathbf x_0)}p(\mathbf x_t\mid \mathbf x_0) \nabla_{\mathbf x_t} \cdot \mathbf f(\mathbf x_t, t)\right]\\
		&\quad+ \frac{1}{2}g_t^2 \left[\frac{p(\mathbf x_T \mid \mathbf x_t)}{p(\mathbf x_T \mid \mathbf x_0)} \nabla_{\mathbf x_t} \cdot \nabla_{\mathbf x_t} p(\mathbf x_t \mid \mathbf x_0)- \frac{p(\mathbf x_t \mid \mathbf x_0)}{p(\mathbf x_T \mid \mathbf x_0)}\nabla_{\mathbf x_t} \cdot \nabla_{\mathbf x_t} p(\mathbf x_T\mid \mathbf x_t) \right]\\
		&=-\left[\mathbf f(\mathbf x_t, t) \cdot \nabla_{\mathbf x_t} p(\mathbf x_t \mid \mathbf x_0, \mathbf x_T) + p(\mathbf x_t \mid \mathbf x_0, \mathbf x_T)\cdot \nabla_{\mathbf x_t} \mathbf f(\mathbf x_t, t) \right]\\
		&\quad+\frac{1}{2}g_t^2 \left[\frac{p(\mathbf x_T \mid \mathbf x_t)}{p(\mathbf x_T \mid \mathbf x_0)} \nabla_{\mathbf x_t} \cdot \nabla_{\mathbf x_t} p(\mathbf x_t \mid \mathbf x_0)- \frac{p(\mathbf x_t \mid \mathbf x_0)}{p(\mathbf x_T \mid \mathbf x_0)}\nabla_{\mathbf x_t} \cdot \nabla_{\mathbf x_t} p(\mathbf x_T\mid \mathbf x_t) \right]\\
		&= - \nabla_{\mathbf x_t} \cdot \left[\mathbf f(\mathbf x_t, t) p(\mathbf x_t \mid \mathbf x_0, \mathbf x_T) \right]\\
		&\quad+\frac{1}{2}g_t^2 \left[\frac{p(\mathbf x_T \mid \mathbf x_t)}{p(\mathbf x_T \mid \mathbf x_0)} \nabla_{\mathbf x_t} \cdot \nabla_{\mathbf x_t} p(\mathbf x_t \mid \mathbf x_0)- \frac{p(\mathbf x_t \mid \mathbf x_0)}{p(\mathbf x_T \mid \mathbf x_0)}\nabla_{\mathbf x_t} \cdot \nabla_{\mathbf x_t} p(\mathbf x_T\mid \mathbf x_t) \right]
	\end{aligned}
\end{equation}
For the second term, we have:
\begin{equation}
	\begin{aligned}
		&\frac{1}{2}g_t^2 \left[\frac{p(\mathbf x_T \mid \mathbf x_t)}{p(\mathbf x_T \mid \mathbf x_0)} \nabla_{\mathbf x_t} \cdot \nabla_{\mathbf x_t} p(\mathbf x_t \mid \mathbf x_0)
		- \frac{p(\mathbf x_t \mid \mathbf x_0)}{p(\mathbf x_T \mid \mathbf x_0)}\nabla_{\mathbf x_t} \cdot \nabla_{\mathbf x_t} p(\mathbf x_T\mid \mathbf x_t) \right]\\
		=&\frac{1}{2}g_t^2\left[\frac{p(\mathbf x_T \mid \mathbf x_t)}{p(\mathbf x_T \mid \mathbf x_0)} \nabla_{\mathbf x_t} \cdot \nabla_{\mathbf x_t} p(\mathbf x_t \mid \mathbf x_0)+\frac{1}{p(\mathbf x_T \mid \mathbf x_0)}\nabla_{\mathbf x_t} p(\mathbf x_T \mid \mathbf x_t)\cdot \nabla_{\mathbf x_t}\ p(\mathbf x_t \mid \mathbf x_0) \right.\\
		&\left. +\frac{1}{p(\mathbf x_T \mid \mathbf x_0)}\nabla_{\mathbf x_t} p(\mathbf x_T \mid \mathbf x_t)\cdot \nabla_{\mathbf x_t}\ p(\mathbf x_t \mid \mathbf x_0)+\frac{p(\mathbf x_t \mid \mathbf x_0)}{p(\mathbf x_T \mid \mathbf x_0)}\nabla_{\mathbf x_t} \cdot \nabla_{\mathbf x_t} p(\mathbf x_T\mid \mathbf x_t) \right]\\
		&-g_t^2\left[\frac{1}{p(\mathbf x_T \mid \mathbf x_0)}\nabla_{\mathbf x_t} p(\mathbf x_T \mid \mathbf x_t)\cdot \nabla_{\mathbf x_t}\ p(\mathbf x_t \mid \mathbf x_0) + \frac{p(\mathbf x_t \mid \mathbf x_0)}{p(\mathbf x_T \mid \mathbf x_0)}\nabla_{\mathbf x_t} \cdot \nabla_{\mathbf x_t} p(\mathbf x_T\mid \mathbf x_t) \right]\\
		=&\frac{1}{2}g_t^2\left[\frac{1}{p(\mathbf x_T \mid \mathbf x_0)}\nabla_{\mathbf x_t}\cdot \left[p(\mathbf x_T \mid \mathbf x_t)\nabla_{\mathbf x_t}p(\mathbf x_t \mid \mathbf x_0) \right] + \frac{1}{p(\mathbf x_T \mid \mathbf x_0)}\nabla_{\mathbf x_t}\cdot \left[p(\mathbf x_t \mid \mathbf x_0)\nabla_{\mathbf x_t}p(\mathbf x_T \mid \mathbf x_t) \right] \right]\\
		&-g_t^2 \frac{1}{p(\mathbf x_T \mid \mathbf x_0)}\nabla_{\mathbf x_t}\cdot \left[p(\mathbf x_t \mid \mathbf x_0)\nabla_{\mathbf x_t}p(\mathbf x_T \mid \mathbf x_t) \right]\\
		=&\frac{1}{2}g_t^2\left[\nabla_{\mathbf x_t}\cdot \left[p(\mathbf x_t \mid \mathbf x_0, \mathbf x_T)\nabla_{\mathbf x_t}\log p(\mathbf x_t \mid \mathbf x_0) \right] + \nabla_{\mathbf x_t}\cdot \left[p(\mathbf x_t \mid \mathbf x_0, \mathbf x_T)\nabla_{\mathbf x_t}\log p(\mathbf x_T \mid \mathbf x_t) \right] \right]\\
		&-g_t^2 \nabla_{\mathbf x_t}\cdot \left[p(\mathbf x_t \mid \mathbf x_0, \mathbf x_T)\nabla_{\mathbf x_t}\log p(\mathbf x_T \mid \mathbf x_t) \right]\\
		=&\frac{1}{2}g_t^2\left[\nabla_{\mathbf x_t}\cdot \left[p(\mathbf x_t \mid \mathbf x_0, \mathbf x_T)\nabla_{\mathbf x_t}\log p(\mathbf x_t \mid \mathbf x_0, \mathbf x_T) \right] \right]-g_t^2 \nabla_{\mathbf x_t}\cdot \left[p(\mathbf x_t \mid \mathbf x_0, \mathbf x_T)\nabla_{\mathbf x_t}\log p(\mathbf x_T \mid \mathbf x_t) \right]\\
		=&\frac{1}{2}g_t^2 \nabla_{\mathbf x_t} \cdot \nabla_{\mathbf x_t} p(\mathbf x_t \mid \mathbf x_0, \mathbf x_T) -g_t^2 \nabla_{\mathbf x_t}\cdot \left[p(\mathbf x_t \mid \mathbf x_0, \mathbf x_T)\nabla_{\mathbf x_t}\log p(\mathbf x_T \mid \mathbf x_t) \right]
	\end{aligned}
\end{equation}
Bring it back to \eqref{51}:
\begin{equation}
	\begin{aligned}
		\frac{\partial}{\partial t} p(\mathbf x_t \mid \mathbf x_0, \mathbf x_T) 
		&=- \nabla_{\mathbf x_t} \cdot \left[\mathbf f(\mathbf x_t, t) p(\mathbf x_t \mid \mathbf x_0, \mathbf x_T) \right] + \frac{1}{2}g_t^2 \nabla_{\mathbf x_t} \cdot \nabla_{\mathbf x_t} p(\mathbf x_t \mid \mathbf x_0, \mathbf x_T) \\
		&\quad -g_t^2 \nabla_{\mathbf x_t}\cdot \left[p(\mathbf x_t \mid \mathbf x_0, \mathbf x_T)\nabla_{\mathbf x_t}\log p(\mathbf x_T \mid \mathbf x_t) \right]\\
		&=- \nabla_{\mathbf x_t} \cdot \left[[\mathbf f(\mathbf x_t, t) + g_t^2 \nabla_{\mathbf x_t}\log p(\mathbf x_T \mid \mathbf x_t)]p(\mathbf x_t \mid \mathbf x_0, \mathbf x_T) \right] + \frac{1}{2}g_t^2 \nabla_{\mathbf x_t} \cdot \nabla_{\mathbf x_t} p(\mathbf x_t \mid \mathbf x_0, \mathbf x_T)
	\end{aligned}
\end{equation}
This is the definition of FP equation of conditional transition probability $p(\mathbf x_t \mid \mathbf x_0, \mathbf x_T)$, which represents the evolution that follows the SDE:
\begin{equation}
	\mathrm{d}\mathbf{x}_t=\left[\mathbf{f}( \mathbf{x}_t,t) + g^2_t\nabla_{\mathbf x_t} \log p(\mathbf x_T \mid \mathbf x_t) \right]\mathrm{d}t + g_t \mathrm{d}\mathbf{w}_t
\end{equation}
This concludes the proof of the \textbf{Theorem \ref{them:A}} in Sec.~\ref{sec:DBM}.

\setcounter{equation}{0}
\renewcommand{\theequation}{\thesection.\arabic{equation}} 
\renewcommand{\thesection}{B}
\section{Mean-Reverting Ornstein–Uhlenbeck Process}\label{sec:suppl_B}
\begin{theorem}
The SDE formulation of the Ornstein–Uhlenbeck process with its predefined coefficients $\theta_t,\sigma_t$ is:
\begin{equation}\label{eq1}
	d \mathbf{x}_t = \theta_t (\boldsymbol{\mu} - \mathbf{x}_t) dt + \sigma_t d w_t,
\end{equation}
where $\boldsymbol{\mu}$ represents the mean value that $\mathbf{x}_t$ will approximate at $t=T$. The solution of OU process can be calculated as:
\begin{equation}
	\!\mathbf{x}_t \!=\! \boldsymbol{\mu} \!+\! (x_0\!-\!\boldsymbol{\mu})e^{-\!\int_0^t \theta_s ds} \!+\! e^{-\int_0^t\! \theta_s \!ds}\!\int_0^t \!\sigma_s e^{\int_0^s \theta_u du}\! dw_s,
\end{equation}
\end{theorem}
\textit{Proof.} We define a surrogate differentiable function $\psi(\mathbf{x},t) = \mathbf{x}e^{\int_0^t \theta_z dz} = \mathbf{x}e^{\overline{\theta}_t}$ and expand it by $It\hat{o}$ formula:
\begin{equation}
	\begin{aligned}
		d\psi(\mathbf{x},t) &= \frac{\partial \psi }{\partial t}(\mathbf{x},t) dt + \frac{\partial \psi }{\partial \mathbf{x}}(\mathbf{x},t) d\mathbf{x} + \frac{1}{2} \frac{\partial^2 \psi }{\partial \mathbf{x}^2}(\mathbf{x},t)d\mathbf{x}^2\\
		&=\theta_t \mathbf{x} e^{\overline{\theta}_t} dt + e^{\overline{\theta}_t}(\theta_t(\boldsymbol{\mu} - \mathbf{x})dt + \sigma_t dw_t) \\
		&=\boldsymbol{\mu} \theta_t e^{\overline{\theta}_t} dt + \sigma_t e^{\overline{\theta}_t} dw_t
	\end{aligned}
\end{equation}
Then, we can solve $\mathbf{x}_t$ conditioned on $\mathbf{x}_s$ where $s<t$, as:
\begin{align}
	&\psi(\mathbf{x}_t,t)\! - \!\psi(\mathbf{x}_s,s) \! = \! \int_s^t \boldsymbol{\mu} \theta_z e^{\overline{\theta}_z} dz \!+ \! \int_s^t \sigma_z e^{\overline{\theta}_z} dw_z,\\
	&\mathbf{x}_te^{\overline{\theta}_t} - \mathbf{x}_se^{\overline{\theta}_s} = \boldsymbol{\mu}(e^{\overline{\theta}_t} - e^{\overline{\theta}_s}) + \int_s^t \sigma_z e^{\overline{\theta}_z} dw_z,\\
	&\mathbf{x}_t = \boldsymbol{\mu} + (\mathbf{x}_s - \boldsymbol{\mu})e^{-\overline{\theta}_{s:t}}+\int_s^t \sigma_z e^{-\overline{\theta}_{z:t}} dw_z.\label{Sec_B:eq11}
\end{align}
where ${-\overline{\theta}_{s:t}} = {-\int_s^t \theta_z dz}$, and thus we complete the proof. The expectation and variance of Eq.~(\ref{Sec_B:eq11}) can be rewritten:
\begin{equation}
	E[x_t] = \boldsymbol{\mu} + (x_s - \boldsymbol{\mu})e^{-\overline{\theta}_{s:t}},
\end{equation}
\begin{equation}
	Var[x_t] = \int_s^t \sigma_z^2 e^{-2\overline{\theta}_{z:t}} dz,
\end{equation}
This concludes the derivations in Sec.~\ref{sec:OUP}.

\setcounter{equation}{0}
\renewcommand{\theequation}{\thesection.\arabic{equation}} 
\renewcommand{\thesection}{C}
\section{RDBM Formulation}\label{sec:suppl_C}
\textbf{Proposition~\ref{prop_1}}: Let $\mathbf{x}_t$ be a finite random variable governed by the generalized OU process, with terminal condition $\mathbf{x}_T = \boldsymbol{\mu}$. The evolution of its marginal distribution $p(\mathbf{x}_t \mid \mathbf{x}_T)$ satisfies the following SDE under a fixed drift-to-diffusion coefficient ratio $\lambda$:
\begin{equation}\label{sec:suppl_C:eq1}
	d\mathbf{x}_t = \theta_t\coth(\overline{\theta}_{t:T})(\boldsymbol{\mu}-\mathbf{x}_t)dt + \sqrt{2\boldsymbol{\pi}^2\lambda \theta_t} dw_t,
\end{equation}
where $\overline{\theta}_{t:T} = \int_t^T \theta_z dz$ and $\boldsymbol{\pi}\in \mathbb{R}$ is the predefined parameter.\\ 
\textit{Proof:} First, we define a generalized OU process with the properties of mean-reverting:
\begin{equation}\label{sec:suppl_C:eq2}
	d \mathbf{x}_t = \theta_t(\boldsymbol{\mu} - \mathbf{x}_t)dt + \boldsymbol{\pi}\sigma_t dw_t.
\end{equation}
In our formulation, $\boldsymbol{\pi} = \boldsymbol{\mu} - \mathbf{x}_0$ is considered as the residual of given distributions. When $\boldsymbol{\pi} = 1$ or $\boldsymbol{\pi} = 0$, Eq.~\eqref{sec:suppl_C:eq2} can degenerate to other bridge models, as discussed in Suppl.~\ref{sec:suppl_F}. Here, we solve this SDE step by step, akin to Suppl.~\ref{sec:suppl_B}. First, we define a surrogate differentiable function $\psi(\mathbf{x},t) = \mathbf{x}e^{\int_0^t \theta_z dz} = \mathbf{x}e^{\overline{\theta}_t}$ and expand it by $It\hat{o}$ formula:
\begin{equation}
	\begin{aligned}\label{sec:suppl_C:eq3}
		d\psi(\mathbf{x},t) &= \frac{\partial \psi }{\partial t}(\mathbf{x},t) dt + \frac{\partial \psi }{\partial \mathbf{x}}(\mathbf{x},t) d\mathbf{x} + \frac{1}{2} \frac{\partial^2 \psi }{\partial \mathbf{x}^2}(\mathbf{x},t)d\mathbf{x}^2\\
		&=\theta_t \mathbf{x} e^{\overline{\theta}_t} dt + e^{\overline{\theta}_t}(\theta_t(\boldsymbol{\mu} - \mathbf{x})dt + \boldsymbol{\pi}\sigma_t dw_t) \\
		&=\boldsymbol{\mu} \theta_t e^{\overline{\theta}_t} dt + \boldsymbol{\pi}\sigma_t e^{\overline{\theta}_t} dw_t
	\end{aligned}
\end{equation}
Then, we can solve $\mathbf{x}_t$ conditioned on $\mathbf{x}_s$ where $s<t$, as:
\begin{align}\label{sec:suppl_C:eq4}
	&\psi(\mathbf{x}_t,t)\! - \!\psi(\mathbf{x}_s,s) \! = \! \int_s^t \boldsymbol{\mu} \theta_z e^{\overline{\theta}_z} dz \!+ \! \int_s^t \boldsymbol{\pi}\sigma_z e^{\overline{\theta}_z} dw_z,\\
	&\mathbf{x}_te^{\overline{\theta}_t} - \mathbf{x}_se^{\overline{\theta}_s} = \boldsymbol{\mu}(e^{\overline{\theta}_t} - e^{\overline{\theta}_s}) + \int_s^t \boldsymbol{\pi}\sigma_z e^{\overline{\theta}_z} dw_z,\\
	&\mathbf{x}_t = \boldsymbol{\mu} + (\mathbf{x}_s - \boldsymbol{\mu})e^{-\overline{\theta}_{s:t}}+\int_s^t \boldsymbol{\pi}\sigma_z e^{-\overline{\theta}_{z:t}} dw_z.\label{eq11}
\end{align}
where ${-\overline{\theta}_{s:t}} = {-\int_s^t \theta_z dz}$. The expectation and variance of Eq.~(\ref{eq11}) can be written as below:
\begin{equation}\label{sec:suppl_C:eq5}
	E[\mathbf{x}_t] = \boldsymbol{\mu} + (\mathbf{x}_s - \boldsymbol{\mu})e^{-\overline{\theta}_{s:t}},
\end{equation}
\begin{equation}\label{sec:suppl_C:eq6}
	Var[\mathbf{x}_t] = \boldsymbol{\pi}^2 \int_s^t \sigma_z^2 e^{-2\overline{\theta}_{z:t}} dz,
\end{equation}
To derive the analytical form of Eq.~\eqref{sec:suppl_C:eq6}, we assume that $\lambda = \frac{\sigma_t^2}{2\theta_t}$ is pre-defined stationary variance, and obtain:
\begin{equation}\label{sec:suppl_C:eq7}
	Var[\mathbf{x}_t] \!=\! \lambda \boldsymbol{\pi}^2 \int_s^t 2\theta_z e^{-2\overline{\theta}_{z:t}} dz = \lambda \boldsymbol{\pi}^2 (1 - e^{-2\overline{\theta}_{s:t}}),
\end{equation}
We can conclude that:
\begin{align}\label{sec:suppl_C:eq8}
	&p(\mathbf{x}_t\vert \mathbf{x}_s) \!\sim \mathcal{N}(\boldsymbol{\mu} \!+\! (\mathbf{x}_s\! -\! \boldsymbol{\mu})e^{-\overline{\theta}_{s:t}},\lambda \boldsymbol{\pi}^2 (1 \!-\! e^{-2\overline{\theta}_{s:t}})),
\end{align}
To ensure that the final state of time point $t=T$ conforms to the distribution of low-quality image $\mathbf{x}_T = \boldsymbol{\mu} \sim p_{LQ}(\boldsymbol{x})$, we leverage the Doob's $h$ transform by modifying the forward SDE from Eq.~\eqref{sec:suppl_C:eq9} to Eq.~\eqref{sec:suppl_C:eq10}:
\begin{align}
	&d\mathbf{x}_t \! = \! \mathbf{f}(\mathbf{x},t)dt \!+\! g(t)dw_t, \label{sec:suppl_C:eq9}\\
	&d\mathbf{x}_t \! = \! [\mathbf{f}(\mathbf{x},t) \!+\! g(t)^2\nabla_{\mathbf{x}_t} \log p(\mathbf{x}_T\vert \mathbf{x}_t)] dt \!+\! g(t) dw_t \label{sec:suppl_C:eq10},
\end{align}
where term $\nabla_{\mathbf{x}_t} \log p(\mathbf{x}_T\vert x_t)$ can be calculated by setting $s = 0, t= T$ in Eq.~(\ref{sec:suppl_C:eq8}):
\begin{equation}
	\nabla_{\mathbf{x}_t} \log p(\mathbf{x}_T\vert \mathbf{x}_t) = (\boldsymbol{\mu} - \mathbf{x}_t)\frac{e^{-2\overline{\theta}_{t:T}}}{\lambda \boldsymbol{\pi}^2 (1 \!-\! e^{-2\overline{\theta}_{t:T}})}.
\end{equation}
The mean-reverting OU process turns into a mean-arriving process, which can be formulated as:
\begin{equation}\label{eq19}
	\begin{aligned}
		&d\mathbf{x}_t = (\theta_t + \frac{\sigma^2_te^{-2\overline{\theta}_{t:T}}}{\lambda (1 - e^{-2\overline{\theta}_{t:T}})})(\boldsymbol{\mu}-\mathbf{x}_t)dt + \boldsymbol{\pi}\sigma_t dw_t,\\
		&=\theta_t(1 + \frac{2e^{-2\overline{\theta}_{t:T}}}{1 - e^{-2\overline{\theta}_{t:T}}})(\boldsymbol{\mu}-\mathbf{x}_t)dt+\boldsymbol{\pi}\sigma_tdw_t\\
		&=\theta_t(\frac{1 + e^{-2\overline{\theta}_{t:T}}}{1 - e^{-2\overline{\theta}_{t:T}}})(\boldsymbol{\mu}-\mathbf{x}_t)dt + \boldsymbol{\pi}\sigma_t dw_t,\\
		&=\theta_t\coth(\overline{\theta}_{t:T})(\boldsymbol{\mu}-\mathbf{x}_t)dt + \boldsymbol{\pi}\sigma_t dw_t, (\sigma^2_t = 2\lambda \theta_t),\\
		&=\theta_t\coth(\overline{\theta}_{t:T})(\boldsymbol{\mu}-\mathbf{x}_t)dt + \sqrt{2\boldsymbol{\pi}^2\lambda \theta_t} dw_t,
	\end{aligned}
\end{equation}
Eq.~(\ref{eq19}) can be converted into an analytical formula as follows. First, we substitute $y_t = \mathbf{x}_t - \boldsymbol{\mu}$, then the SDE of $y_t$ becomes:
\begin{equation}
	dy_t = -\theta_t\coth(\overline{\theta}_{t:T})y_tdt + \sqrt{2\boldsymbol{\pi}^2\lambda \theta_t} dw_t,
\end{equation}
Second, we introduce $\Psi_t = \exp(\int_0^t \theta_s\coth(\overline{\theta}_{s:T})ds)$ as the integrating factor and expand $\Psi_ty_t$ by It$\hat{o}$ formula:
\begin{equation}
	d(\Psi_ty_t) = \Psi_tdy_t + y_t d\Psi_t + d\Psi_tdy_t,
\end{equation}
Since $\Psi$ is a deterministic function, it satisfies $d\Psi = \Psi\theta_t\coth(\overline{\theta}_{t:T})$. $d\Psi dy_t$ produces $(dt)^2, dtdw_t$, which are the higher order infinitesimal of $dt$ and can be omitted. Thus, we obtain:
\begin{equation}\label{eq22}
	\begin{aligned}
		d(\Psi_ty_t) & =\! \Psi_t(-\theta_t\coth(\overline{\theta}_{t:T})y_tdt \!+\!  \sqrt{2\boldsymbol{\pi}^2\lambda \theta_t} dw_t) + \Psi_t \theta_t\coth(\overline{\theta}_{t:T})y_tdt= \Psi_t \sqrt{2\boldsymbol{\pi}^2\lambda \theta_t} dw_t,
	\end{aligned}
\end{equation}
Furthermore, we integrate both sides of Eq.~(\ref{eq22}):
\begin{equation}
	\Psi_t y_t = y_0 + \int_0^t\Psi_s\sqrt{2\boldsymbol{\pi}^2\lambda\theta_s}dw_s.
\end{equation}
Consequently, we have:
\begin{equation}\label{eq24}
	\begin{aligned}
		\mathbf{x}_t = \boldsymbol{\mu} + (\mathbf{x}_0-\boldsymbol{\mu})\Psi_t^{e^{-\int_0^t \theta_s\coth(\overline{\theta}_{s:T})ds}} + \int_0^t \sqrt{2\boldsymbol{\pi}^2\lambda\theta_s} e^{-\int_s^t \theta_z\coth(\overline{\theta}_{z:T})dz} dw_s.
	\end{aligned}
\end{equation}
We next analyze the analytical formulation of $\Psi_t$. Considering the internal integral $\int_0^t\theta_s \coth(\overline{\theta}_{s:T})ds$ at first, we set $u=\overline{\theta}_{s:T}$ satisfying $du = -\theta_sds$:
\begin{equation}
	\begin{aligned}
		&\int_0^t\theta_s \coth(\overline{\theta}_{s:T})ds = -\int_{\overline{\theta}_{0:T}}^{\overline{\theta}_{t:T}} \coth(u) du = -\ln \vert \sinh(u) \vert \bigg\vert_{\overline{\theta}_{0:T}}^{\overline{\theta}_{t:T}} = \ln \vert \frac{\sinh(\overline{\theta}_{0:T})}{\sinh(\overline{\theta}_{t:T})} \vert,
	\end{aligned}
\end{equation}
Therefore, the analytical expression of $\Psi_t$ is:
\begin{equation}
	\Psi_{t} = \frac{\sinh(\overline{\theta}_{0:T})}{\sinh(\overline{\theta}_{t:T})}.
\end{equation}
Finally, we can compute the closed-form of $\mathbf{x}_t$ in Eq.~(\ref{eq24}):
\begin{equation}\label{eq27}
	\mathbf{x}_t = \boldsymbol{\mu} + (\mathbf{x}_0 - \boldsymbol{\mu})\frac{\sinh(\overline{\theta}_{t:T})}{\sinh(\overline{\theta}_{0:T})}+\int_0^t \sqrt{2\boldsymbol{\pi}^2\lambda\theta_s} \frac{\sinh(\overline{\theta}_{t:T})}{\sinh(\overline{\theta}_{s:T})} dw_s.
\end{equation}
Eq.~(\ref{eq27}) preserves the properties of diffusion bridge models, whose initial state $\mathbf{x}_0$ and final state $\mathbf{x}_T$ are determined. The formulation of variance can be further simplified as follows:
\begin{equation}
	\begin{aligned}
		Var[x_t] = \int_0^t 2\boldsymbol{\pi}^2\lambda\theta_s (\frac{\sinh(\overline{\theta}_{t:T})}{\sinh(\overline{\theta}_{s:T})})^2 ds &= 2\boldsymbol{\pi}^2\lambda\sinh^2(\overline{\theta}_{t:T})\int_{\overline{\theta}_{0:T}}^{\overline{\theta}_{t:T}}-\frac{du}{\sinh^2(u)}= 2\boldsymbol{\pi}^2\lambda\sinh^2(\overline{\theta}_{t:T}) \coth(u)\bigg\vert_{\overline{\theta}_{0:T}}^{\overline{\theta}_{t:T}}\\
		& = 2\boldsymbol{\pi}^2\lambda\sinh^2(\overline{\theta}_{t:T})(\coth(\overline{\theta}_{t:T}) - \coth(\overline{\theta}_{0:T}))\\
		& = 2\boldsymbol{\pi}^2\lambda\sinh^2(\overline{\theta}_{t:T})(\frac{\sinh(\overline{\theta}_{0:T} - \overline{\theta}_{t:T})}{\sinh(\overline{\theta}_{0:T})\sinh(\overline{\theta}_{t:T})})\\
		& =  2\boldsymbol{\pi}^2\lambda\frac{\sinh(\overline{\theta}_{0:t})\sinh(\overline{\theta}_{t:T})}{\sinh(\overline{\theta}_{0:T})}
	\end{aligned}
\end{equation}
This concludes the derivations in Sec.~\ref{sec:GFP}. The expectation and variance of Eq.~(\ref{eq27}) are summarized as follows:
\begin{align}
	E[\mathbf{x}_t] &= \boldsymbol{\mu} + (\mathbf{x}_0 - \boldsymbol{\mu})\frac{\sinh(\overline{\theta}_{t:T})}{\sinh(\overline{\theta}_{0:T})},\label{eq28}\\
	Var[\mathbf{x}_t] & = 2\boldsymbol{\pi}^2\lambda\frac{\sinh(\overline{\theta}_{0:t})\sinh(\overline{\theta}_{t:T})}{\sinh(\overline{\theta}_{0:T})}.\label{eq29}
\end{align}

\setcounter{equation}{0}
\renewcommand{\theequation}{\thesection.\arabic{equation}} 
\renewcommand{\thesection}{D}
\section{In-depth analysis of $\boldsymbol{\pi}$ selection}\label{sec:suppl_D}
\noindent\textbf{Rethinking the diffusion process}. Mainstream diffusion models perturb the entire image with Gaussian noise and then perform pixel-wise reconstruction, aiming to handle noise corruption and recover high-quality information in parallel. For global degradations (e.g., low-light, noise), this approach achieves favorable performance by leveraging the known distribution of noise to guide the restoration of missing details. However, for mask-based degradations (e.g., rain, snow), only the degraded regions require restoration, while unaffected areas remain nearly identical to high-quality images. This approach introduces additional task complexity, which not only enables recovery of degraded regions, but also simultaneously compromises the quality of intact areas through redundant reconstruction. Moreover, severely degraded regions (with limited preserved information) benefit from enhanced noise perturbation to facilitate reconstruction, while mildly degraded regions require noise suppression to retain valid information. Drawing from the above analyses, $\boldsymbol{\pi}$ should possess weighted masking properties, effectively equivalent to the image residual $\boldsymbol{\pi} = \mathbf{x}_T-\mathbf{x}_0$.

\noindent\textbf{Power analysis}. Eq.~(\ref{eq27}) reveals that the diffusion process is determined by two terms given the final state $\mathbf{x}_T=\boldsymbol{\mu}$. The power ratio between residual component and noise component at pixel $i,j$ can be defined as residual-to-noise ratio $R(t,i,j)$:
\begin{equation}
	\begin{aligned}
		&R(t,i,j) \!=\! \frac{(\mathbf{x}_T(i,j)\!-\!\mathbf{x}_0(i,j))^2}{2\boldsymbol{\pi}^2(i,j) \lambda }\frac{(\frac{\sinh(\overline{\theta}_{t:T})}{\sinh(\overline{\theta}_{0:T})})^2}{ \frac{\sinh(\overline{\theta}_{0:t})\sinh(\overline{\theta}_{t:T})}{\sinh(\overline{\theta}_{0:T})}}
		= \frac{(\mathbf{x}_T(i,j)\!-\!\mathbf{x}_0(i,j))^2}{2\boldsymbol{\pi}^2(i,j) \lambda }\frac{\sinh(\overline{\theta}_{t:T})}{\sinh(\overline{\theta}_{0:t})\sinh(\overline{\theta}_{0:T})},
	\end{aligned}
\end{equation}
The first part is determined by predefined parameters $\boldsymbol{\pi},\lambda$ given initial and final states. The second part is entirely determined by the sequence of $\theta_t$ values, which approaches infinity at time 0 and converges to infinitesimal at time T. If $\boldsymbol{\pi}$ is a globally predefined parameter, when pixel residual $(\mathbf{x}_T(i,j)-\mathbf{x}_0(i,j))$ approaches zero, $R(t,i,j)\rightarrow \infty$. In this context, the high-quality regions are disrupted by noise and cannot be perfectly reconstructed due to the predicted error. Besides,  the refinement of low-quality regions with varying degradation degrees is dominated by their respective residual magnitudes. To make the $R(t,i,j)$ smooth, we leverage the setting of $\boldsymbol{\pi} = \mathbf{x}_T-\mathbf{x}_0$, and obtain:
\begin{equation}\label{eq45}
	R(t,i,j)  \!=\! R(t) = \frac{\sinh(\overline{\theta}_{t:T})}{\sinh(\overline{\theta}_{0:t})\sinh(\overline{\theta}_{0:T})}.
\end{equation}
Let us check the monotonic properties of $R(t)$ by its logarithm derivatives:
\begin{equation}
	A(t) = \sinh(\overline{\theta}_{t:T}),\qquad B(t) = \sinh(\overline{\theta}_{0:t}),\qquad C = \sinh(\overline{\theta}_{0:T})
\end{equation}
\begin{equation}
	A^\prime(t) = \cosh(\overline{\theta}_{t:T}) \cdot \frac{d}{dt}\overline{\theta}_{t:T} = -\theta_t \cosh(\overline{\theta}_{t:T}),B^\prime(t) = \cosh(\overline{\theta}_{0:t}) \cdot \frac{d}{dt}\overline{\theta}_{0:t} = \theta_t \cosh(\overline{\theta}_{t:T}),
\end{equation}
\begin{equation}
	\frac{d R(t)}{dt} \!=\! \frac{A^\prime(t)B(t)\! -\! A(t)B^\prime(t)}{B^2(t)C} = -\theta_t \frac{\cosh(\overline{\theta}_{t:T})\sinh(\overline{\theta}_{0:t}) + \sinh(\overline{\theta}_{t:T})\cosh(\overline{\theta}_{0:t})}{ \sinh^2(\overline{\theta}_{0:t}) \sinh(\overline{\theta}_{0:T})}  = - \frac{\theta_t}{\sinh^2(\overline{\theta}_{0:t})}.
\end{equation}
For the typical condition that $\theta_t \ge 0$, and $\sinh^2(\overline{\theta}_{t:T})\ge 0$, we can conclude that $R(t)$ is a monotonically decreasing function starting from $R(0)\rightarrow \infty$ to $R(T) \rightarrow 0$, as $\frac{d}{dt}R(t) \le 0$. It can be observed that if $\boldsymbol{\pi}=\mathbf{x}_T-\mathbf{x}_0$ is set, $R(t)$ decreases evenly for each pixel without being affected by image contents. Hence, we set $\boldsymbol{\pi}$ as residual component in Sec.~\ref{sec:GFP}.

\setcounter{equation}{0}
\renewcommand{\theequation}{\thesection.\arabic{equation}} 
\renewcommand{\thesection}{E}
\section{Process of Reverse Inference}\label{sec:suppl_E}
For simplicity, we use $\Theta_t$ and $\Sigma_t$ to represent the coefficients in Eq.~\eqref{eq28} and Eq.~\eqref{eq29}, respectively. We have:
\begin{equation}
	\Theta_t \equiv \frac{\sinh(\overline{\theta}_{t:T})}{\sinh(\overline{\theta}_{0:T})},\quad \Sigma_t \equiv 2\lambda\frac{\sinh(\overline{\theta}_{0:t})\sinh(\overline{\theta}_{t:T})}{\sinh(\overline{\theta}_{0:T})}
\end{equation}
\noindent\textbf{DDPM Reverse Process.} Leveraging the properties of Bayesian formula, we obtain:
\begin{align}
	p(\mathbf{x}_{t-1}\vert \mathbf{x}_t,\mathbf{x}_0,\mathbf{x}_T) &= \frac{p(\mathbf{x}_{t}\vert \mathbf{x}_{t-1},\mathbf{x}_0,\mathbf{x}_T)p(\mathbf{x}_{t-1}\vert \mathbf{x}_0,\mathbf{x}_T)}{p(\mathbf{x}_{t}\vert \mathbf{x}_0,\mathbf{x}_T)},\label{eq55}\\
	\mathbf{x}_{t-1} &= \boldsymbol{\mu} + (\mathbf{x}_0 - \boldsymbol{\mu})\Theta_{t-1} + \boldsymbol{\pi}\Sigma_{t-1}\epsilon_{t-1},\\
	\mathbf{x}_{t} &= \boldsymbol{\mu} + (\mathbf{x}_0 - \boldsymbol{\mu})\Theta_{t} + \boldsymbol{\pi}\Sigma_{t}\epsilon_{t}.
\end{align}
Eliminating the variable $\mathbf{x}_0$, we have:
\begin{align}
	\mathbf{x}_t &\!= \boldsymbol{\mu} + \Theta_{t} \frac{\mathbf{x}_{t-1} - \boldsymbol{\mu} - \Sigma_{t-1}\epsilon_{t-1}}{\Theta_{t-1}} + \boldsymbol{\pi}\Sigma_{t}\epsilon_{t} \\
	&=\boldsymbol{\mu} + \frac{\Theta_{t}}{\Theta_{t-1}}(\mathbf{x}_{t-1}-\boldsymbol{\mu}) + \boldsymbol{\pi}\sqrt{\Sigma_t^2 - \frac{\Theta_{t}^2}{\Theta_{t-1}^2}\Sigma_{t-1}^2}\epsilon
\end{align}
Back to Eq.~(\ref{eq55}), we have:
\begin{equation}
	\begin{aligned}
		&\log p(\mathbf{x}_{t-1}\vert \mathbf{x}_0,\mathbf{x}_t,\mathbf{x}_T) = \log p(\mathbf{x}_{t}\vert \mathbf{x}_{t-1}, \mathbf{x}_0,\mathbf{x}_T) + \log p(\mathbf{x}_{t-1}\vert \mathbf{x}_0,\mathbf{x}_T) - \log p(\mathbf{x}_{t}\vert \mathbf{x}_0,\mathbf{x}_T)\\
		&\propto -\frac{1}{2\boldsymbol{\pi}^2}\bigg[\frac{(\mathbf{x}_t - \boldsymbol{\mu} - \frac{\Theta_{t}}{\Theta_{t-1}}(\mathbf{x}_{t-1}-\boldsymbol{\mu}))^2}{\Sigma_t^2 - \frac{\Theta_{t}^2}{\Theta_{t-1}^2}\Sigma_{t-1}^2} + \frac{(\mathbf{x}_{t-1}-\boldsymbol{\mu} -(\mathbf{x}_0-\boldsymbol{\mu})\Theta_{t-1})^2}{\Sigma_{t-1}^2} - \frac{(\mathbf{x}_{t}-\boldsymbol{\mu} -(\mathbf{x}_0-\boldsymbol{\mu})\Theta_{t})^2}{\Sigma_{t}^2}\bigg]\\
		& = -\frac{1}{2\boldsymbol{\pi}^2}\!\bigg[
		\frac{(\mathbf{x}_{t-1}\!-\!\boldsymbol{\mu} \!-\! \frac{\Theta_{t-1}}{\Theta_{t}}(\mathbf{x}_t\!-\!\boldsymbol{\mu}))^2}{\frac{\Theta_{t-1}^2}{\Theta_{t}^2}\Sigma_t^2 \!-\! \Sigma_{t-1}^2} \!+\! \frac{(\mathbf{x}_{t-1}\!-\!\boldsymbol{\mu} \!-\!(\mathbf{x}_0\!-\!\boldsymbol{\mu})\Theta_{t-1})^2}{\Sigma_{t-1}^2} \!+\! C
		\bigg]\\
		& = -\frac{1}{2\boldsymbol{\pi}^2}\bigg[
		\frac{(\mathbf{x}_{t-1}^2\!-2(\boldsymbol{\mu} \!+\! \frac{\Theta_{t-1}}{\Theta_{t}}(\mathbf{x}_t\!-\!\boldsymbol{\mu}))\mathbf{x}_{t-1} + (\boldsymbol{\mu} \!+\! \frac{\Theta_{t-1}}{\Theta_{t}}(\mathbf{x}_t\!-\!\boldsymbol{\mu}))^2 }{\frac{\Theta_{t-1}^2}{\Theta_{t}^2}\Sigma_t^2 \!-\! \Sigma_{t-1}^2}
		\\& \phantom{**********************} + \frac{(\mathbf{x}_{t-1}^2\!-2(\boldsymbol{\mu} \!+\!(\mathbf{x}_0\!-\!\boldsymbol{\mu})\Theta_{t-1})\mathbf{x}_{t-1} + (\boldsymbol{\mu} \!+\!(\mathbf{x}_0\!-\!\boldsymbol{\mu})\Theta_{t-1})^2 }{\Sigma_{t-1}^2} + C
		\bigg]\\
	\end{aligned}
\end{equation}
Furthermore, all the terms not related to $\mathbf{x}_{t-1}$ are categorized as $C$:
\begin{equation}\label{eq60}
	\begin{aligned}
		\log p(\mathbf{x}_{t-1}\vert \mathbf{x}_0,\mathbf{x}_t,\mathbf{x}_T) = -& \frac{1}{2\boldsymbol{\pi}^2}\bigg[
		(\frac{1}{\frac{\Theta_{t-1}^2}{\Theta_{t}^2}\Sigma_t^2 \!-\! \Sigma_{t-1}^2} + \frac{1}{\Sigma_{t-1}^2}) \mathbf{x}_{t-1}^2 -2\big[\Sigma_{t-1}^2(\boldsymbol{\mu} \!+\! \frac{\Theta_{t-1}}{\Theta_{t}}(\mathbf{x}_t\!-\!\boldsymbol{\mu})) \\
		& + (\frac{\Theta_{t-1}^2}{\Theta_{t}^2}\Sigma_t^2 \!-\! \Sigma_{t-1}^2)(\boldsymbol{\mu} \!+\!(\mathbf{x}_0\!-\!\boldsymbol{\mu})\Theta_{t-1})\big] \mathbf{x}_{t-1} + C.
		\bigg]
	\end{aligned}
\end{equation}
We can reformulate the Eq.~(\ref{eq60}) in Gaussian distribution format:
\begin{equation}\label{eq61}
	\begin{aligned}
		&Var[\mathbf{x}_{t-1}] = \boldsymbol{\pi}^2 (\frac{1}{\frac{\Theta_{t-1}^2}{\Theta_{t}^2}\Sigma_t^2 \!-\! \Sigma_{t-1}^2} + \frac{1}{\Sigma_{t-1}^2})^{-1} = \boldsymbol{\pi}^2 \frac{\Sigma_{t-1}^2(\Theta_{t-1}^2\Sigma_t^2 -\Theta_{t}^2\Sigma_{t-1}^2)}{\Theta_{t-1}^2\Sigma_t^2}\\
	\end{aligned}
\end{equation}
\begin{equation}\label{eq62}
	\begin{aligned}
		E[\mathbf{x}_{t-1}] &= \big[\Sigma_{t-1}^2(\boldsymbol{\mu} \!+\! \frac{\Theta_{t-1}}{\Theta_{t}}(\mathbf{x}_t\!-\!\boldsymbol{\mu})) + (\frac{\Theta_{t-1}^2}{\Theta_{t}^2}\Sigma_t^2 \!-\! \Sigma_{t-1}^2)(\boldsymbol{\mu} \!+\!(\mathbf{x}_0\!-\!\boldsymbol{\mu})\Theta_{t-1})\big] \cdot \frac{\Sigma_{t-1}^2(\Theta_{t-1}^2\Sigma_t^2 -\Theta_{t}^2\Sigma_{t-1}^2)}{\Theta_{t-1}^2\Sigma_t^2}\\ 
		& = \frac{\Sigma_{t-1}^2(\Theta_{t-1}^2\Sigma_t^2 - \Theta_t^2\Sigma^2_{t-1}) }{\Theta_t^2}[\boldsymbol{\mu} + (\mathbf{x}_0 - \boldsymbol{\mu})\Theta_{t-1}] + \frac{\Sigma_{t-1}^4(\Theta_{t-1}^2\Sigma_t^2 - \Theta_t^2\Sigma_{t-1}^2)}{\Theta_{t-1}\Sigma_t\Theta_t}\boldsymbol{\pi}\epsilon_t
	\end{aligned}
\end{equation}
\noindent\textbf{DDIM Reverse Process.} A common forward process in our framework can be determined as follows:
\begin{equation}
	\mathbf{x}_{t-1} = \boldsymbol{\mu} + (\mathbf{x}_0 - \boldsymbol{\mu})\Theta_{t-1} + \boldsymbol{\pi}\Sigma_{t-1}\epsilon_{t-1},
\end{equation}
\begin{equation}
	\mathbf{x}_{t} = \boldsymbol{\mu} + (\mathbf{x}_0 - \boldsymbol{\mu})\Theta_{t} + \boldsymbol{\pi}\Sigma_{t}\epsilon_{t}.
\end{equation}
We assume the reverse process follows a Gaussian distribution:
\begin{equation}\label{eq65}
	\begin{aligned}
		& \mathbf{x}_{t-1} = \kappa_t \mathbf{x}_t + \eta_t \boldsymbol{\mu} + \gamma_{t} \mathbf{x}_0 + \dot{\sigma_t}\boldsymbol{\pi}\epsilon_{t}\\
		&= \kappa_t ( \boldsymbol{\mu} + (\mathbf{x}_0 - \boldsymbol{\mu})\Theta_{t} + \boldsymbol{\pi}\Sigma_{t}\epsilon_{t}) + \eta_t \boldsymbol{\mu} + \gamma_t \mathbf{x}_0 + \dot{\sigma_t}\boldsymbol{\pi}\epsilon_{t}\\
		&=(\kappa_t + \eta_t - \kappa_t\Theta_{t})\boldsymbol{\mu} + (\kappa_t\Theta_{t}+\gamma_{t})\mathbf{x}_0 + \boldsymbol{\pi}(\kappa_t^2\Sigma_{t}^2 + \dot{\sigma_t}^2)^{\frac{1}{2}}\epsilon_{t},
	\end{aligned}
\end{equation}
we have:
\begin{align}
	\kappa_t + \eta_t - \kappa_t\Theta_{t}&= 1 - \Theta_{t-1},\\
	\kappa_t\Theta_{t}+\gamma_{t} &= \Theta_{t-1},\\
	\Sigma_{t-1}^2 &= \kappa_t^2\Sigma_{t}^2 + \dot{\sigma_t}^2.
\end{align}
By setting $\dot{\sigma_t} = 0$:
\begin{equation}
	\begin{aligned}
		\kappa_t = {\frac{\Sigma_{t-1}}{\Sigma_{t}}}&,\gamma_t=\Theta_{t-1} - \Theta_{t} {\frac{\Sigma_{t-1}}{\Sigma_{t}}}\\
		\eta_t & = 1 - \Theta_{t-1} - (1- \Theta_{t}){\frac{\Sigma_{t-1}}{\Sigma_{t}}},
	\end{aligned}
\end{equation}
substituting into Eq.~(\ref{eq65}):
\begin{equation}
	\begin{aligned}
		\mathbf{x}_{t-1} & = {\frac{\Sigma_{t-1}}{\Sigma_{t}}} \mathbf{x}_t + (1 - \Theta_{t-1} - (1- \Theta_{t}){\frac{\Sigma_{t-1}}{\Sigma_{t}}})\boldsymbol{\mu}  + (\Theta_{t-1} - \Theta_{t} {\frac{\Sigma_{t-1}}{\Sigma_{t}}})\mathbf{x}_0\\
		& = {\frac{\Sigma_{t-1}}{\Sigma_{t}}} \mathbf{x}_t + (1 - \frac{\Sigma_{t-1}}{\Sigma_{t}} - (\Theta_{t-1} - \Theta_{t} \frac{\Sigma_{t-1}}{\Sigma_{t}}))\boldsymbol{\mu} + (\Theta_{t-1} - \Theta_{t} {\frac{\Sigma_{t-1}}{\Sigma_{t}}})\mathbf{x}_0\\
		& = {\frac{\Sigma_{t-1}}{\Sigma_{t}}} \mathbf{x}_t + (1 - \frac{\Sigma_{t-1}}{\Sigma_{t}})\boldsymbol{\mu} + (\Theta_{t-1} - \Theta_{t} {\frac{\Sigma_{t-1}}{\Sigma_{t}}})(\mathbf{x}_0 - \boldsymbol{\mu})\\
		& = {\frac{\Sigma_{t-1}}{\Sigma_{t}}} \mathbf{x}_t + (1 - \frac{\Sigma_{t-1}}{\Sigma_{t}})\boldsymbol{\mu} +(\Theta_{t-1} - \Theta_{t} {\frac{\Sigma_{t-1}}{\Sigma_{t}}})(\frac{\mathbf{x}_t - \boldsymbol{\mu} - \boldsymbol{\pi}\Sigma_{t}\epsilon_t}{\Theta_{t}})\\
		& = \boldsymbol{\mu} + \frac{\Theta_{t-1}}{\Theta_t}(\mathbf{x}_t - \boldsymbol{\mu}) - \boldsymbol{\pi}({\frac{\Theta_{t-1}}{\Theta_{t}}\Sigma_{t} - \Sigma_{t-1}})\epsilon_t
	\end{aligned}
\end{equation}
This concludes the derivations in Sec.~\ref{sec:RPTO}.
\setcounter{equation}{0}
\renewcommand{\theequation}{\thesection.\arabic{equation}} 
\renewcommand{\thesection}{F}
\section{Connections Among Existing Diffusion Bridge}\label{sec:suppl_F}
Suppose the high-quality image $\mathbf{x}$ is sampled from the data distribution $p_{HQ}(\boldsymbol{x})$ and the paired degraded image $\boldsymbol{\mu}$ is sampled from prior distribution $p_{LQ}(\boldsymbol{x})$. We redefine the generalized meaning-reverting process as:
\begin{equation}\label{sec:suppl_F:eq1}
	d \mathbf{x}_t = \theta_t (\boldsymbol{\mu} - \mathbf{x}_t) dt + \boldsymbol{\pi} \sigma_t d \omega_t,
\end{equation}
where $\boldsymbol{\pi}$ is a predefined value. By applying the Doob's $h$-transform, we can establish the bridge SDE that connects the paired distribution under a fixed drift-to-diffusion coefficient ratio $\lambda = \sigma_t^2/(2\theta_t)$: 
\begin{equation}\label{sec:suppl_F:eq2}
	d\mathbf{x}_t = \theta_t\coth(\overline{\theta}_{t:T})(\boldsymbol{\mu}-\mathbf{x}_t)dt + \sqrt{2\boldsymbol{\pi}^2\lambda \theta_t} d\omega_t,
\end{equation}

\subsection{Connections to Variance-Exploding (VE) and Variance-Preserving (VP) SDEs} SMLD~\cite{songscore} primarily introduces two mainstream diffusion formulations, namely VP and VE. For a given generalized OU process in Eq.~\eqref{sec:suppl_F:eq1}, there exists relationships:
\begin{equation}\label{sec:suppl_F:eq3}
	\begin{aligned}
		\lim\limits_{\theta_{t}\rightarrow 0}^{\boldsymbol{\pi}=1} \text{Eq.~\eqref{sec:suppl_F:eq1}}  &= \lim\limits_{\theta_{t}\rightarrow 0}^{\boldsymbol{\pi}=1}\{d \mathbf{x}_t = \theta_t (\boldsymbol{\mu} - \mathbf{x}_t) dt + \boldsymbol{\pi} \sigma_t d \omega_t\}\\
		&= \lim\limits_{\theta_{t}\rightarrow 0}^{\boldsymbol{\pi}=1}\{d \mathbf{x}_t = \sigma_t d\omega_t \}\\
		&= \text{VE},
	\end{aligned}
\end{equation}
where $\sigma_t$ can be any noise schedule. Besides, we have:
\begin{equation}\label{sec:suppl_F:eq4}
	\begin{aligned}
		\lim\limits_{\boldsymbol{\mu}\rightarrow 0,\theta_t \rightarrow \sigma_t^2}^{\boldsymbol{\pi}=1} \text{Eq.~\eqref{sec:suppl_F:eq1}}  &= \lim\limits_{\boldsymbol{\mu}\rightarrow 0,\theta_t \rightarrow \sigma_t^2}^{\boldsymbol{\pi}=1}\{d \mathbf{x}_t = \theta_t (\boldsymbol{\mu} - \mathbf{x}_t) dt + \boldsymbol{\pi} \sigma_t d \omega_t\}\\
		&= \lim\limits_{\boldsymbol{\mu}\rightarrow 0,\theta_t \rightarrow \sigma_t^2}^{\boldsymbol{\pi}=1}\{d \mathbf{x}_t = \theta_t \boldsymbol{\mu} dt - \theta_t \mathbf{x}_t dt +  \sigma_t d\omega_t \}\\
		&= \lim\limits_{\boldsymbol{\mu}\rightarrow 0,\theta_t \rightarrow \sigma_t^2}^{\boldsymbol{\pi}=1}\{d \mathbf{x}_t = -\frac{1}{2} \sigma_t^2 \mathbf{x}_t dt +  \sigma_t d\omega_t \}\\
		&= \text{VP},
	\end{aligned}
\end{equation}
where we set the $\theta_t \rightarrow \sigma_t^2$, implying $\lambda = \frac{1}{2}$. On this basis, DDBM~\cite{zhoudenoising} further extends such diffusion configuration to bridge models, which are also special cases of our formulation under specific configurations:
\begin{equation}\label{sec:suppl_F:eq5}
	\begin{aligned}
		\lim\limits_{\theta_{t}\rightarrow 0,\sigma_t^2 \rightarrow C}^{\boldsymbol{\pi}=1} \text{Eq.~\eqref{sec:suppl_F:eq2}}  &= \lim\limits_{\theta_{t}\rightarrow 0,\sigma_t^2 \rightarrow C}^{\boldsymbol{\pi}=1}\{d\mathbf{x}_t = \theta_t\coth(\overline{\theta}_{t:T})(\boldsymbol{\mu}-\mathbf{x}_t)dt + \sqrt{2\pi^2\lambda \theta_t} d\omega_t\}\\
		&= \lim\limits_{\theta_{t}\rightarrow 0,\sigma_t^2 \rightarrow C}^{\boldsymbol{\pi}=1}\{d \mathbf{x}_t = \frac{\sigma_t^2}{\sigma_T^2 - \sigma_t^2} (\boldsymbol{\mu} - \mathbf{x}_t) dt + \sigma_t d\omega_t  \}\\
		&= \text{VE Bridge},
	\end{aligned}
\end{equation}
where $C$ denotes a constant and $\lambda = \frac{\sigma_t^2}{2\theta_t}\rightarrow \infty$. We utilize the following approximation:
\begin{equation}\label{sec:suppl_F:eq6}
	\lim\limits_{\theta_{t}\rightarrow 0}\theta_t\coth(\overline{\theta}_{t:T}) = \theta \coth({\theta}(T-t)) = \frac{1}{T-t}.
\end{equation}
VP bridge drives terminate state towards a zero-mean Gaussian distribution, which satisfies:
\begin{equation}\label{sec:suppl_F:eq7}
	\begin{aligned}
		\lim\limits_{\boldsymbol{\mu}\rightarrow 0,\theta_t \rightarrow \sigma_t^2}^{\boldsymbol{\pi}=1} \text{Eq.~\eqref{sec:suppl_F:eq2}}  &= \lim\limits_{\boldsymbol{\mu}\rightarrow 0,\theta_t \rightarrow \sigma_t^2}^{\boldsymbol{\pi}=1}\{d\mathbf{x}_t = \theta_t\coth(\overline{\theta}_{t:T})(\boldsymbol{\mu}-\mathbf{x}_t)dt + \sqrt{2\pi^2\lambda \theta_t} d\omega_t\}\\
		&= \lim\limits_{\boldsymbol{\mu}\rightarrow 0,\theta_t \rightarrow \sigma_t^2}^{\boldsymbol{\pi}=1}\{d\mathbf{x}_t = -\sigma_t^2\coth(\overline{\sigma}^2_{t:T})\mathbf{x}_tdt + \sigma_t d\omega_t\}\\
		&= \text{VP Bridge},
	\end{aligned}
\end{equation}
\subsection{Connections to Brownian Bridge SDEs} 
Brownian bridge is a fundamental architecture for diffusion model, which are widely adopted in BBDM~\cite{li2023bbdm}, I$^2$SB~\cite{liu20232}. By setting $\theta_t \rightarrow 0$ with condition $2\lambda\theta_t = 1 = \sigma_t^2$, we can derive the Brownian Bridge as formulated below:
\begin{equation}\label{sec:suppl_F:eq8}
	\begin{aligned}
		\lim\limits_{\theta_{t}\rightarrow 0,\sigma_t^2 \rightarrow 1}^{\boldsymbol{\pi}=1} \text{Eq.~\eqref{sec:suppl_F:eq2}}  &= \lim\limits_{\theta_{t}\rightarrow 0,\sigma_t^2 \rightarrow 1}^{\boldsymbol{\pi}=1}\{d\mathbf{x}_t = \theta_t\coth(\overline{\theta}_{t:T})(\boldsymbol{\mu}-\mathbf{x}_t)dt + \sqrt{2\pi^2\lambda \theta_t} d\omega_t\}\\
		&= \lim\limits_{\theta_{t}\rightarrow 0,\sigma_t^2 \rightarrow 1}^{\boldsymbol{\pi}=1}\{d \mathbf{x}_t =  \frac{\boldsymbol{\mu} - \mathbf{x}_t}{T-t}dt + d\omega_t  \}\\
		&= \text{Brownian Bridge},
	\end{aligned}
\end{equation}
where the corresponding expectation and variance are:
\begin{equation}
	\mathbf{x}_t = \boldsymbol{\mu} + (\mathbf{x}_0 - \boldsymbol{\mu})(1-\frac{t}{T}) + \int_0^t \frac{T-t}{T-s}dw_s,
\end{equation}
\begin{align}
	E[\mathbf{x}_t] &= \boldsymbol{\mu} + (\mathbf{x}_0 - \boldsymbol{\mu})(1-\frac{t}{T}),\\
	Var[\mathbf{x}_t] & = t(1-\frac{t}{T}),
\end{align}

\subsection{Connections to Flow Matching}
Flow-based generative models~\cite{lipman2023flow,liuflow} design a deterministic probability path that linearly interpolates between a prior and the data distribution, and then directly learn a time-dependent vector field whose integral trajectories realize this path. By discarding the stochastic noise ($\sigma_t = 0$) and adopting the Brownian bridge configuration, Eq.~\eqref{sec:suppl_F:eq2} can be transformed into:
\begin{equation}
	\begin{aligned}
		\lim\limits_{\theta_{t}\rightarrow 0}^{\boldsymbol{\pi}=0} \text{Eq.~\eqref{sec:suppl_F:eq2}}  &= \lim\limits_{\theta_{t}\rightarrow 0}^{\boldsymbol{\pi}=0}\{d\mathbf{x}_t = \theta_t\coth(\overline{\theta}_{t:T})(\boldsymbol{\mu}-\mathbf{x}_t)dt + \sqrt{2\pi^2\lambda \theta_t} d\omega_t\}\\
		&= \lim\limits_{\theta_{t}\rightarrow 0}^{\boldsymbol{\pi}=0}\{d \mathbf{x}_t =  \frac{\boldsymbol{\mu} - \mathbf{x}_t}{T-t}dt \}\\
		&= \text{Flow Matching},
	\end{aligned}
\end{equation}
whose trajectories satisfy:
\begin{equation}
	\mathbf{x}_t = \boldsymbol{\mu} + (\mathbf{x}_0 - \boldsymbol{\mu})(1-\frac{t}{T}).
\end{equation}
\subsection{Connections to OU Bridge SDEs}
Eq.~\eqref{sec:suppl_F:eq2} can be transformed into naive OU bridge~\cite{yue2024image} by setting $\pi = 1$ to recover global noise perturbation:
\begin{equation}
	\begin{aligned}
		\lim\limits_{\theta_{t},\lambda}^{\boldsymbol{\pi}=1} \text{Eq.~\eqref{sec:suppl_F:eq2}}  &= \lim\limits_{\theta_{t},\lambda}^{\boldsymbol{\pi}=1}\{d\mathbf{x}_t = \theta_t\coth(\overline{\theta}_{t:T})(\boldsymbol{\mu}-\mathbf{x}_t)dt + \sqrt{2\pi^2\lambda \theta_t} d\omega_t\}\\
		&= \lim\limits_{\theta_{t},\lambda}^{\boldsymbol{\pi}=1}\{d \mathbf{x}_t =  \frac{\boldsymbol{\mu} - \mathbf{x}_t}{T-t}dt \}\\
		&= \text{OU Bridge},
	\end{aligned}
\end{equation}
\subsection{Connections to Stochastic Interpolants}
Stochastic interpolants~\cite{albergo2023stochastic} define a unified framework for flows and diffusions, which can be expressed as:
\begin{equation}
	\mathbf{x}_t = I(t,\mathbf{x}_0,\mathbf{x}_T) + \gamma(t)z, t\in[0,T],
\end{equation}
whose boundary conditions are $I(0,\mathbf{x}_0,\mathbf{x}_T) \!=\! \mathbf{x}_0$ and $I(T,\mathbf{x}_0,\mathbf{x}_T) \!=\! \mathbf{x}_T$. Eq.~\eqref{sec:suppl_F:eq2} describes our probability path as:
\begin{align}
	E[\mathbf{x}_t] &= \boldsymbol{\mu} + (\mathbf{x}_0 - \boldsymbol{\mu})\frac{\sinh(\overline{\theta}_{t:T})}{\sinh(\overline{\theta}_{0:T})},\\
	Var[\mathbf{x}_t] & = 2\boldsymbol{\pi}^2\lambda\frac{\sinh(\overline{\theta}_{0:t})\sinh(\overline{\theta}_{t:T})}{\sinh(\overline{\theta}_{0:T})}.
\end{align}
Hence, the above process can be regarded as stochastic interpolants. The derivative of $I(t,\mathbf{x}_0,\mathbf{x}_T)$ to time $t$ is fixed as:
\begin{equation}
	\partial_t I(t,\mathbf{x}_0,\mathbf{x}_T) = \partial_t \frac{\sinh(\overline{\theta}_{t:T})}{\sinh(\overline{\theta}_{0:T})} (\mathbf{x}_0 - \boldsymbol{\mu}),
\end{equation}
and $\gamma(t)$ with boundary conditions $\gamma(0) = \gamma(T) = 0$ is:
\begin{equation}
	\gamma(t)^2 = 2\boldsymbol{\pi}^2\lambda\frac{\sinh(\overline{\theta}_{0:t})\sinh(\overline{\theta}_{t:T})}{\sinh(\overline{\theta}_{0:T})}.
\end{equation}
These relationships are summarized in Tab.~\ref{tab:analysis} in Sec.~\ref{sec:analysis}.

\setcounter{equation}{0}
\renewcommand{\theequation}{\thesection.\arabic{equation}} 
\renewcommand{\thesection}{G}
\section{Training Objective}\label{sec:suppl_G}
\begin{proposition}\label{props:3}
Let $\mathbf{x}_t$ be a finite random variable described by the given residual diffusion bridge in Eq.~\eqref{sec:suppl_F:eq2}. For a fixed final state $\mathbf{x}_T = \boldsymbol{\mu}$. the expectation of log-likelihood $\mathbb{E}_{p(\mathbf{x}_0)}[\log p_\theta(\mathbf{x}_0\vert \boldsymbol{\mu})]$ possesses an Evidence Lower Bound (ELBO):
\begin{equation}
	ELBO = \mathbb{E}_{p(\mathbf{x}_0)}\bigg[\mathbb{E}_{p(\mathbf{x}_1\vert \mathbf{x}_0,\boldsymbol{\mu})}[\log p_\theta(\mathbf{x}_0\vert \mathbf{x}_1,\mathbf{x}_T)] - \sum\limits_{t>1}\mathbb{E}_{p(\mathbf{x}_t\vert \mathbf{x}_0,\boldsymbol{\mu})}[D_{KL}(p(\mathbf{x}_{t-1}\vert \mathbf{x}_0,\mathbf{x}_t,\mathbf{x}_T))\|p_{\theta}(\mathbf{x}_{t-1}\vert \mathbf{x}_t,\mathbf{x}_T)]
	\bigg]
\end{equation}
Assuming $p_{\theta}(\mathbf{x}_{t-1}\vert \mathbf{x}_t,\mathbf{x}_T)$ follows a Gaussian distribution with a constant variance $\mathcal{N}(\boldsymbol{\mu}_{\theta,t-1},\sigma^2_{\theta,t-1}\boldmath{I})$, maximizing the ELBO is equivalent to minimizing:
\begin{equation}
	\mathcal{L} = \mathbb{E}_{t,\mathbf{x}_0,\mathbf{x}_t,\mathbf{x}_T}\bigg[\frac{1}{2\sigma^2_{\theta,t-1}}\|\boldsymbol{\mu}_{t-1} - \boldsymbol{\mu}_{\theta,t-1}\|^2
	\bigg]
\end{equation}
\end{proposition}
where $\mu_{t-1}$ is the expectation at time $t-1$ and $\mu_{\theta,t-1}$ is predicted by a neural network parameterized by $\theta$.\\
\textit{Proof.} For the conditional marginal likelihood of the data $\mathbf{x}_0$, we have
\begin{equation}\label{suppl_F:eq1}
	 p_\theta(\mathbf{x}_0\vert \boldsymbol{\mu}) = \int p_\theta(\mathbf{x}_{0:T}\vert \boldsymbol{\mu}) d \mathbf{x}_{1:T} = \int \frac{p_\theta(\mathbf{x}_{0:T}\vert \boldsymbol{\mu})}{p(\mathbf{x}_{1:T}\vert \mathbf{x}_0,\boldsymbol{\mu})}p(\mathbf{x}_{1:T}\vert \mathbf{x}_0,\boldsymbol{\mu}) d \mathbf{x}_{1:T}
\end{equation}
To maximize Eq.~\eqref{suppl_F:eq1}, we leverage the property of Jensen's inequality:
\begin{align}
	\log p_\theta(\mathbf{x}_0\vert \boldsymbol{\mu}) &\ge \mathbb{E}_{{p(\mathbf{x}_{1:T}\vert \mathbf{x}_0,\boldsymbol{\mu})}}\bigg[\log \frac{p_\theta(\mathbf{x}_{0:T}\vert \boldsymbol{\mu})}{p(\mathbf{x}_{1:T}\vert \mathbf{x}_0,\boldsymbol{\mu})}\bigg]=\mathbb{E}\bigg[\log p_\theta(\mathbf{x}_{T}\vert \boldsymbol{\mu} ) + \log \frac{p_\theta(\mathbf{x}_{0:T-1}\vert \boldsymbol{\mu})}{p(\mathbf{x}_{1:T}\vert \mathbf{x}_0,\boldsymbol{\mu})}\bigg]\\
	&= \mathbb{E}\bigg[\log p_\theta(\mathbf{x}_{T}\vert \boldsymbol{\mu} ) + \sum\limits_{t\ge 1}\log \frac{p_\theta(\mathbf{x}_{t-1}\vert \mathbf{x}_t,\boldsymbol{\mu})}{p(\mathbf{x}_t\vert \mathbf{x}_{t-1},\boldsymbol{\mu})}\bigg]\\
	&=\mathbb{E}\bigg[\log p_\theta(\mathbf{x}_{T}\vert \boldsymbol{\mu} ) + \sum\limits_{t > 1}\log \frac{p_\theta(\mathbf{x}_{t-1}\vert \mathbf{x}_t,\boldsymbol{\mu})}{p(\mathbf{x}_t\vert \mathbf{x}_{t-1},\mathbf{x}_0,\boldsymbol{\mu})} + \log \frac{p_\theta(\mathbf{x}_0\vert \mathbf{x}_1,\boldsymbol{\mu})}{p(\mathbf{x}_1\vert \mathbf{x}_0,\boldsymbol{\mu})}\bigg]\\
	&=\mathbb{E}\bigg[\log p_\theta(\mathbf{x}_{T}\vert \boldsymbol{\mu} ) + \sum\limits_{t > 1}\log \frac{p_\theta(\mathbf{x}_{t-1}\vert \mathbf{x}_t,\boldsymbol{\mu})}{p(\mathbf{x}_{t-1}\vert \mathbf{x}_{t},\mathbf{x}_0,\boldsymbol{\mu})}\cdot \frac{p(\mathbf{x}_{t-1}\vert \mathbf{x}_0,\boldsymbol{\mu})}{p(\mathbf{x}_t\vert \mathbf{x}_0,\boldsymbol{\mu})} + \log \frac{p_\theta(\mathbf{x}_0\vert \mathbf{x}_1,\boldsymbol{\mu})}{p(\mathbf{x}_1\vert \mathbf{x}_0,\boldsymbol{\mu})}\bigg]\\
	&=\mathbb{E}\bigg[\log p_\theta(\mathbf{x}_{T}\vert \boldsymbol{\mu} ) + \sum\limits_{t > 1}\log \frac{p_\theta(\mathbf{x}_{t-1}\vert \mathbf{x}_t,\boldsymbol{\mu})}{p(\mathbf{x}_{t-1}\vert \mathbf{x}_{t},\mathbf{x}_0,\boldsymbol{\mu})}+ \sum\limits_{t > 1}\log \frac{p(\mathbf{x}_{t-1}\vert \mathbf{x}_0,\boldsymbol{\mu})}{p(\mathbf{x}_t\vert \mathbf{x}_0,\boldsymbol{\mu})} + \log \frac{p_\theta(\mathbf{x}_0\vert \mathbf{x}_1,\boldsymbol{\mu})}{p(\mathbf{x}_1\vert \mathbf{x}_0,\boldsymbol{\mu})}\bigg]\\
	&=\mathbb{E}\bigg[\log p_\theta(\mathbf{x}_{T}\vert \boldsymbol{\mu} ) + \sum\limits_{t > 1}\log \frac{p_\theta(\mathbf{x}_{t-1}\vert \mathbf{x}_t,\boldsymbol{\mu})}{p(\mathbf{x}_{t-1}\vert \mathbf{x}_{t},\mathbf{x}_0,\boldsymbol{\mu})}+ \log \frac{p(\mathbf{x}_{1}\vert \mathbf{x}_0,\boldsymbol{\mu})}{p(\mathbf{x}_T\vert \mathbf{x}_0,\boldsymbol{\mu})} + \log \frac{p_\theta(\mathbf{x}_0\vert \mathbf{x}_1,\boldsymbol{\mu})}{p(\mathbf{x}_1\vert \mathbf{x}_0,\boldsymbol{\mu})}\bigg]\\
	&=\mathbb{E}\bigg[\log \frac{p_\theta(\mathbf{x}_{T}\vert \boldsymbol{\mu} )}{p(\mathbf{x}_T\vert \mathbf{x}_0,\boldsymbol{\mu})}\bigg] + \sum\limits_{t > 1}\mathbb{E}\bigg[\log \frac{p_\theta(\mathbf{x}_{t-1}\vert \mathbf{x}_t,\boldsymbol{\mu})}{p(\mathbf{x}_{t-1}\vert \mathbf{x}_{t},\mathbf{x}_0,\boldsymbol{\mu})}\bigg] + \mathbb{E}\bigg[\log {p_\theta(\mathbf{x}_0\vert \mathbf{x}_1,\boldsymbol{\mu})}\bigg]\\ 
	&= \mathbb{E}_{p(\mathbf{x}_1\vert \mathbf{x}_0,\boldsymbol{\mu})}\bigg[\log {p_\theta(\mathbf{x}_0\vert \mathbf{x}_1,\boldsymbol{\mu})}\bigg] - \sum\limits_{t>1}\mathbb{E}_{p(\mathbf{x}_t\vert \mathbf{x}_0,\boldsymbol{\mu})}\bigg[ D_{KL}(p(\mathbf{x}_{t-1}\vert \mathbf{x}_{t},\mathbf{x}_0,\boldsymbol{\mu})\|p_\theta(\mathbf{x}_{t-1}\vert \mathbf{x}_T,\boldsymbol{\mu}))\bigg].
\end{align}   
Accordingly, 
\begin{equation}
	\begin{aligned}
		&D_{KL}\left( p\left( \mathbf{\mathbf{x}}_{t-1}\mid \mathbf{\mathbf{x}}_0, \mathbf{\mathbf{x}}_t, \mathbf{\mathbf{x}}_T \right) ||p_{\theta}\left( \mathbf{\mathbf{x}}_{t-1}\mid \mathbf{\mathbf{x}}_t,\mathbf{\mathbf{x}}_T \right) \right)\\
		=&\mathbb{E}_{p\left( \mathbf{\mathbf{x}}_{t-1}\mid \mathbf{\mathbf{x}}_0, \mathbf{\mathbf{x}}_t, \mathbf{\mathbf{x}}_T \right)}\left[\log \frac{ \frac{1}{\sqrt{2\pi}\sigma_{t-1}}e^{-(\mathbf \mathbf{x}_{t-1}-\boldsymbol{\mu}_{t-1})^2/{2\sigma_{t-1}^2}} } {\frac{1}{\sqrt{2\pi}\sigma_{\theta,t-1}}e^{-(\mathbf \mathbf{x}_{t-1}-{\boldsymbol{\mu}}_{\theta,t-1})^2/{2\sigma_{\theta,t-1}^2}}} \right]\\ 
		=&\mathbb{E}_{p\left( \mathbf{\mathbf{x}}_{t-1}\mid \mathbf{\mathbf{x}}_0, \mathbf{\mathbf{x}}_t, \mathbf{\mathbf{x}}_T \right)}\left[\log\sigma_{\theta,t-1} - \log\sigma_{t-1} - (\mathbf \mathbf{x}_{t-1}-\boldsymbol{\mu}_{t-1})^2/{2\sigma^{2}_{t-1}} + (\mathbf \mathbf{x}_{t-1}-{\boldsymbol{\mu}}_{\theta,t-1})^2/{2\sigma_{\theta,t-1}^{2}} \right]\\ 
		=&\log\sigma_{\theta,t-1}-\log\sigma_{t-1}-\frac{1}{2} + \frac{\sigma_{t-1}^2}{2\sigma_{\theta,t-1}^2} + \frac{(\boldsymbol{\mu}_{t-1}-{\boldsymbol{\mu}}_{\theta,t-1})^2}{2\sigma_{\theta,t-1}^2}
	\end{aligned}
\end{equation}
Ignoring unlearnable constant, the training objective that involves minimizing the negative ELBO is :
\begin{equation}\label{eq222}
	\mathcal{L}= \mathbb{E}_{t,\mathbf{\mathbf{x}}_0,\mathbf{\mathbf{x}}_t,\mathbf{\mathbf{x}}_T}\left[\frac{1}{2\sigma_{\theta,t-1}^{2}}\|{\boldsymbol{\mu}}_{t-1}-{\boldsymbol{\mu}}_{\theta,t-1}\|^2\right].
\end{equation}
By substituting Eq.~\eqref{eq62} into Eq.~\eqref{eq222} yields the equivalent loss:
\begin{equation} 
	\mathcal{L}= \mathbb{E}_{t,\mathbf{\mathbf{x}}_0,\mathbf{\mathbf{x}}_t,\mathbf{\mathbf{x}}_T}\left[C_\theta\|{\boldsymbol{\pi}\epsilon_{t-1}}-{\boldsymbol{\pi}\epsilon_{\theta,t-1}}\|^2\right].
\end{equation}
Where $C_\theta$ are corresponding weights. This concludes the proof of the \textbf{Proposition \ref{props:3}} in Sec.~\ref{sec:RPTO}.
\setcounter{equation}{0}
\renewcommand{\theequation}{\thesection.\arabic{equation}} 
\renewcommand{\thesection}{H}
\section{More Experiments}\label{sec:suppl_H}
\subsection{Summary about the Datasets}\label{sec:suppl_H-1}
We evaluate the proposed method on five natural image restoration tasks, including deraining, low-light enhancement, desnowing, dehazing, and deblurring. We select the most widely used datasets for each task, as summarized in Tab.~\ref{sec:suppl_H_tab1}.

\renewcommand\thetable{S1}
\begin{table}[h]
	\caption{Summary of the image restoration datasets utilized in this paper.}
	\label{sec:suppl_H_tab1}
	\centering
		\begin{tabular}{ccccc}
			\toprule
			\multicolumn{1}{c}{{Task}}&\multicolumn{1}{c}{{Dataset}}   & Synthetic/Real & {Train samples }      & {Test samples }     \\
			\midrule
			\multirow{8}{*}{\textbf{\shortstack{Deraining}}}
			&DID~\cite{zhang2018density}          & Synthetic   &  -     & 1,200   \\
			&Rain13K~\cite{Kui_2020_CVPR}         & Synthetic   &  13,711 &   -   \\
			&Rain\_100~\cite{yang2017deep}       & Synthetic   &  -     &  200   \\
			& DeRaindrop~\cite{qian2018attentive} &  Real       &  861   &  307  \\
			& GT-Rain~\cite{ba2022gt-rain}        &  Real       & 26,125  &  2,100 \\
			& RealRain-1k~\cite{li2022toward}   &  Real       &  1792   &  448  \\
			\hline
			\multirow{5}{*}{\textbf{\shortstack{Low-light\\Enhancement}}}
			& LOL~\cite{wei2018deep}      &         Real          &      485           &        15 \\
			& MEF~\cite{7120119}      &         Real          &        -           &        17 \\
			& VE-LOL-L~\cite{ll_benchmark} &       Synthetic/Real       &    900/400        &        100/100 \\
			&NPE~\cite{wang2013naturalness} &         Real          &      -           &        8 \\
			& DICM~\cite{lee2013contrast} &         Real          &      -           &        64 \\
			\hline	
			\multirow{2}{*}{\textbf{Desnowing}}
			& CSD~\cite{chen2021all}  & Synthetic&  8,000        &       2,000    \\
			& Snow100K-Real~\cite{liu2018desnownet}  & Real & -         &          1,329 \\
			\hline
			\multirow{5}{*}{\textbf{Dehazing}}
			& SOTS~\cite{li2018benchmarking}        & Synthetic & -	&  500 	 	 \\
			& ITS\_v2~\cite{li2018benchmarking}     & Synthetic & 13,990	&  - 	 	 \\
			& D-HAZY~\cite{Ancuti_D-Hazy_ICIP2016}     & Synthetic & 1,178	&  294   	 	 \\
			& NH-HAZE~\cite{NH-Haze_2020}    & Real      & -  & 55   	 	 \\
			& Dense-Haze~\cite{ancuti2019dense}  & Real      & -  & 55     \\
			& NHRW~\cite{zhang2017fast} & Real & -  & 150 \\
			\hline
			\multirow{2}{*}{\textbf{Deblur}} 
			& GoPro~\cite{nah2017deep}   & Synthetic & 2,103    &  1,111  \\
			&RealBlur~\cite{rim_2020_ECCV} & Real      & 3,758    &  980	 \\
			\bottomrule
		\end{tabular}
\end{table}

\subsection{More Visual Comparisons on Image Restoration}\label{sec:suppl_H-2}

We show the visualization results of other degradation categories in Fig.~\ref{fig:haze}, Fig.~\ref{fig:blur}, Fig.~\ref{fig:snow}, and Fig.~\ref{fig:lowlight}, to further demonstrate our superiority. Evidently, our method generates more stable image samples with high fidelity than other universal image restoration methods. Benefiting from the adaptivity of residual bridge score matching, we achieve the outstanding reconstruction of the missing details and preserve undegraded regions well.

\renewcommand\thefigure{S1}
\begin{figure*}[t]
	\centering
	\includegraphics[width=0.99\linewidth]{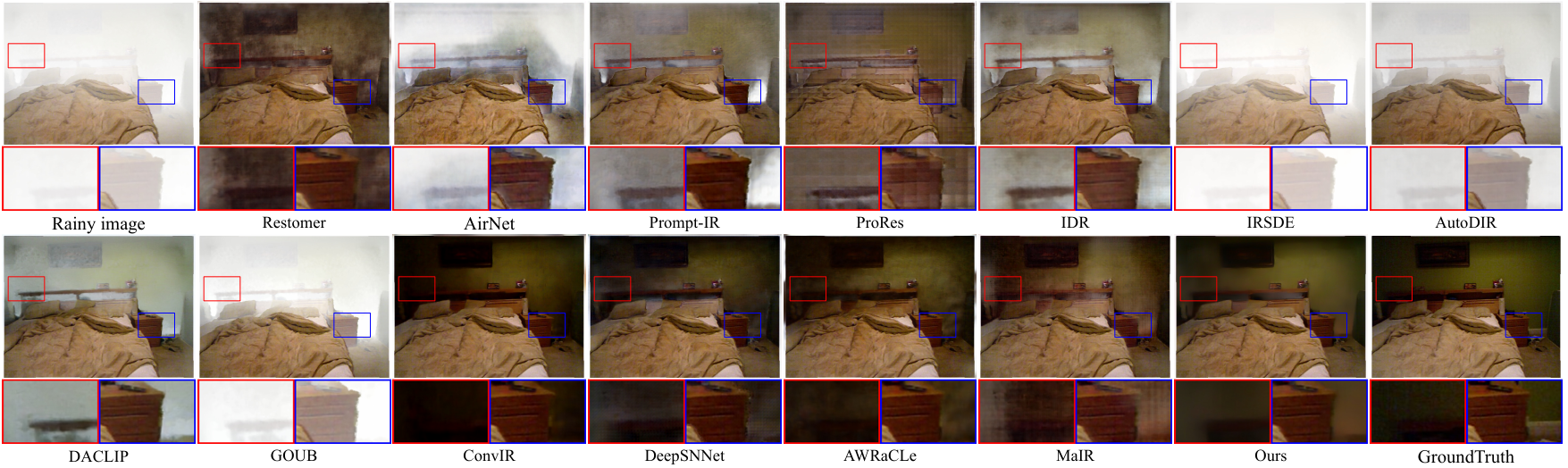}
	\vspace{-0.14in}
	\caption{Visualization comparison with state-of-the-art methods on dehazing. Zoom in for best view.}
	\label{fig:haze}
\end{figure*} 

\renewcommand\thefigure{S2}
\begin{figure*}[t]
	\centering
	\includegraphics[width=0.99\linewidth]{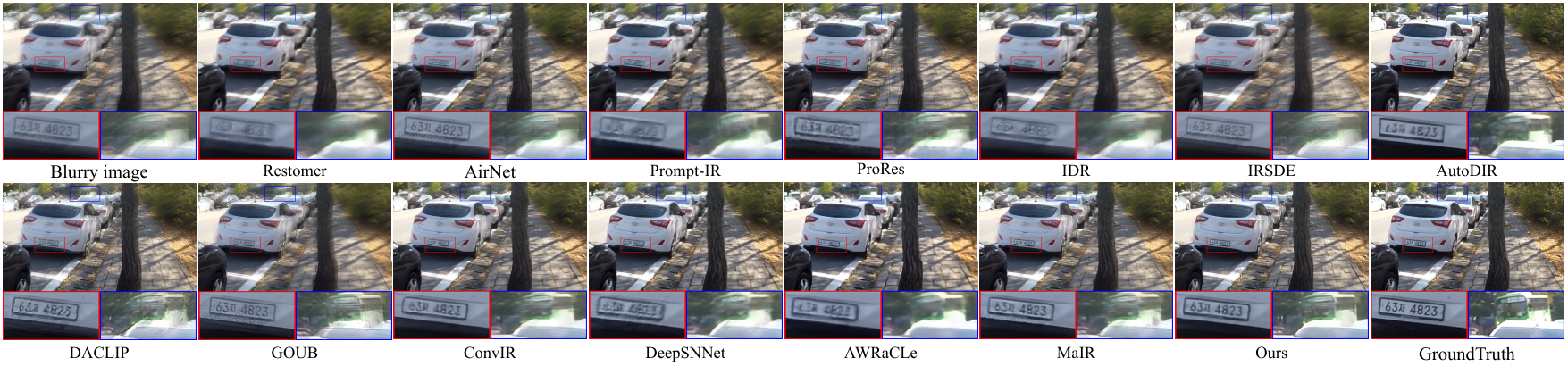}
	\vspace{-0.14in}
	\caption{Visualization comparison with state-of-the-art methods on deblurring. Zoom in for best view.}
	\label{fig:blur}
\end{figure*}

\renewcommand\thefigure{S3}
\begin{figure*}[t]
	\centering
	\includegraphics[width=0.99\linewidth]{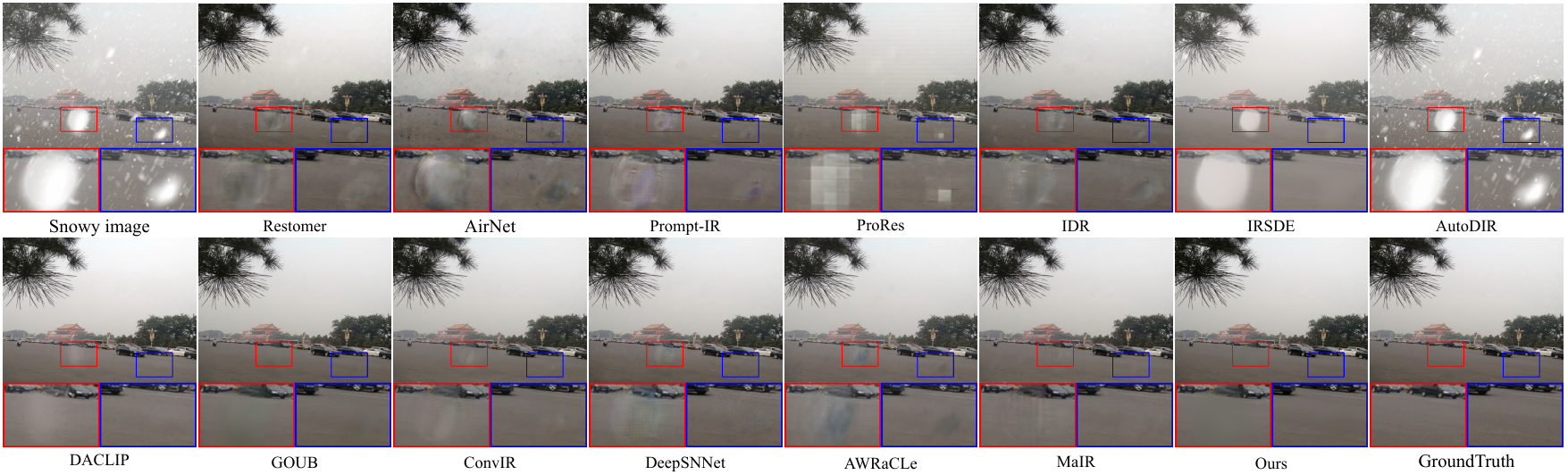}
	\vspace{-0.14in}
	\caption{Visualization comparison with state-of-the-art methods on desnowing. Zoom in for best view.}
	\label{fig:snow}
\end{figure*}

\renewcommand\thefigure{S4}
\begin{figure*}[t]
	\centering
	\includegraphics[width=0.99\linewidth]{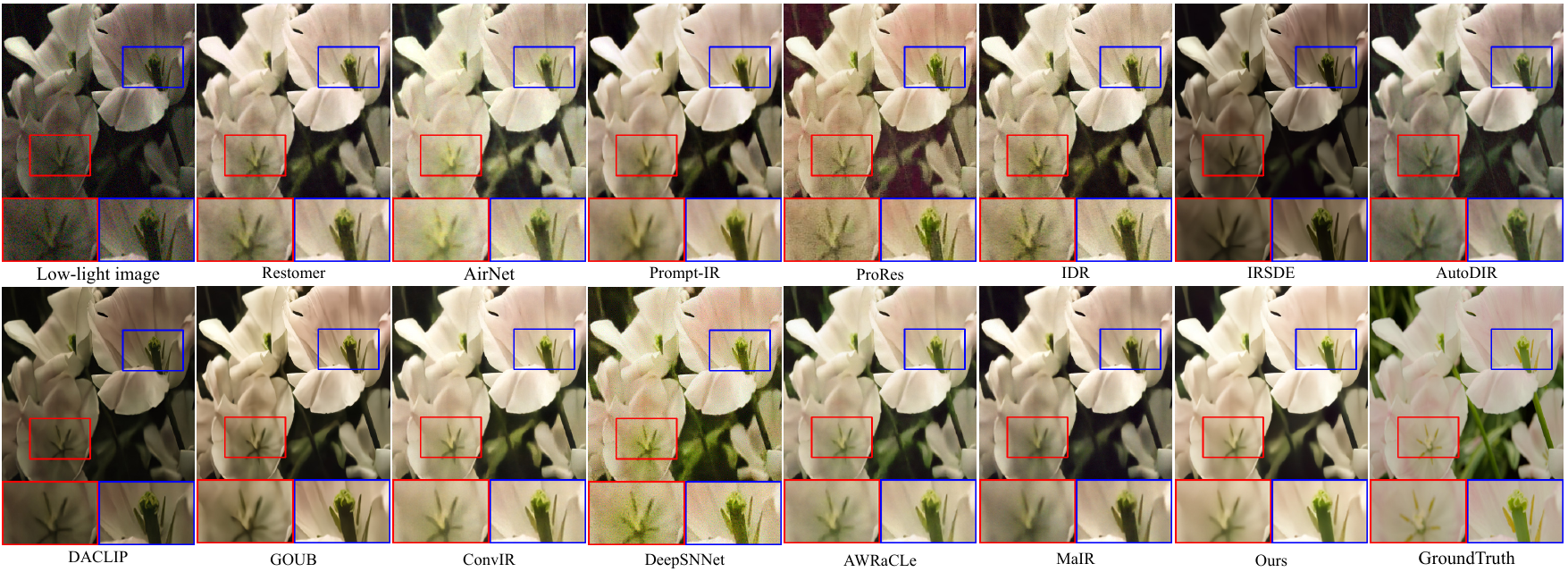}
	\vspace{-0.14in}
	\caption{Visualization comparison with state-of-the-art methods on low-light enhancement. Zoom in for best view.}
	\label{fig:lowlight}
\end{figure*} 

\renewcommand\thefigure{S5}
\begin{figure*}[t]
	\centering
	\includegraphics[width=0.99\linewidth]{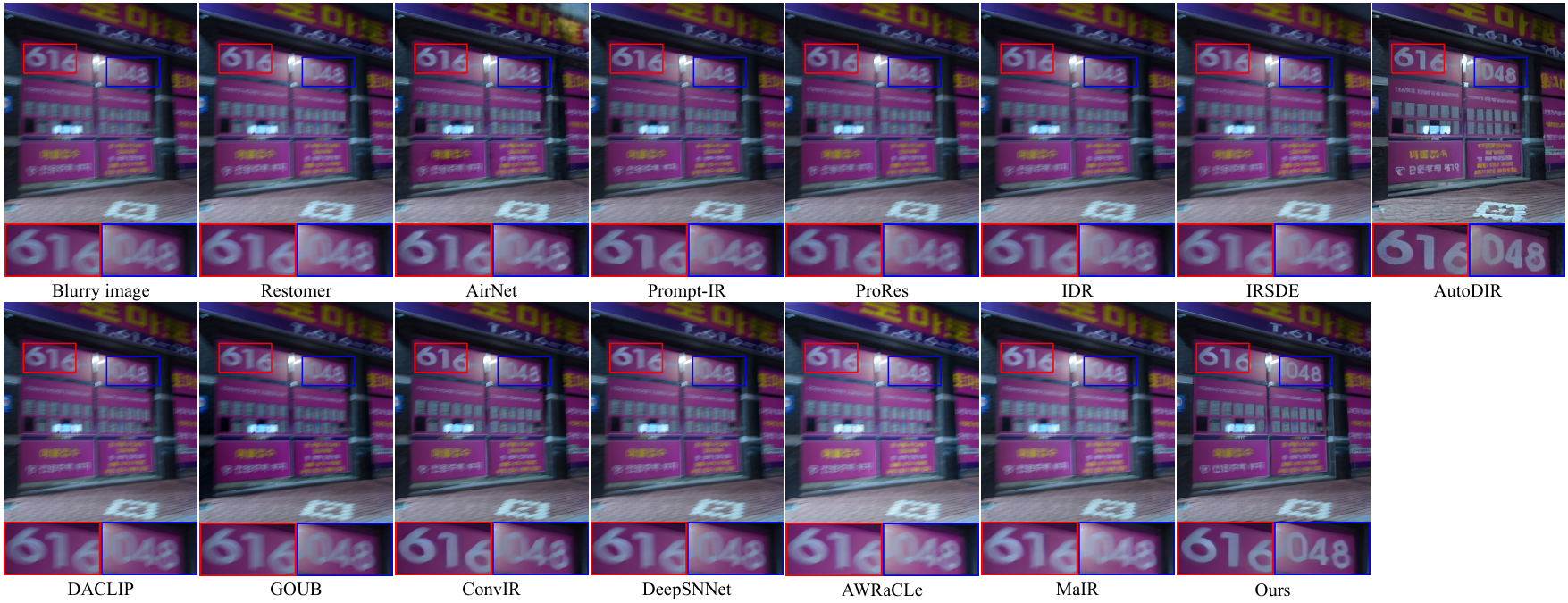}
	\vspace{-0.14in}
	\caption{Visualization comparison of deblurring task in real-world scenarios. Zoom in for best view.}
	\label{fig:real_blur}
\end{figure*} 

\renewcommand\thefigure{S6}
\begin{figure*}[t]
	\centering
	\includegraphics[width=0.99\linewidth]{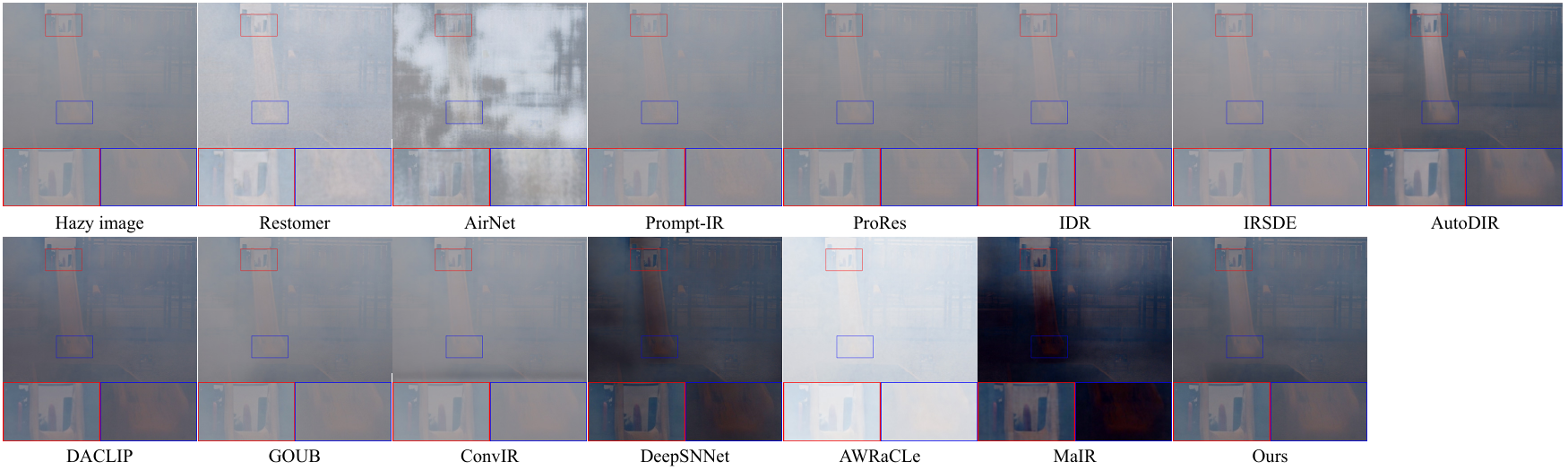}
	\vspace{-0.14in}
	\caption{Visualization comparison of dehazing task in real-world scenarios. Zoom in for best view.}
	\label{fig:real_haze}
\end{figure*} 

\renewcommand\thefigure{S7}
\begin{figure*}[t]
	\centering
	\includegraphics[width=0.99\linewidth]{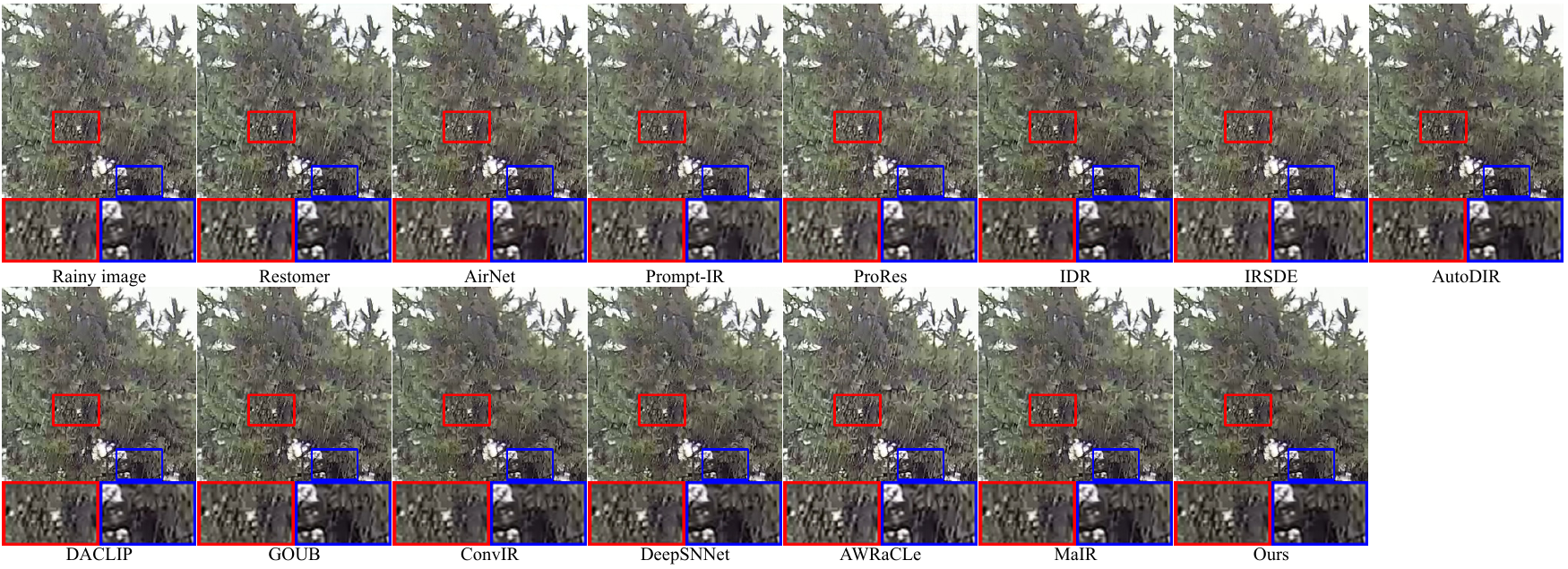}
	\vspace{-0.14in}
	\caption{Visualization comparison of deraining task in real-world scenarios. Zoom in for best view.}
	\label{fig:real_rain}
\end{figure*} 

\renewcommand\thefigure{S8}
\begin{figure*}[t]
	\centering
	\includegraphics[width=0.99\linewidth]{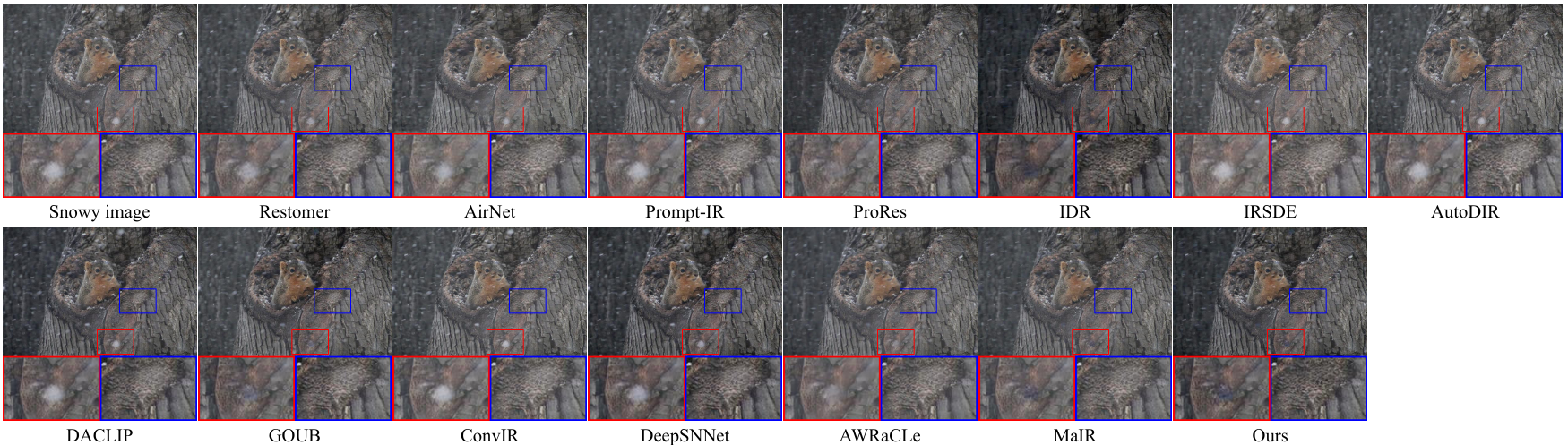}
	\vspace{-0.14in}
	\caption{Visualization comparison of desnowing task in real-world scenarios. Zoom in for best view.}
	\label{fig:real_snow}
\end{figure*} 

\renewcommand\thefigure{S9}
\begin{figure*}[t]
	\centering
	\includegraphics[width=0.99\linewidth]{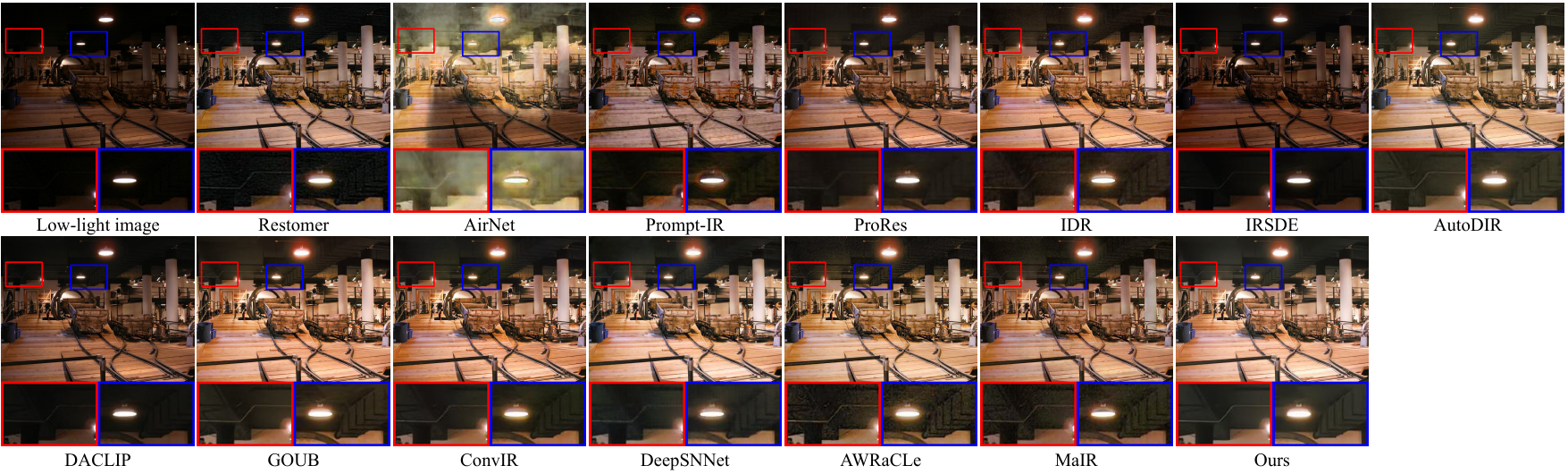}
	\vspace{-0.14in}
	\caption{Visualization comparison of low-light enhancement task in real-world scenarios. Zoom in for best view.}
	\label{fig:real_lowlight}
\end{figure*} 
 
\subsection{More Visual Comparisons on Real-world Scene Generalization}\label{sec:suppl_H-3}
\noindent{\textbf{Known task generalization.}} 
We randomly select 20 samples for each task to conduct the non-reference assessment, as presented in Tab.~\ref{tab:known}. Furthermore, to fully demonstrate that our method can handle the real-world restoration tasks, we have generalized all well-optimized models to five known tasks within real-world scenarios. Visual comparisons are displayed in Fig.~\ref{fig:real_blur}, Fig.~\ref{fig:real_haze}, Fig.~\ref{fig:real_rain}, Fig.~\ref{fig:real_snow}, and Fig.~\ref{fig:real_lowlight}, respectively. Clearly, our method produces the highest-quality restored images.

\noindent{\textbf{Unknown task generalization.}} Unknown task image restoration is performed on both POLED and TOLED~\cite{zhou2021image}. Visual comparisons on the POLED dataset are provided in Fig.~\ref{fig:poled}. The results show that our method can generalize to real-world scenes and achieve competitive visual results.

\renewcommand\thefigure{S10}
\begin{figure*}[t]
	\centering
	\includegraphics[width=0.99\linewidth]{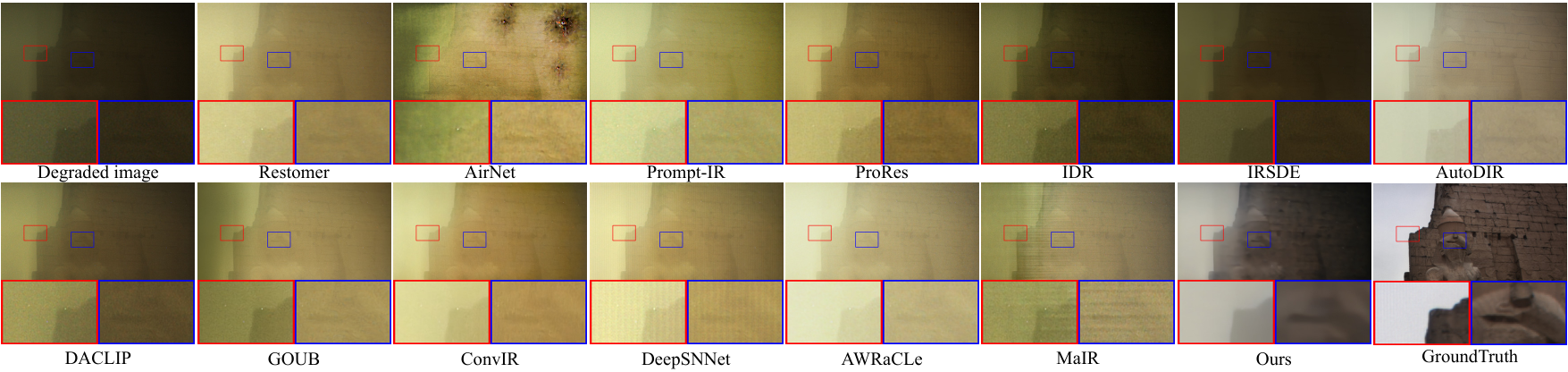}
	\vspace{-0.14in}
	\caption{Visualization results of zero-shot generalization in real-world POLED dataset. Zoom in for best view.}
	\label{fig:poled}
\end{figure*}

\subsection{More Visual Comparisons on Image Translation and Inpainting}\label{sec:suppl_H-4}
To further show the visual advantages of our approach across tasks, we present additional comparisons for image translation (Fig.~\ref{fig:trans2}) and image inpainting (Fig.~\ref{fig:inpai2}). In image translation, our method better preserves semantic and structural consistency, produces more faithful colors, sharper edges, and richer details. In image inpainting, it synthesizes textures and boundaries highly consistent with the surrounding context while avoiding oversmoothing and texture drift. Overall, our qualitative results show clearer details, stronger global consistency, and fewer visual artifacts than competing methods.

\renewcommand\thefigure{S11}
\begin{figure*}[t]
	\centering
	\includegraphics[width=0.99\linewidth]{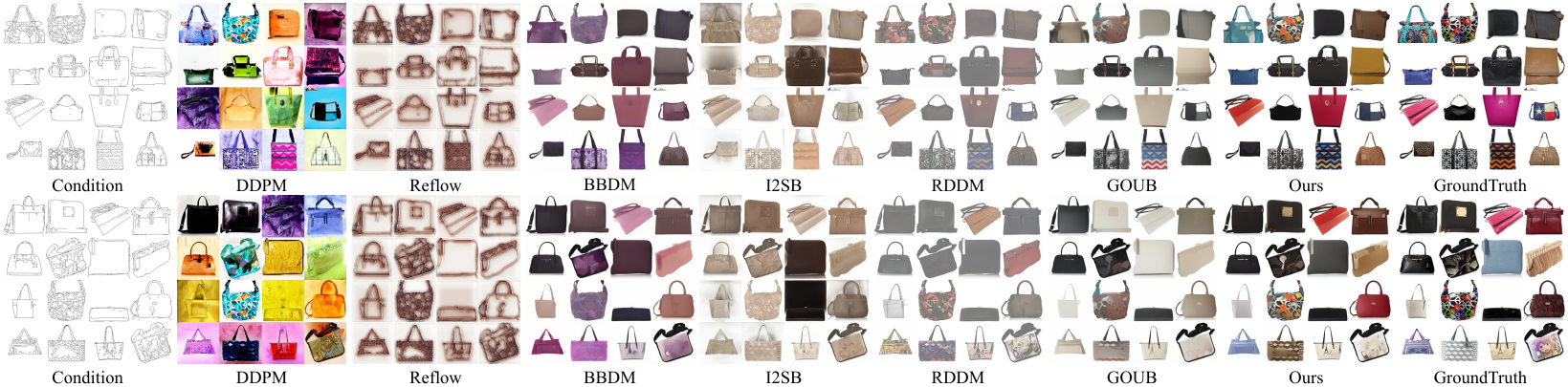}
	\vspace{-0.14in}
	\caption{Visualization results of image translation. Zoom in for best view.}
	\label{fig:trans2}
\end{figure*}

\renewcommand\thefigure{S12}
\begin{figure*}[t]
	\centering
	\includegraphics[width=0.99\linewidth]{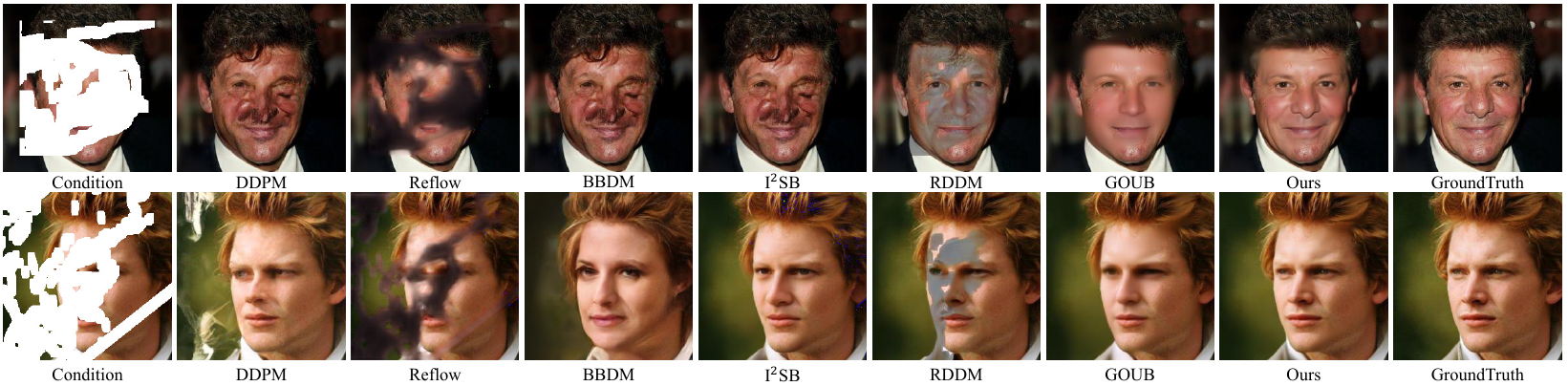}
	\vspace{-0.14in}
	\caption{Visualization results of image inpainting. Zoom in for best view.}
	\label{fig:inpai2}
\end{figure*}

\subsection{Efficiency Comparison}
Our mixed dataset consists of images with resolutions ranging from 256 to 1024 pixels. Accordingly, we evaluate model efficiency under three representative resolution settings, as summarized in Tab.~\ref{tab:effiency}. Evidently, our methods are moderately efficient with reasonable resource consumption. Overall, RDBM strikes a balance between efficiency and performance.

\renewcommand\thetable{S2}
\begin{table}[htbp]
	\centering
	\small 
	\caption{Efficiency comparisons among universal methods. '---' means out of memmory.}
	\label{tab:effiency}
	\begin{tabular}{l*{9}{c}}
		\toprule
		Resolution & \multicolumn{3}{c}{256$\times$256} & \multicolumn{3}{c}{512$\times$512} & \multicolumn{3}{c}{1024$\times$1024} \\
		\cmidrule(lr){2-4}\cmidrule(lr){5-7}\cmidrule(lr){8-10}
		Method & Mem.(G) & Time(s) & FPS & Mem.(G) & Time(s) & FPS & Mem.(G) & Time(s) & FPS \\
		\midrule
		Restomer~\cite{zamir2022restormer} & 1.959 & 0.105 & 9.563 & 6.670 & 0.381 & 2.622 & 25.419 & 1.773 & 0.564 \\ 
		AirNet~\cite{li2022all}  &1.039 & 0.194 & 5.159 & 3.480 & 0.738 & 1.355 & 11.244 & 20.499 & 0.049 \\
		Prompt-IR~\cite{potlapalli2023promptir}& 2.544 & 0.111 & 8.981 & 7.255 & 0.399 & 2.508 & 26.005 & 1.845 & 0.542 \\
		ProRes~\cite{ma2023prores} & 2.027 & 0.318 & 3.149 & 2.514 & 0.766 & 1.305 & 6.025  & 1.715 & 0.583 \\
		IDR~\cite{zhang2023ingredient} & 1.340 & 0.052 & 19.253& 4.313 & 0.136 & 7.373 & 16.110 & 0.615 & 1.626 \\ 
		IRSDE~\cite{luo2023image} & 1.554 & 5.017 & 0.199& 2.743 & 18.493 & 0.054& 9.997 & 72.289 & 0.014 \\
		AutoDIR~\cite{jiang2024autodir} & 7.023 & 6.266 & 0.160 & 11.021& 11.986& 0.083 & ---     & ---    & ---    \\
		DA-CLIP~\cite{luo2024controlling}& 2.119 & 2.585 & 0.387 & 6.775 & 7.937 & 0.126 & 58.548 & 60.893& 0.016 \\ 
		GOUB~\cite{yue2024image} & 1.554 & 4.996 & 0.200& 2.868 & 18.442 & 0.054& 10.122 & 72.239 & 0.014  \\
		ConvIR~\cite{cui2024revitalizing} & 0.708 & 0.035 & 28.570& 1.184 & 0.055 & 18.020& 2.809 & 0.192 & 5.202 \\
		DeepSNNet~\cite{deng2025deepsn} & 0.862 & 0.100 & 9.974& 0.989 & 0.102 & 9.801& 1.364 & 0.267 & 3.749 \\
		AWRaCLe~\cite{rajagopalan2025awracle} & 1.929 & 0.101 & 9.922& 4.264 & 0.354 & 2.822& 13.608 & 1.374 & 0.728\\
		MaIR~\cite{li2025mair} & 2.091 & 1.297 & 0.771& 6.593 & 4.744 & 0.211& 24.593 & 18.029 & 0.055 \\
		RDBM-T & 0.813 & 0.418 & 2.392& 1.907 & 1.621 & 0.617& 13.407 & 6.648 & 0.150 \\
		RDBM-S & 0.825 & 0.431 & 2.322& 1.921 & 1.648 & 0.607& 13.421 & 6.775 & 0.148 \\
		RDBM-B & 1.124 & 0.480 & 2.081& 2.186 & 1.926 & 0.519& 14.938 & 7.872 & 0.127 \\
		RDBM-L & 1.150 & 0.504 & 1.986& 2.307 & 1.982 & 0.505& 15.059 & 8.135 & 0.123  \\
		\bottomrule
	\end{tabular}
\end{table}

\setcounter{equation}{0}
\renewcommand{\theequation}{\thesection.\arabic{equation}} 
\renewcommand{\thesection}{I}
\section{Discussions, Limitations, and Future Work}\label{sec:suppl_I} 
\noindent{\textbf{Limitations and broader impact.}} The main challenge lies in fully exploring the connections between the data and prior distribution to modify the diffusion process. Although we have theoretically proposed a general and analytical formulation for diffusion bridge models, our core analysis assumes a fixed drift-to-diffusion coefficient ratio $\lambda = \sigma_t^2 / (2\theta_t) $ to admit closed-form solutions of SDEs. In the fields of image restoration, translation and inpainting where the data and prior distributions share semantic or structural affinity, our method is highly flexible and robust with competitive performance. However, it may be sub-optimal when applied to the generative tasks, where the distributions lack direct correspondence. Despite current limitations, we believe our unified model offers a strong foundation for diffusion bridge models.

\noindent{\textbf{Future Work.}} Future work could be explored in several promising directions. (1) With the rise of high-resolution imagery (e.g., 4K, 8K), developing multi-dimensional latent diffusion bridge models is crucial to address the computational demands. (2) Exploring more efficient network architectures to reduce memory usage and enhance efficiency. (3) Expanding the model capacity and datasets to strengthen restoration performance and generalization. (4) Designing adaptive learning rate schedules or applying model distillation to reduce sampling steps and improve restoration quality.

\noindent{\textbf{Reproducibility.}} Source code is provided in supplementary materials and will be released upon camera-ready submission. 
\end{document}